\documentclass{article}

\usepackage{microtype}
\usepackage{graphicx}
\usepackage{adjustbox}
\usepackage{subcaption}
\usepackage{wrapfig}
\usepackage{float}
\usepackage{booktabs}
\usepackage{hyperref}

\newcommand{\emojiimg}[1]{\raisebox{-0.15em}{\includegraphics[height=0.9em]{#1}}}
\newcommand{\variota}{\mathrm{I}}  %

\usepackage[preprint]{icml2026}
\usepackage{colortbl}
\usepackage{multirow}
\usepackage{amsmath}
\usepackage{amssymb}
\usepackage{mathtools}
\usepackage{amsthm}
\usepackage{epigraph} 
\usepackage{centernot}

\usepackage[capitalize,noabbrev]{cleveref}
\usepackage{titletoc}
\usepackage{tikz}
\usetikzlibrary{arrows.meta,positioning,calc,shapes.geometric,fit,backgrounds}
\usepackage[skins,breakable]{tcolorbox}
\tcbset{breakable}

\newcommand{\wrapfigspan}{0.37\textwidth}
\newcommand{\besidewrap}{\dimexpr\linewidth-\wrapfigspan-\columnsep\relax}
\tcbset{besidewrap/.style={width=\besidewrap}}

\newtcolorbox{takeaway}{
  colback=black!4, colframe=black!60, boxrule=0.5pt, arc=1pt,
  left=6pt, right=6pt, top=4pt, bottom=4pt,
  fonttitle=\bfseries, title=Takeaway
}

\newtheorem{theorem}{Theorem}[section]

\theoremstyle{definition}
\newtheorem{definition}[theorem]{Definition}

\theoremstyle{remark}

\definecolor{cellgain}{HTML}{2E9E5B}
\definecolor{cellloss}{HTML}{D1495B}
\definecolor{perspO}{HTML}{4C72B0}  %
\definecolor{perspS}{HTML}{DD8452}  %
\definecolor{perspC}{HTML}{55A868}  %
\definecolor{boundary}{HTML}{6C6C6C} %
\definecolor{routerExp}{HTML}{8172B3} %
\definecolor{routerRef}{HTML}{B8860B} %
\definecolor{routerRst}{HTML}{C0392B} %
\definecolor{routerStop}{HTML}{404040} %

\renewcommand{\algorithmiccomment}[1]{\hfill{\color{black!55}\#\,#1}}
\definecolor{ethosurvey}{HTML}{4C72B0}
\definecolor{ethotheorize}{HTML}{DD8452}
\definecolor{ethoformalize}{HTML}{55A868}
\definecolor{ethorulemake}{HTML}{C44E52}
\definecolor{ethoenact}{HTML}{8172B3}
\definecolor{ethorevise}{HTML}{937860}
\newcommand{\ethokey}[2]{{\setlength{\fboxsep}{1.2pt}\colorbox{#1}{\textcolor{white}{\sffamily\bfseries\small #2}}}}

\icmltitlerunning{Small Models Think Better When They Reflect Polycontexturally}

\begin{document}

\makeatletter
\if@filesw
  \expandafter\newwrite\csname tf@toc\endcsname
  \immediate\openout \csname tf@toc\endcsname \jobname.toc\relax
\fi
\makeatother

\twocolumn[
\icmltitle{Small Models Think Better When They Reflect Polycontexturally}

\icmlsetsymbol{equal}{*}

\begin{icmlauthorlist}
\icmlauthor{Habiba Sarhan}{gaia}
\icmlauthor{Simon Hegelich}{gaia}
\icmlauthor{Morteza Shahrezaye}{gaia}
\icmlauthor{Linda Schannen}{gaia}
\icmlauthor{Matthias Balk}{gaia}
\icmlauthor{Erik Cambria}{singapore}
\icmlauthor{Christian Montag}{macau}
\end{icmlauthorlist}

\icmlaffiliation{gaia}{Evercursion -- GaiaLogic Unipessoal Lda., Portugal}
\icmlaffiliation{singapore}{Nanyang Technological University, Singapore}
\icmlaffiliation{macau}{University of Macau}

\icmlcorrespondingauthor{Simon Hegelich}{simon.hegelich@evercursion.ai}

\icmlkeywords{self-reflection, test-time compute, polycontextural logic, meta-cognition, LLM reasoning}

\vskip 0.3in
]

\printAffiliationsAndNotice{}

\begin{abstract}
Scaling test-time compute with longer chains or more votes within a single valuation frame shows diminishing returns: extended chains can drift or restate their premises with greater confidence. We study a different variable, the structure of reflection across irreducible standpoints. We introduce a two-layer architecture for small language models. A reflection engine recursively splits an input across three perspectives (Objective, Subjective, Conceptual) and their pairwise divergence boundaries, then integrates the outermost leaves into one summary. A router has a language model judge each completed pass from the same three standpoints, and a deterministic state machine maps the eight verdict triples onto four global values that decide whether to expand, refine, reset, or stop. On GPQA Diamond, ExploreToM, Hi-ToM, five BIG-Bench Hard tasks and MMLU-Pro across five small models, the router with a four-cycle budget beats direct answering in 19 of 27 model--task pairs, by up to \texttimes 5.2 on state tracking. It loses on Hi-ToM, ExploreToM and exact word sorting for both 3B models, and on MMLU-Pro for Qwen2.5-3b.
\end{abstract}

\section{Introduction}
\epigraph{\textit{[$\dots$] we've achieved peak data, and there'll be no more [$\dots$]}}{Ilya Sutskever --- NeurIPS'24.}
Progress toward more general machine intelligence is, at present, pursued largely as a scaling problem. Train a larger model on more data with more compute, and capability is assumed to follow as a predictable, near-automatic consequence~\citep{kaplan2020scaling}. That premise has already been revised once. The original scaling laws overstated the share of a fixed compute budget that should go to model size rather than to training data. The correction changed how compute-optimal models are trained~\citep{hoffmann2022chinchilla}.

Further work shows that some of the scale's most celebrated successes are not a real property of the underlying model. Capabilities that appear to switch on suddenly once a model crosses a size threshold are better explained as an artifact of the metric used to score them~\citep{schaeffer2023emergent}. This is no longer only an academic objection. By late 2024, researchers at the frontier labs driving the scaling bet were themselves reporting that returns from pretraining scale had plateaued~\citep{hu2024reuters}. That is part of why effort across the field shifted toward scaling \emph{inference-time} compute instead. We take the plateau in training-time scaling as a reason to apply the same scrutiny to this newer claim.

The conventional wisdom has long suggested that letting a model think longer, critique its own output, and explore multiple candidate paths---chain-of-thought prompting~\citep{wei2022chain}, self-consistency over reasoning paths~\citep{wang2022self}, iterative self-refinement~\citep{madaan2023self}, and Tree-of-Thoughts~\citep{yao2023tree}---reliably improves performance on complex tasks by investing more structured computation per answer. 

Recent evidence complicates this view. Longer reasoning chains can introduce drift, compound errors, and restate original answers with added confidence rather than actual correctness. Extended sequential thinking can produce an illusion of improved reasoning because of how evaluation metrics interact with increased output variance, causing accuracy to decline past an early peak rather than continuing to climb~\citep{ghosal2025mirage}. Furthermore, parallel research highlights a fundamental disagreement within the field regarding what test-time scaling methods are actually spending their computational budgets on, which ultimately limits how broadly any single reported scaling curve can generalize~\citep{hariri2026testtime}. The difficulty is not confined to how compute is spent. Evaluations of deductive, inductive, and abductive reasoning that score the soundness of a model's explanation, and not only its final answer, find that LLMs are considerably weaker logical reasoners than answer accuracy suggests~\citep{xu2025large}. More broadly, models that operate on statistical patterns can produce fluent, contextually appropriate text without the deeper comprehension that natural language understanding requires~\citep{cambria2024understanding}.\\

We ask whether a second variable matters besides the amount of compute: the \emph{structure} of a reasoning process, that is, what decides to expand a line of reasoning, refine it, or stop. Our experiments compare against direct answering only, so they do not separate structure from the additional model calls it makes (\Cref{sec:limitations}). We therefore treat the role of structure as a hypothesis that our results bear on, not one they settle.

This paper studies recursive reflection, and its underlying intuition is simple. A piece of reasoning can always be examined from three angles at once: what in it is objectively the case, what a reasoner subjectively takes to be the case, and what general principle organizes the two. We formalize a reflection operator that recursively splits a text across these three perspectives into a self-similar tree whose leaves reintegrate into a single summary, and a meta-controller that has a language model judge the completed recursion from the same three perspectives and, through a deterministic state machine, decides whether to expand, refine, reset, or stop. Both layers are fixed by construction. The branching factor, the judging prompts, and the transitions are neither learned nor searched over, so the procedure we test is fixed before any benchmark is run. The architecture is therefore neurosymbolic: a language model generates every text, and a fixed symbolic layer decides what happens to it~\citep{senticnet}.

\section{Method}\label{sec:method}

The architecture rests on two claims.

The first is that a system reflecting on its own output is in two positions simultaneously. It writes the reasoning, and it evaluates it. G\"unther argues that one place of judgment cannot cover both, and that comparing two standpoints needs a third one to mediate between them~\citep{gunther1973life}. We therefore use three perspectives and treat them equally.

The second concerns what repeated reflection buys. Running the same reflection again on the same material adds nothing~\citep{hegelich2026corecursive}. Getting past that point means changing the input to the next pass from outside the reflection itself. Our router does this with the judges' verdicts. Whatever the judges reject is written into the next prompt as an explicit constraint. This is why they return reasons in words, which can indicate what to change.

\Cref{sec:gunther} gives the full argument of our choice and the four-valued logic behind it.

\subsection{Layer 1: The Sierpinski Reflection Engine}\label{sec:engine}
\begin{figure}[t]
\centering
\resizebox{\linewidth}{!}{%
\begin{tikzpicture}[scale=0.9, every node/.style={font=\footnotesize}]
  \coordinate (O) at (0,3.2);
  \coordinate (S) at (-2.1,0);
  \coordinate (B) at (2.1,0);
  \coordinate (mOS) at ($(O)!0.5!(S)$);
  \coordinate (mOB) at ($(O)!0.5!(B)$);
  \coordinate (mSB) at ($(S)!0.5!(B)$);

  \draw[thick] (O)  --  (S)  --  (B)  --  cycle;
  \filldraw[black!4] (mOS) -- (mOB) -- (mSB) -- cycle;
  \draw[dashed, boundary] (mOS)  --  (mOB)  --  (mSB)  --  cycle;
  \filldraw[boundary] (mOS) circle (1.4pt);
  \filldraw[boundary] (mOB) circle (1.4pt);
  \filldraw[boundary] (mSB) circle (1.4pt);

  \filldraw[perspO] (O) circle (2.1pt);
  \filldraw[perspS] (S) circle (2.1pt);
  \filldraw[perspC] (B) circle (2.1pt);

  \node[above, perspO] at (O) {\textbf{O}: $\varphi_O$};
  \node[below left, perspS] at (S) {\textbf{S}: $\varphi_S$};
  \node[below right, perspC] at (B) {\textbf{C}: $\varphi_C$};

  \node[left,align=right, boundary] at ($(mOS)+(-0.2,0.05)$) {$\psi_{OS}$};
  \node[right,align=left, boundary] at ($(mOB)+(0.2,0.05)$) {$\psi_{OC}$};
  \node[below, boundary] at (mSB) {$\psi_{SC}$};

  \node[align=center] at (0,1.05) {$T(x,\Phi)$};

  \draw[-{Latex[length=2.2mm]},thick] (0.3,-0.4) to[bend right=50]
    node[below,pos=0.5,align=center,font=\scriptsize]{recurse into each corner:\\depth $n \to n{+}1$} (3.6,-0.4);

  \begin{scope}[xshift=5cm,scale=0.5]
    \coordinate (O2) at (0,3.2);
    \coordinate (S2) at (-2.1,0);
    \coordinate (B2) at (2.1,0);
    \draw[thick] (O2)  --  (S2)  --  (B2)  --  cycle;
    \coordinate (mOS2) at ($(O2)!0.5!(S2)$);
    \coordinate (mOB2) at ($(O2)!0.5!(B2)$);
    \coordinate (mSB2) at ($(S2)!0.5!(B2)$);
    \draw[dashed, boundary] (mOS2)  --  (mOB2)  --  (mSB2)  --  cycle;
    \filldraw[perspO] (O2) circle (3.0pt);
    \filldraw[perspS] (S2) circle (3.0pt);
    \filldraw[perspC] (B2) circle (3.0pt);
    \node[above,font=\scriptsize, perspO] at (O2) {O};
    \node[below left,font=\scriptsize, perspS] at (S2) {S};
    \node[below right,font=\scriptsize, perspC] at (B2) {C};
    \node[align=center,font=\scriptsize] at (0,-0.9) {depth $n{+}1$};
  \end{scope}
\end{tikzpicture}%
}
\caption{\textbf{One level of the Sierpinski reflection engine.} $\varphi_O,\varphi_S,\varphi_C$ sit at the vertices, divergence boundaries $\psi_{OS},\psi_{OC},\psi_{SC}$ sit at the edge midpoints, and the outermost corner leaves $\Phi$ are what the pass $T(x,\Phi)$ integrates with the input into $T^n(x)$.}
\label{fig:sierpinski}
\end{figure}
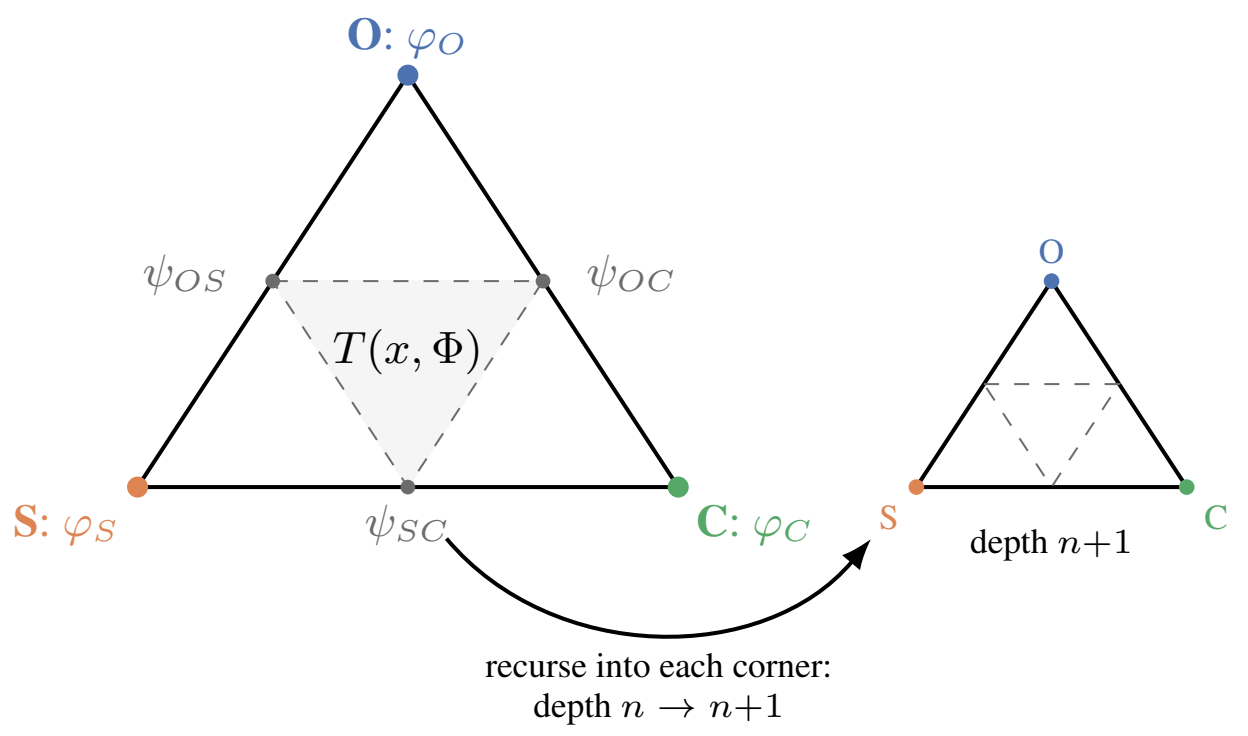

We give Günther's three-contexture reflection a recursive geometric form.\footnote{\url{https://github.com/gaialogicag/polyreflexion}} The three perspectives are placed at the vertices of an equilateral triangle, and the construction is applied recursively to its own corner sub-triangles after the Sierpinski gasket (\Cref{fig:sierpinski}). The triangle is a picture of the construction, not a space we compute in. The three vertices are the perspectives. The midpoint of each edge is the divergence statement for that pair of perspectives. The middle triangle those three midpoints form is the one the gasket removes, which is why the boundaries are computed and carried into the corner regions as constraints rather than recursed into. 

Let $\mathcal{X}$ be the set of finite text strings and $V = \{O, S, C\}$ the vertex set (\textbf{O}bjective, \textbf{S}ubjective, \textbf{C}onceptual). 

\begin{definition}
\label{def:perspective}
\textit{A perspective projection is a family of functions $\{\varphi_v : \mathcal{X} \times \mathcal{R} \to \mathcal{X}\}_{v \in V}$ conditioned on a region context $r \in \mathcal{R}$.}
\end{definition}
Each projection is one prompt (\S\ref{sec:prompt-perspective}). $\varphi_O$ extracts the facts, quantities, and constraints that must hold; $\varphi_S$ names the standpoint a reasoner takes and the distractors that mislead it; $\varphi_C$ states the principle that organizes the other two.

\begin{definition}
\label{def:boundary}
\textit{For each edge $\{u,v\} \subset V$, a {boundary function} $\psi_{uv} : \mathcal{X} \times \mathcal{R} \to \mathcal{X}$ maps an input $x$ within region $r$ to an explicit statement of where the two adjacent contextures $u$ and $v$ diverge. }
\end{definition}%
There are three boundary functions, $\psi_{OS}, \psi_{OC}, \psi_{SC}$, one per edge of the triangle. Each is one prompt (\S\ref{sec:prompt-boundary}). 

\begin{definition}
\label{def:reflection}
\textit{The {reflection operator} $T : \mathcal{X} \times \mathcal{X}^{|V|} \to \mathcal{X}$ integrates an input with the three corner perspectives it was split into, into a single combined summary,
\begin{equation}
\label{eq:reflection}
y = T(x, \Phi), \qquad \Phi = \{\varphi_v(x,r_v)\}_{v\in V},
\end{equation}
where $r_v$ carries the adjacent boundary statements $\psi_{uv}$ into corner $v$.}
\end{definition}
The integration itself is one prompt (\S\ref{sec:prompt-integration}), and the answering function $\beta$ is another (\S\ref{sec:prompt-answering}).

\begin{figure}[t]
\centering
\resizebox{\columnwidth}{!}{%
\begin{tikzpicture}[
  >=Latex, line width=0.5pt,
  box/.style={draw, align=left, inner sep=3pt, font=\scriptsize},
  boxO/.style={box, draw=perspO, fill=perspO!7, line width=0.7pt},
  boxS/.style={box, draw=perspS, fill=perspS!7, line width=0.7pt},
  boxC/.style={box, draw=perspC, fill=perspC!7, line width=0.7pt},
  shaded/.style={box, fill=black!5},
  psi/.style={align=center, inner sep=1pt, font=\tiny, text width=3.0cm, text=boundary},
  vtx/.style={circle, fill=black, inner sep=1.3pt},
  stepnum/.style={circle, fill=black!78, text=white, minimum size=0.30cm, inner sep=0pt,
               font=\tiny\bfseries}
]
\node[shaded, text width=8.0cm, anchor=south] (story) at (0,1.55) {%
\textit{x}: ``Brandon effortlessly dealt with the \textbf{stuffed rabbit}, disposing of it by
placing it within a \textbf{cardboard box}\dots\ meanwhile, Parker secretly applied his
observational skills to Brandon's movements\dots\ Brandon moved straight to a trusty
\textbf{storage bin} and slipped the stuffed rabbit inside.''};

\node[boxO, text width=3.8cm, anchor=south, align=center] (aO) at (0,0.35) {%
{\color{perspO}$\varphi_O$:} ``Brandon arrived and placed a stuffed rabbit in a \textbf{cardboard box}, then moved
it to a \textbf{storage bin}.''};

\coordinate (O) at (0,0);   \coordinate (S) at (-2.5,-3.3); \coordinate (C) at (2.5,-3.3);
\coordinate (mOS) at (-1.25,-1.65); \coordinate (mOC) at (1.25,-1.65); \coordinate (mSC) at (0,-3.3);
\draw[line width=0.8pt] (O) -- (S) -- (C) -- cycle;
\filldraw[black!4] (mOS) -- (mOC) -- (mSC) -- cycle;
\draw[dashed, boundary] (mOS) -- (mOC) -- (mSC) -- cycle;
\filldraw[perspO] (O) circle (2.2pt);
\filldraw[perspS] (S) circle (2.2pt);
\filldraw[perspC] (C) circle (2.2pt);
\foreach \p in {mOS,mOC,mSC}{\filldraw[boundary] (\p) circle (1.6pt);}
\node[circle, draw, fill=white, minimum size=0.42cm, inner sep=0pt, font=\tiny] (alpha) at (0,-2.2) {$T$};
\draw[dashed, boundary, ->, shorten >=1pt, looseness=1.5] (-3.9,1.55) to[out=-90,in=215] (alpha);
\draw[dashed, perspO, ->, shorten >=1pt] (O) -- (alpha);
\draw[dashed, perspS, ->, shorten >=1pt] (S) to[out=40,in=200] (alpha);
\draw[dashed, perspC, ->, shorten >=1pt] (C) to[out=140,in=340] (alpha);
\draw[dashed, boundary, ->, shorten <=1pt, looseness=1.6] (alpha) to[out=0,in=60] (2.6,-6.55);
\draw[->, boundary] (mOS) -- (-1.97,-1.15);
\draw[->, boundary] (mOC) -- (1.97,-1.15);
\draw[->, boundary] (mSC) -- (0,-4.75);

\node[psi, rotate=53] at (-2.78,-1.3) {%
$\psi_{OS}$: ``\dots observable actions and events, like \textbf{Brandon placing a rabbit in a box},
such as \textbf{Parker understanding the significance}\dots''};
\node[psi, rotate=-53] at (2.78,-1.36) {%
$\psi_{OC}$: ``\dots the sun's glow or \textbf{the placement of a stuffed rabbit}, versus the
internal thoughts of characters, like \textbf{Parker understanding the significance}\dots''};
\node[psi, text width=3.8cm, anchor=north] at (0,-4.85) {%
$\psi_{SC}$: ``\dots the objective descriptions of the scene and characters' actions, and
\textbf{the internal thoughts} and interpretations of those actions\dots\ \textbf{Parker's
subjective understanding}\dots''};

\node[boxS, text width=3.4cm, anchor=north east] at (-0.85,-3.7) {%
{\color{perspS}$\varphi_S$:} ``\textbf{I was curious about} what he was doing as he discreetly put a stuffed rabbit
into a box.''};
\node[boxC, text width=3.4cm, anchor=north west] at (0.85,-3.7) {%
{\color{perspC}$\varphi_C$:} ``\textbf{Parker understood the} hidden meaning behind Brandon's seemingly simple
actions.''};

\node[stepnum] at ($(story.north west)+(0.26,-0.21)$) {1};
\node[stepnum] at (-1.64,-1.36) {2};
\node[stepnum] at (1.64,-1.36)  {2};
\node[stepnum] at (0,-4.26)  {2};
\node[stepnum] at (0.34,0.07)   {3};
\node[stepnum] at (-2.86,-3.3)  {3};
\node[stepnum] at (2.86,-3.3)   {3};
\node[stepnum] at (3.1,-5.52)  {4};
\node[shaded, text width=8.0cm, anchor=north] (sum) at (0,-6.6) {%
Summary $T^1(x)$: ``Brandon arrived and discreetly placed a stuffed rabbit into a \textbf{cardboard
box}, then moved it to a \textbf{storage bin}, an action \textbf{Parker observed and understood the
significance of}.''};
\end{tikzpicture}%
}
\caption{\textbf{${T^1(x)}$ on a real ExploreToM story}. The engine runs the input (1), the three boundary statements (2), the three corner answers, resolved in parallel (3), and the integration (4). Each corner answers under the two boundaries that touch it, and $T$ merges the three corner answers with the input.}
\label{fig:worked-example}
\end{figure}
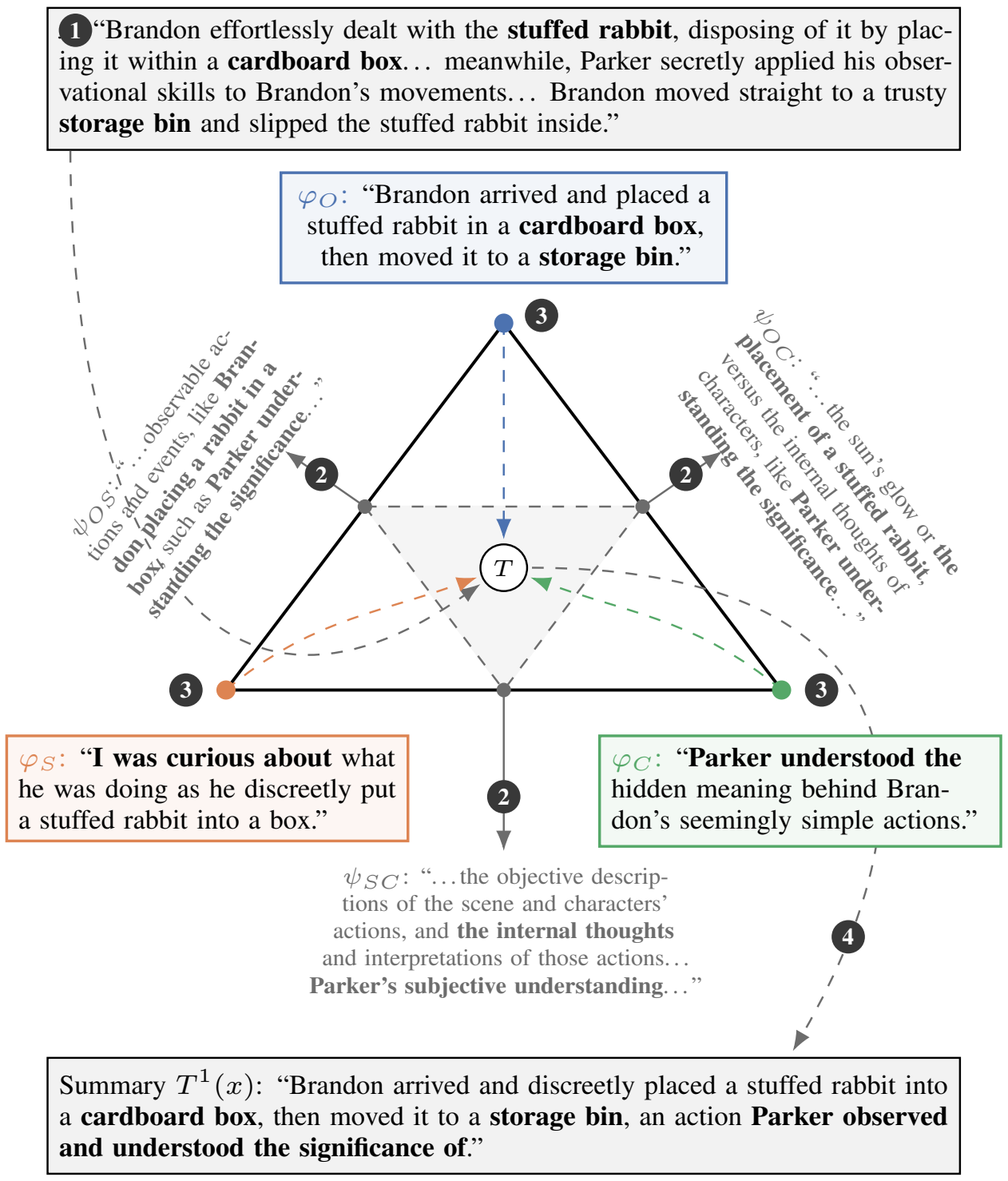

\Cref{fig:worked-example} shows one level. For each edge, the engine writes a short statement of where its two perspectives disagree about the input. Each corner then takes the two sentences written on the edges that meet at it. The \textbf{O}bjective corner takes $\psi_{OS}$ and $\psi_{OC}$. Inside that corner, the engine resolves all three perspectives on the input under the corner's region context, which names the corner's own perspective as the apex and carries the two constraints. It keeps the apex perspective, here the objective one, as the corner answer. The \textbf{S}ubjective and \textbf{C}onceptual corners work the same way. At depth $1$, the three corner answers are what $T$ merges with the input. A deeper recursion repeats the same step inside each corner, and the answer taken from a corner is the apex perspective of the leaf reached by staying in that corner all the way down. An $n$-depth pass $T^n$ therefore builds $3^n$ leaves, reads the three outermost corners, and applies $T$ once to those. We write $\Phi^n(x) = \{\varphi^n_v(x)\}_{v \in V}$ for those three corners, with $\varphi^1_v(x) = \varphi_v(x, r_v)$, and abbreviate the whole pass as $T^n(x) = T\bigl(x, \Phi^n(x)\bigr)$. The perspectives are an input to the pass, not a second output of it, so the router can read the same $\Phi^n(x)$ the engine integrated. $T^n$ is not $T$ composed with itself $n$ times: the input $x$ is passed down the chain, and only the region is refined, so the integration runs once per pass rather than once per level. In this notation, $x \equiv T^0(x)$. A question $q$ is then answered from that summary by $\beta(T^n(x), q)$, so direct answering is $\beta(T^0(x), q) \equiv \beta(x, q)$.

$T^n$ cannot judge its own output. Therefore, there is no way to tell whether the reported summary dropped information, hallucinated, or just restated what $T^{n-1}$ did. This is why we need a meta-steering layer on top.

\subsection{Layer 2: The Polycontextural Logic Router}
\label{sec:meta}

The polycontextural logic router is that layer. It runs the engine more than once. One engine run, the judgment of what that run produced, and the decision of what to do next together make one \emph{cycle}. Cycles are numbered $c = 0, 1, \dots, c_{\max}-1$, with the budget $c_{\max}$ fixed before the run. Cycle $0$ runs $T^n$ on the original input $x$. Each later cycle runs $T^n$ on an input the router wrote. The router therefore steers the engine by choosing what it reads next, and never by editing what it has already written.

Let $y_c$ be the summary the engine produces in cycle $c$. Within the same cycle, the router then does three things, each defined below. It scores $y_c$ from the same three perspectives the engine itself used (\Cref{def:polyjudge}). It treats the three scores as a single value (\Cref{def:interpreter}). It turns that value into the input the engine reads in cycle $c+1$ (\Cref{def:strategy}). The scoring is what the engine could not do on its own. The value determines whether cycle $c+1$ expands the reasoning, refines it, or resets it, or whether the loop stops. This is the co-recursive regime transition of~\citet{hegelich2026corecursive}.

\subsubsection{Step 1: Judging the Pass}
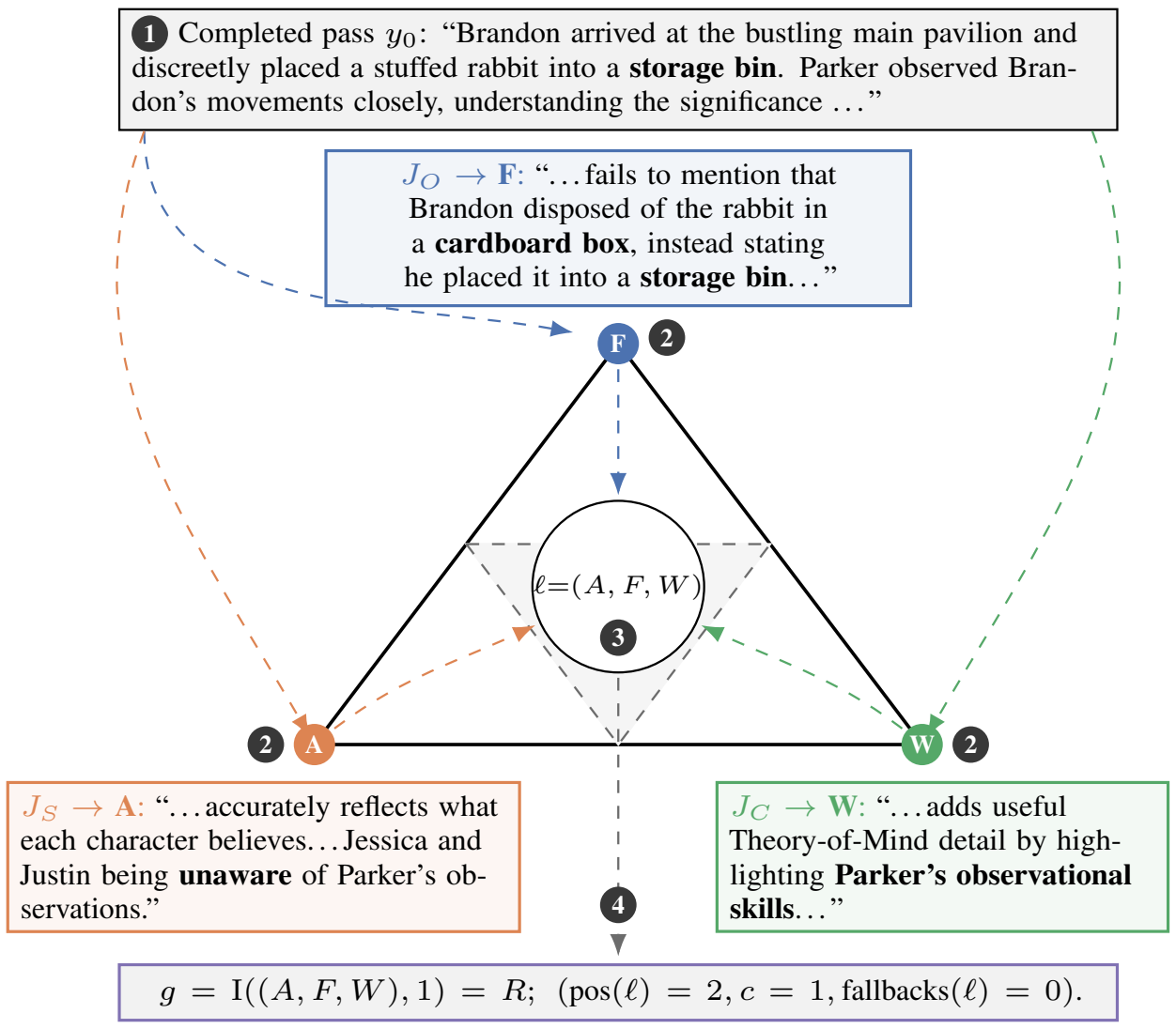
\begin{figure}[t]
\centering
\resizebox{\columnwidth}{!}{%
\begin{tikzpicture}[
  >=Latex, line width=0.5pt,
  box/.style={draw, align=left, inner sep=3pt, font=\scriptsize},
  boxO/.style={box, draw=perspO, fill=perspO!7, line width=0.7pt},
  boxS/.style={box, draw=perspS, fill=perspS!7, line width=0.7pt},
  boxC/.style={box, draw=perspC, fill=perspC!7, line width=0.7pt},
  shaded/.style={box, fill=black!5},
  vdict/.style={circle, fill=#1, text=white, minimum size=0.34cm, inner sep=0pt,
               font=\tiny\bfseries},
  vdict/.default=black!55,
  stepnum/.style={circle, fill=black!78, text=white, minimum size=0.30cm, inner sep=0pt,
               font=\tiny\bfseries}
]
\node[shaded, text width=8.0cm, anchor=south] (sum) at (0,1.75) {%
\quad\;\;Completed pass $y_0$: ``Brandon arrived at the bustling main pavilion and discreetly placed a stuffed
rabbit into a \textbf{storage bin}. Parker observed Brandon's movements closely, understanding
the significance \dots''};

\node[boxO, text width=4.6cm, anchor=south, align=center] (jO) at (0,0.3) {%
{\color{perspO}$J_O$ $\to$ \textbf{F}:} ``\dots fails to mention that Brandon disposed of the
rabbit in a \textbf{cardboard box}, instead stating he placed it into a \textbf{storage bin}\dots''};

\coordinate (O) at (0,0);   \coordinate (S) at (-2.5,-3.3); \coordinate (C) at (2.5,-3.3);
\coordinate (mOS) at (-1.25,-1.65); \coordinate (mOC) at (1.25,-1.65); \coordinate (mSC) at (0,-3.3);
\draw[line width=0.8pt] (O) -- (S) -- (C) -- cycle;
\filldraw[black!4] (mOS) -- (mOC) -- (mSC) -- cycle;
\draw[dashed, boundary] (mOS) -- (mOC) -- (mSC) -- cycle;

\node[circle, draw, fill=white, minimum size=0.78cm, inner sep=0pt, font=\tiny] (ell) at (0,-2.0)
  {$\ell{=}(A,F,W)$};

\draw[dashed, perspO, ->, shorten >=2pt] (-3.9,1.75) to[out=-90,in=170] (-0.30,0.08);
\draw[dashed, perspS, ->, shorten >=2pt] (-3.9,1.75) to[out=-110,in=120] (S);
\draw[dashed, perspC, ->, shorten >=2pt] (3.9,1.75)  to[out=-70,in=60]  (C);
\draw[dashed, perspO, ->, shorten >=1pt] (O) -- (ell);
\draw[dashed, perspS, ->, shorten >=1pt] (S) to[out=40,in=205] (ell);
\draw[dashed, perspC, ->, shorten >=1pt] (C) to[out=140,in=335] (ell);
\draw[dashed, boundary, ->, shorten <=1pt, shorten >=1pt] (ell) -- (0,-5.08);

\node[vdict=perspO] at (O) {F};
\node[vdict=perspS] at (S) {A};
\node[vdict=perspC] at (C) {W};

\node[boxS, text width=4cm, anchor=north east] at (-0.80,-3.6) {%
{\color{perspS}$J_S$ $\to$ \textbf{A}:} ``\dots accurately reflects what each character believes\dots
Jessica and Justin being \textbf{unaware} of Parker's observations.''};
\node[boxC, text width=3.5cm, anchor=north west] at (0.80,-3.6) {%
{\color{perspC}$J_C$ $\to$ \textbf{W}:} ``\dots adds useful Theory-of-Mind detail by highlighting
\textbf{Parker's observational skills}\dots''};

\node[stepnum] at ($(sum.north west)+(0.26,-0.21)$) {1};
\node[stepnum] at (0.40,0.05)   {2};
\node[stepnum] at (-2.90,-3.3)  {2};
\node[stepnum] at (2.90,-3.3)   {2};
\node[stepnum] at (0,-2.42)  {3};
\node[stepnum] at (0,-4.62)  {4};

\node[shaded, draw=routerExp, line width=0.7pt, text width=8.0cm, anchor=north, align=center]
  (act) at (0,-5.10) {%
$g=\variota((A,F,W), 1) = R; \;(\text{pos}(\ell)=2, c = 1, \text{fallbacks}(\ell)=0)$.};
\end{tikzpicture}%
}
\caption{\textbf{Steps 1--2 on the same ExploreToM story as \Cref{fig:worked-example}} (a different run, so its first pass differs from the summary shown there). The three judges read the completed first pass $y_0$ (cycle $c = 0$) and each returns a verdict and a rationale. This pass dropped the \textit{cardboard box}, and $J_O$ is the judge that catches it, so the triple is $(A,F,W)$.}
\label{fig:judging}
\end{figure}

The first step asks how the last pass went. One judge cannot answer that, for the same reason one perspective cannot read the input: the question takes a different form depending on which standpoint asks it.

The judges sit at the same vertices $V = \{O,S,C\}$ as the perspectives, and each asks the one question its own vertex can ask. \Cref{fig:judging} runs all three on a pass of the story of \Cref{fig:worked-example}. The judges run on a fixed meta-judge model, which is gpt-4o-mini in \Cref{tab:main,tab:mmlu-pro} and differs from the answerer except when gpt-4o-mini itself answers.

\noindent$\bullet$ $J_S$: \textit{is $y_c$ coherent with the three perspective texts?} Authentic ($A$) or inauthentic ($F$).

\noindent$\bullet$ $J_O$: \textit{is $y_c$ objectively defensible given the input $x$?} The default prompt asks whether $y_c$ is correct and defensible given $x$; the task-specific prompts ask whether it is faithful to $x$, penalizing invented facts and dropped constraints. Warranted ($W$) or false ($F$).

\noindent$\bullet$ $J_C$: \textit{does $y_c$ add anything to $x$ and to the previous cycle's summary $y_{c-1}$?} Productive ($W$) or redundant ($R$).

\noindent Each scale is written with its positive value first: $\mathcal{L}_S = \{A,F\}$, $\mathcal{L}_O = \{W,F\}$ and $\mathcal{L}_C = \{W,R\}$ (see §\ref{sec:gunther} for their derivation).

\begin{definition}
\label{def:polyjudge}
\textit{Poly-judge is the family of evaluation functions $\{J_v : \mathcal{X}^6 \to \mathcal{L}_v \times \mathcal{X}\}_{v \in V}$, run on the observation $(x, y_c, \Phi^n, y_{c-1})$. Each $J_v$ returns a verdict $\ell_v \in \mathcal{L}_v$ and a rationale $\rho_v \in \mathcal{X}$ naming the statements that drove it.}
\end{definition}
Each judge is one prompt, task-framed per benchmark (\S\ref{sec:prompt-judge}). A parse failure or an off-schema response defaults to the negative value of $\mathcal{L}_v$ and is marked with a fallback flag, so three unreadable responses give $(F,F,R)$. Because a fallback is always negative, a unanimous triple never rests on an unread response.

\subsubsection{Step 2: Interpreting the Triple}
Three separate verdicts are not yet a decision. The router needs one value it can act on. Across the three binary contextures, $\mathcal{L}_S \times \mathcal{L}_O\times\mathcal{L}_C$ contains $2^3 = 8$ possible triples $\ell = (\ell_S,\ell_O,\ell_C)$. Let $\text{pos}(\ell)\subseteq \{O,S,C\}$ denote the subset of vertices judged positively and $k = |\text{pos}(\ell)|$ its cardinality.

\begin{definition}
\label{def:interpreter}
\textit{The Interpreter $\variota: \mathcal{L}_S \times \mathcal{L}_O \times \mathcal{L}_C \times \mathbb{N}_0 \to \{W,R,A,F\}$ maps a valuation triple $\ell$ at cycle $c$ onto the global value set $\{W,R,A,F\}$:}
\begin{equation}\label{eq:interpreter}
   g =  \variota(\ell,c) = \begin{cases}
        W, & k = 3,\ c \geq 1\\
        A, & k = 1\\
        F, & k = 0\\
        R, & \textit{otherwise}
    \end{cases}.
\end{equation}
\end{definition}%

A unanimous triple does not stop the loop at cycle $0$. The first pass has not been steered yet, so at $c = 0$ the Interpreter flips $(A,W,W)$ to $(A,W,R)$: $k$ drops to $2$, the value is $R$, and the router expands. From cycle $1$ on, $(A,W,W)$ gives $W$. What each value then does is the next step. \Cref{tab:truth-router} writes all eight triples out.

\subsubsection{Step 3: Boundary Construction}
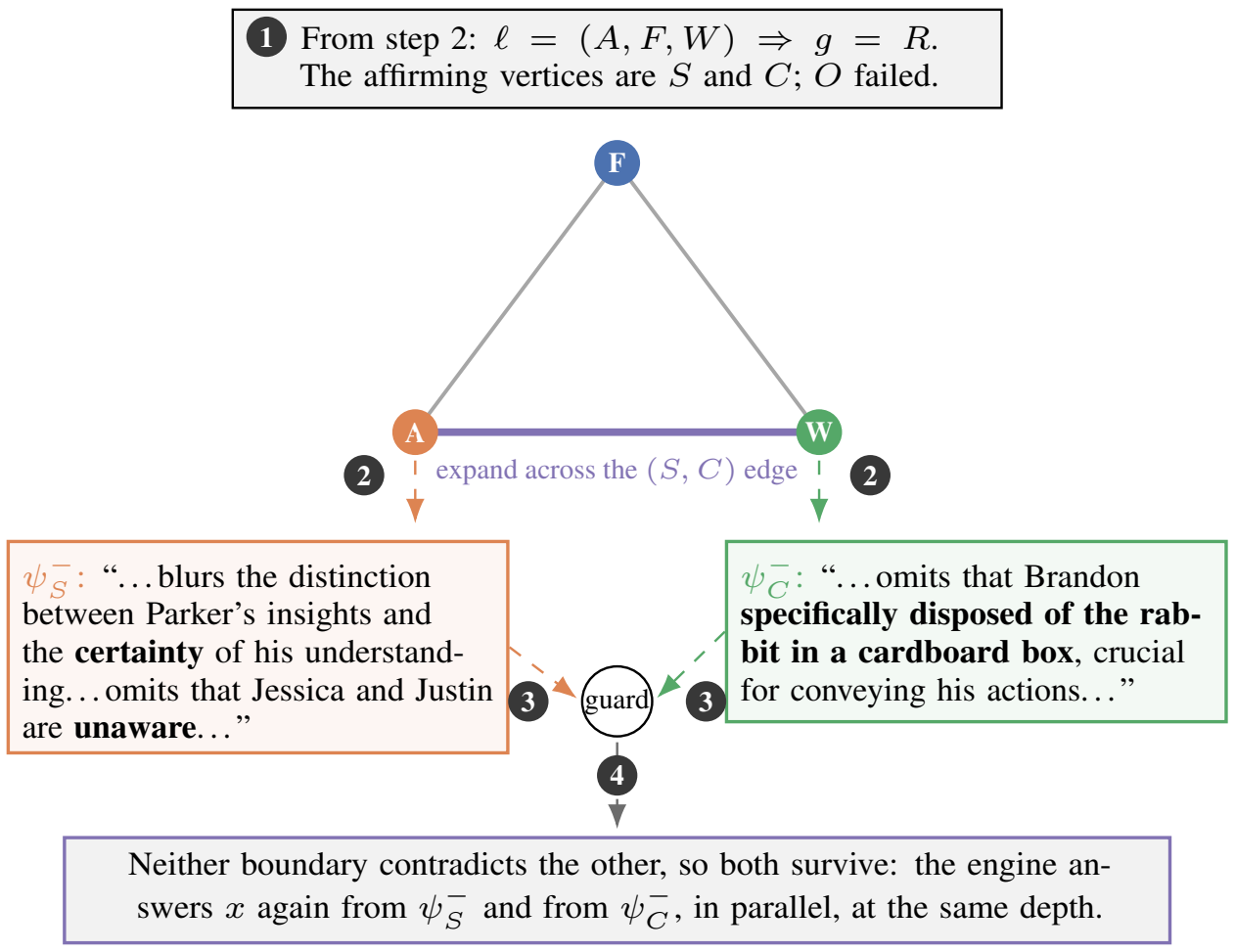
\begin{figure}[t]
\centering
\resizebox{\columnwidth}{!}{%
\begin{tikzpicture}[
  >=Latex, line width=0.5pt,
  box/.style={draw, align=left, inner sep=3pt, font=\scriptsize},
  boxS/.style={box, draw=perspS, fill=perspS!7, line width=0.7pt},
  boxC/.style={box, draw=perspC, fill=perspC!7, line width=0.7pt},
  shaded/.style={box, fill=black!5},
  vdict/.style={circle, fill=#1, text=white, minimum size=0.34cm, inner sep=0pt,
               font=\tiny\bfseries},
  vdict/.default=black!55,
  stepnum/.style={circle, fill=black!78, text=white, minimum size=0.30cm, inner sep=0pt,
               font=\tiny\bfseries}
]
\node[shaded, text width=5.5cm, anchor=south, align=center] (in) at (0,0.40) {%
From step 2: $\ell = (A,F,W) \Rightarrow g=R$. The affirming vertices are $S$ and $C$; $O$ failed.};

\coordinate (O) at (0,0);   \coordinate (S) at (-1.5,-2.0); \coordinate (C) at (1.5,-2.0);
\draw[line width=0.8pt, black!35] (O) -- (S);
\draw[line width=0.8pt, black!35] (O) -- (C);
\draw[line width=1.6pt, routerExp] (S) -- (C);
\node[vdict=perspO] at (O) {F};
\node[vdict=perspS] at (S) {A};
\node[vdict=perspC] at (C) {W};
\node[font=\tiny, text=routerExp, below] at (0,-2.06) {expand across the $(S,C)$ edge};

\draw[dashed, perspS, ->, shorten >=2pt] (S) -- (-1.5,-2.75);
\draw[dashed, perspC, ->, shorten >=2pt] (C) -- (1.5,-2.75);

\node[boxS, text width=3.5cm, anchor=north east] (bS) at (-0.80,-2.80) {%
{\color{perspS}$\psi^-_S$:} ``\dots blurs the distinction between Parker's insights and the
\textbf{certainty} of his understanding\dots omits that Jessica and Justin are
\textbf{unaware}\dots''};
\node[boxC, text width=3.5cm, anchor=north west] (bC) at (0.80,-2.80) {%
{\color{perspC}$\psi^-_C$:} ``\dots omits that Brandon \textbf{specifically disposed of the
rabbit in a cardboard box}, crucial for conveying his actions\dots''};

\node[circle, draw, fill=white, minimum size=0.52cm, inner sep=0pt, font=\tiny, align=center]
  (guard) at (0,-4.00) {guard};
\draw[dashed, perspS, ->, shorten >=1pt] (bS.east) -- (guard.west);
\draw[dashed, perspC, ->, shorten >=1pt] (bC.west) -- (guard.east);
\draw[dashed, boundary, ->, shorten <=1pt] (guard) -- (0,-4.95);

\node[stepnum] at ($(in.north west)+(0.26,-0.21)$) {1};
\node[stepnum] at (-1.88,-2.32) {2};
\node[stepnum] at (1.88,-2.32)  {2};
\node[stepnum] at (-0.66,-4.00) {3};
\node[stepnum] at (0.66,-4.00)  {3};
\node[stepnum] at (0,-4.55)     {4};

\node[shaded, draw=routerExp, line width=0.7pt, text width=8.0cm, anchor=north, align=center]
  (out) at (0,-5.00) {%
Neither boundary contradicts the other, so both survive: the engine answers $x$ again from $\psi^-_S$ and from $\psi^-_C$, in parallel, at the same depth.};
\end{tikzpicture}%
}
\caption{\textbf{Step 3 on the pass judged in \Cref{fig:judging}.} The router takes the verdict triple and $g = R$ from step 2 (1), writes one negation boundary per affirming vertex in parallel (2), checks the two against each other with one guard call (3), and turns each survivor into a text for the engine (4). A negation boundary says what the summary would still have to answer to reach its vertex's opposite. The guard is \textsc{Expand}'s own call asking whether the two boundaries contradict (not one of the loop guards of \Cref{sec:loop}).}
\label{fig:expand}
\end{figure}

The global value says what kind of move to make. It does not yet say what the engine reads next. \Cref{fig:expand} carries the pass of \Cref{fig:judging} through this step. 

\begin{definition}
\label{def:strategy} 
\textit{The Strategy Planner is the function}
\begin{equation*}
  \Delta : \{W,R,A,F\} \times \textstyle\prod_{v \in V}(\mathcal{L}_v \times \mathcal{X}) \times \mathcal{X} \times \mathcal{X} \;\to\; \mathcal{X}^{\leq 2},
\end{equation*}
\textit{that maps the global value $g$, the three verdicts with their rationales $\ell = (\ell_v, \rho_v)_{v \in V}$, the input $x$ and the summary $y_c$ to what the engine must run next. $\mathcal{X}^{\leq 2}$ is the set of at most two texts.}

The four values of $g$ give four moves. Let $n_0$ be the base depth fixed before the run.

\noindent$\bullet$ \textsc{Stop} ($g = W$): the loop ends. The summary $y_c$ has three positive verdicts, so it is the summary the controller returns.

\noindent$\bullet$ \textsc{Expand} ($g = R$): two vertices affirmed the summary, and one did not (\cref{fig:judging}). For each affirming vertex $v^{+}$, one call writes a short statement $\psi^{-}_{v^{+}}$ of what the summary would still have to answer to reach the opposite of that vertex, given the failing judge's reason $\rho_{v^{-}}$. A second call drops one of the two statements if they contradict each other. Both calls go to the meta-judge model. Each statement that survives becomes one text, which asks the engine to answer $x$ again from there. The engine gets one or two texts, at the base depth $n_0$.

\noindent$\bullet$ \textsc{Refine} ($g = A$): one vertex affirmed the summary, and two did not. No call is needed here, because the two failing judges already said what is wrong. Their reasons $\rho_{v^{-}_{1}}, \rho_{v^{-}_{2}}$, written by the meta-judge model, are copied word for word into the two boundaries of the corner at the surviving vertex, $\bigl[v^{+};\, \rho_{v^{-}_{1}},\, \rho_{v^{-}_{2}}\bigr]$. The engine gets one text, at the base depth $n_0$, which asks it to answer $x$ again inside that corner.

\noindent$\bullet$ \textsc{Reset} ($g = F$): no vertex affirmed the summary. The engine gets $x$ itself again, one level deeper, at depth $n_0+1$. This is the only move that changes the depth, and the change does not persist: a later \textsc{Expand} or \textsc{Refine} runs at $n_0$ again. A second reset ends the run.
\end{definition}
There is a duality between \textsc{Refine} and \textsc{Expand}. In \textsc{Refine}, two contextures have failed, so their negative rationales $\bigl(\rho_{v^{-}_{1}}, \rho_{v^{-}_{2}}\bigr)$ already furnish the two boundary walls needed to enclose a sub-triangle around the single surviving vertex $v^{+}$; the failed derivation trace becomes the boundaries of the next call. In \textsc{Expand}, by contrast, two contextures have succeeded, and only one has failed; to move beyond, the router applies determinate negation $\psi^{-}_{v^{+}}$ to the two validated dimensions themselves, generating up to two parallel candidate branches.

\subsection{The Meta-Controller}
\label{sec:loop}

Layer 1 answers. Layer 2 reads that answer and writes the next one the engine has to attempt. Neither closes on its own. The meta-controller runs the two against each other.

One cycle runs the engine on the texts it was given, judges every result, keeps the one with the most positive verdicts (\Cref{def:polyjudge}), reads that triple as a global value (\Cref{def:interpreter}), and writes the next texts (\Cref{def:strategy}). Those texts go back to the engine, and the cycle repeats. We write $T^1, c=4$ for the meta-controller at engine depth $n_0 = 1$ with budget $c_{\max} = 4$; a run may stop earlier through one of the guards below.

Three guards can end the loop early. The \emph{drift} guard is a yes$/$no call, also made by the meta-judge model, that asks whether the new summary still addresses $x$ (see its prompt in \S\ref{sec:prompt-drift}); two consecutive drifted cycles end the run, and a drifted summary is never planned from. The \emph{convergence} guard stops on $g = W$ at any cycle after the first, provided the summary did not drift. The \emph{oscillation} guard stops when a cycle repeats a pair of global value and summary already seen. A budget $c_{\max}$ ends the loops, and the controller must choose which cycle's summary to return. \Cref{alg:controller} shows the entire workflow.

\begin{algorithm}[t]
\caption{The meta-controller.}
\label{alg:controller}
\small
\begin{algorithmic}[1]
\STATE {\bfseries input:} $x$, $c_{\max}$, base depth $n_0$, $V = \{O,S,C\}$
\STATE $X \leftarrow \{x\}$,\; $n \leftarrow n_0$,\; $y_{-1} \leftarrow \varnothing$,\; $\mathcal{S}, \mathcal{C} \leftarrow \emptyset$,\; $d \leftarrow 0$,\; $r \leftarrow 0$
\FOR{$c = 0, \dots, c_{\max}-1$}
  \WHILE{$X \neq \emptyset$}
    \STATE $x_i \leftarrow X.\text{pop}()$;\; $\Phi_i \leftarrow \Phi^{n}(x_i)$ 
    \STATE $y_i \leftarrow T(x_i, \Phi_i)$ \COMMENT{one engine pass}
    \STATE $\ell_i \leftarrow \bigl(J_v(x, y_i, \Phi_i, y_{c-1})\bigr)_{v \in V}$ \COMMENT{step 1}
  \ENDWHILE
  \STATE $i^{*} \leftarrow \arg\max_{i} \bigl( |\mathrm{pos}(\ell_i)|,\ i \bigr)$;\; $(y_c, \ell_c) \leftarrow (y_{i^{*}}, \ell_{i^{*}})$
  \STATE $g_c \leftarrow \variota(\ell_c, c)$ \COMMENT{step 2}
  \STATE $\delta \leftarrow (c \geq 1$ and the drift check rejects $y_c)$
  \IF{$\delta$}
    \STATE $d++$;\; \textbf{break} if $d = 2$ \COMMENT{drift happened}
  \ELSE
    \STATE $d \leftarrow 0$;\; $\mathcal{C} \leftarrow \mathcal{C} \cup \{c\}$;\; $(g^{\star}, \ell^{\star}, y^{\star}) \leftarrow (g_c, \ell_c, y_c)$ \COMMENT{cycle $c$ did not drift}
  \ENDIF
  \IF{$(g_c = W \land c \geq 1 \land \lnot\delta) \lor ((g_c,y_c) \in \mathcal{S})$}
  \STATE \textbf{break} \COMMENT{convergence or oscillation}
  \ENDIF
  \STATE $\mathcal{S} \leftarrow \mathcal{S} \cup \{(g_c,y_c)\}$
  \STATE $X \leftarrow \Delta\bigl(g^{\star}, \ell^{\star}, x, y^{\star}\bigr)$ \COMMENT{step 3}
  \IF{$g^{\star} = F$}
    \STATE $n \leftarrow n_0 + 1$;\; $r++$ \COMMENT{a reset deepens the engine}
  \ELSE
    \STATE $n \leftarrow n_0$ \COMMENT{expand and refine run at the base depth}
  \ENDIF
  \IF{$X = \varnothing \lor r = 2$}
    \STATE \textbf{break} \COMMENT{no move left}
  \ENDIF
\ENDFOR
\STATE $\mathcal{C} \leftarrow \{0, \dots, c\}$ if $\mathcal{C} = \emptyset$ \COMMENT{every cycle drifted}
\STATE $c^{*} = \arg\max_{c \in \mathcal{C}} \bigl( |\mathrm{pos}(\ell_{c})|,\ c \bigr)$
\STATE {\bfseries return} $y_{c^{*}}$
\end{algorithmic}
\end{algorithm}

\section{Experiments}
\label{sec:experiments}

\subsection{Setup and Benchmarks}
\label{sec:setup}

Scoring is an exact match against the gold label for every benchmark with a closed answer set; where string matching fails, or where the benchmark has no closed answer set, a fixed language-model judge ({gpt-4.1-nano} in \cref{tab:main} and {gpt-4o-mini} in \cref{tab:mmlu-pro}) decides correctness (\S\ref{sec:appendix-repro}). The meta-judge, which runs the judges, the \textsc{Expand} calls and the drift check, is gpt-4o-mini in both tables.

We evaluate small, non-frontier answerers (gpt-4.1-nano, Qwen2.5-3b, Llama3.2-3b, gpt-4o-mini, and gemini-2.5-flash-lite; the goal-free condition in \S\ref{sec:goalfree-play} also uses gemini-3.5-flash-lite and gpt-5-mini). Small answerers leave more room for the architecture to change the outcome than frontier models, on which many benchmarks have saturated~\citep{akhtar2026ai}. %

The baseline condition $T^0$ is direct answering. The model receives the input and the question and returns an answer, and no reflection pass runs. Answerers run without a reasoning mode in all conditions, with three exceptions. In the goal-free study, gpt-5-mini runs at medium reasoning effort and gemini-3.5-flash-lite, which cannot switch thinking off, runs with a 512-token thinking budget. gemini-2.5-flash-lite runs at its provider default, which we did not set to zero thinking; on MMLU-Pro it recorded 894, 1{,}738 and 2{,}068 reasoning tokens in total over the 3{,}000-question $T^0$, $T^1$ and $T^1, c=4$ runs. $T^0$ therefore differs from $T^1$ and $T^1, c=4$ both in architecture and in the number of model calls. A $T^1$ pass makes 13 calls before the answering call (three boundaries, three perspectives in each of the three leaves, and the integration), and $T^1, c=4$ adds further passes and the meta-judge's calls. We have no chain-of-thought or compute-matched baseline (\Cref{sec:limitations}).

We test on the following benchmarks.

\textbf{GPQA Diamond}~\citep{rein2023gpqa} (198 graduate-level multiple-choice science questions in physics, chemistry, and biology) tests whether multi-perspective decomposition helps isolate the governing mechanism of a problem and set aside distractors.

\textbf{ExploreToM}\footnote{A static dump of stories generated adversarially against Llama-3.1-70B-Instruct, so a model that shares its failures looks worse and one unlike it looks better. Its intended use is to regenerate the stories against the model under test. We do not.}~\citep{sclar2025explore} (we choose 83 questions over 5 stories) and \textbf{Hi-ToM}\footnote{Hi-ToM has 243 stories, not the documented 240, and 6 of them lack 2--3 of their 5 questions. Sampling 240 at seed 42 drops 3 complete stories and keeps all 6 partial ones, giving $234 \times 5 + 15 = 1{,}185$ scoreable questions.}~\citep{he2023hitom} (1,185 of 1,200 questions over 240 stories, a 15-way choice per question) are narrative false-belief tasks: the answer depends on tracking what a character believes across a multi-paragraph story, with the relevant information scattered rather than local.

From \textbf{BIG-Bench Hard (BBH)}~\citep{suzgun2023challenging} we select five tasks spanning three distinct cognitive regimes: symbolic state tracking via tracking shuffled objects with three, five, and seven objects (T3, T5, T7; 250 questions each, 3-, 5-, and 7-way choice), ordinary-language causal attribution via causal judgment (CJ; 187 questions, Yes$/$No), and exact free-form string manipulation via word sorting (WS; 250 questions, lexicographic sorting).

\textbf{MMLU-Pro}~\citep{wang2024mmlupro} (12,032 questions across 14 academic subjects with three to ten
options per item) tests whether the architecture
generalizes across broad domain knowledge and
multi-step quantitative reasoning.

\subsection{Results}\label{sec:results}

Our aim is to find where the architecture helps and where it hurts, not to maximize benchmark scores.\footnote{\url{https://typesafe.ai/blog/antibenchmaxxing}} All benchmark results are single runs at temperature 0, so most differences carry no confidence interval; we report paired tests where we ran them.

\begin{table*}[!t]
\centering
\caption{\textbf{Accuracy (\%) on the main benchmarks.} The judge is gpt-4.1-nano, except for word sorting with Llama3.2-3b, where the judge is gpt-4o-mini. The meta-judge, used by $T^1, c=4$ only, is gpt-4o-mini. $^{\ddagger}$\,Built by extending the $c=3$ run by one cycle instead of by a cold four-cycle run (\S\ref{sec:appendix-repro}); every other $T^1, c=4$ cell is a cold run. Each cell is a single run; on ExploreToM ($n = 83$) one question is 1.2 points. Llama3.2-3b runs with 4-bit weights on a local runtime, except on the five BIG-Bench Hard tasks (T3--WS), which run through OpenRouter with bf16 weights. Bold is the best performance within the same model family for each dataset. \textit{Total API cost: \underline{\$36.87}, plus \$9.72 for the Llama3.2-3b BIG-Bench Hard rerun.}}
\label{tab:main}
\begin{adjustbox}{max width=\linewidth}
\footnotesize
\begin{tabular}{@{}lrrrrrrrrr@{}}
\toprule
 & \multicolumn{3}{c}{gpt-4.1-nano} & \multicolumn{3}{c}{Qwen2.5-3b} & \multicolumn{3}{c}{Llama3.2-3b} \\
\cmidrule(lr){2-4}\cmidrule(lr){5-7}\cmidrule(lr){8-10}
 & $T^0$ & $T^1$ & $T^1, c=4$ & $T^0$ & $T^1$ & $T^1, c=4$ & $T^0$ & $T^1$ & $T^1, c=4$ \\ \midrule
GPQA & 42.9 & \cellcolor{cellgain!17}46.5 {\scriptsize\textcolor{black!65}{+8.4\%}} & \cellcolor{cellgain!18}\textbf{47.0} {\scriptsize\textcolor{black!65}{+9.6\%}} & 36.9 & \cellcolor{cellloss!19}32.8 {\scriptsize\textcolor{black!65}{$-$11.1\%}} & \cellcolor{cellgain!14}\textbf{38.9} {\scriptsize\textcolor{black!65}{+5.4\%}} & 28.3 & \cellcolor{cellgain!11}28.8 {\scriptsize\textcolor{black!65}{+1.8\%}} & \cellcolor{cellgain!18}\textbf{31.3} {\scriptsize\textcolor{black!65}{+10.7\%}} \\
ExploreToM & 61.4 & 61.4 & \cellcolor{cellgain!19}\textbf{68.7}$^{\ddagger}$ {\scriptsize\textcolor{black!65}{+11.9\%}} & \textbf{66.3} & \cellcolor{cellloss!20}57.8 {\scriptsize\textcolor{black!65}{$-$12.8\%}} & \cellcolor{cellloss!22}55.4 {\scriptsize\textcolor{black!65}{$-$16.4\%}} & \textbf{72.3} & \textbf{72.3} & \cellcolor{cellloss!14}68.7 {\scriptsize\textcolor{black!65}{$-$5.0\%}} \\
Hi-ToM & \textbf{41.4} & \cellcolor{cellloss!13}39.5 {\scriptsize\textcolor{black!65}{$-$4.6\%}} & \textbf{41.4}$^{\ddagger}$ & \textbf{32.2} & \cellcolor{cellloss!19}28.5 {\scriptsize\textcolor{black!65}{$-$11.5\%}} & \cellcolor{cellloss!19}28.6 {\scriptsize\textcolor{black!65}{$-$11.3\%}} & \textbf{33.4} & \cellcolor{cellloss!28}25.3 {\scriptsize\textcolor{black!65}{$-$24.2\%}} & \cellcolor{cellloss!22}28.0 {\scriptsize\textcolor{black!65}{$-$16.2\%}} \\
T3 & 33.6 & \cellcolor{cellgain!45}89.2 {\scriptsize\textcolor{black!65}{$\times$2.7}} & \cellcolor{cellgain!45}\textbf{91.2} {\scriptsize\textcolor{black!65}{$\times$2.7}} & 32.4 & \cellcolor{cellgain!24}39.2 {\scriptsize\textcolor{black!65}{$\times$1.2}} & \cellcolor{cellgain!27}\textbf{42.8} {\scriptsize\textcolor{black!65}{$\times$1.3}} & 30.4 & \cellcolor{cellgain!29}42.0 {\scriptsize\textcolor{black!65}{$\times$1.4}} & \cellcolor{cellgain!32}\textbf{46.0} {\scriptsize\textcolor{black!65}{$\times$1.5}} \\
T5 & 17.6 & \cellcolor{cellgain!55}\textbf{91.6} {\scriptsize\textcolor{black!65}{$\times$5.2}} & \cellcolor{cellgain!55}\textbf{91.6} {\scriptsize\textcolor{black!65}{$\times$5.2}} & 14.0 & \cellcolor{cellgain!34}\textbf{22.4} {\scriptsize\textcolor{black!65}{$\times$1.6}} & \cellcolor{cellgain!32}21.2 {\scriptsize\textcolor{black!65}{$\times$1.5}} & 14.0 & \cellcolor{cellgain!46}\textbf{40.0} {\scriptsize\textcolor{black!65}{$\times$2.9}} & \cellcolor{cellgain!46}\textbf{40.0} {\scriptsize\textcolor{black!65}{$\times$2.9}} \\
T7 & 17.6 & \cellcolor{cellgain!54}85.6 {\scriptsize\textcolor{black!65}{$\times$4.9}} & \cellcolor{cellgain!54}\textbf{87.6}$^{\ddagger}$ {\scriptsize\textcolor{black!65}{$\times$5.0}} & 9.6 & \cellcolor{cellgain!42}\textbf{20.8} {\scriptsize\textcolor{black!65}{$\times$2.2}} & \cellcolor{cellgain!38}18.0 {\scriptsize\textcolor{black!65}{$\times$1.9}} & 11.6 & \cellcolor{cellgain!36}20.4 {\scriptsize\textcolor{black!65}{$\times$1.8}} & \cellcolor{cellgain!41}\textbf{24.4} {\scriptsize\textcolor{black!65}{$\times$2.1}} \\
CJ & 61.5 & \cellcolor{cellloss!12}59.4 {\scriptsize\textcolor{black!65}{$-$3.4\%}} & \cellcolor{cellgain!14}\textbf{64.7}$^{\ddagger}$ {\scriptsize\textcolor{black!65}{+5.2\%}} & 49.2 & \cellcolor{cellgain!15}52.4 {\scriptsize\textcolor{black!65}{+6.5\%}} & \cellcolor{cellgain!21}\textbf{56.1} {\scriptsize\textcolor{black!65}{+14.1\%}} & 52.4 & \cellcolor{cellloss!11}51.9 {\scriptsize\textcolor{black!65}{$-$1.0\%}} & \cellcolor{cellgain!13}\textbf{54.5} {\scriptsize\textcolor{black!65}{+4.1\%}} \\
WS & 46.0 & \cellcolor{cellgain!13}48.0 {\scriptsize\textcolor{black!65}{+4.3\%}} & \cellcolor{cellgain!15}\textbf{48.8}$^{\ddagger}$ {\scriptsize\textcolor{black!65}{+6.1\%}} & \textbf{16.0} & \cellcolor{cellloss!21}13.6 {\scriptsize\textcolor{black!65}{$-$15.0\%}} & \cellcolor{cellloss!16}14.8 {\scriptsize\textcolor{black!65}{$-$7.5\%}} & \textbf{38.8} & \cellcolor{cellloss!32}27.2 {\scriptsize\textcolor{black!65}{$-$29.9\%}} & \cellcolor{cellloss!33}26.8 {\scriptsize\textcolor{black!65}{$-$30.9\%}} \\
\bottomrule
\end{tabular}
\end{adjustbox}
\end{table*}

\begin{table*}[!t]
\centering
\caption{\textbf{Macro-averaged F$_1$, with precision and recall.} The large number is macro-F$_1$, and the small pair is macro-precision$/$macro-recall. Bold is the best macro-F$_1$ within the same model family for each dataset. The one dash is causal judgment with gpt-4.1-nano at $T^0$, where only aggregate counts were kept, so the confusion matrix cannot be recovered. Word sorting is omitted, since each gold string is its own class and the macro-average there equals accuracy. The number of classes is in parentheses. $^{\ddagger}$\,Same runs as the marked cells of \Cref{tab:main}.}
\label{tab:macro}
\begin{adjustbox}{max width=\linewidth}
\footnotesize
\begin{tabular}{@{}lrrrrrrrrr@{}}
\toprule
 & \multicolumn{3}{c}{gpt-4.1-nano} & \multicolumn{3}{c}{Qwen2.5-3b} & \multicolumn{3}{c}{Llama3.2-3b} \\
\cmidrule(lr){2-4}\cmidrule(lr){5-7}\cmidrule(lr){8-10}
 & $T^0$ & $T^1$ & $T^1, c=4$ & $T^0$ & $T^1$ & $T^1, c=4$ & $T^0$ & $T^1$ & $T^1, c=4$ \\ \midrule
GPQA (4) & 42.7 {\scriptsize\textcolor{black!65}{44.7$/$43.3}} & \cellcolor{cellgain!17}46.6 {\scriptsize\textcolor{black!65}{48.5$/$46.9}} & \cellcolor{cellgain!17}\textbf{46.7} {\scriptsize\textcolor{black!65}{48.0$/$46.9}} & 36.6 {\scriptsize\textcolor{black!65}{38.7$/$36.7}} & \cellcolor{cellloss!19}32.4 {\scriptsize\textcolor{black!65}{32.5$/$32.8}} & \cellcolor{cellgain!12}\textbf{37.7} {\scriptsize\textcolor{black!65}{38.0$/$38.0}} & 25.6 {\scriptsize\textcolor{black!65}{27.0$/$27.0}} & \cellcolor{cellgain!15}27.4 {\scriptsize\textcolor{black!65}{28.3$/$28.0}} & \cellcolor{cellgain!21}\textbf{29.3} {\scriptsize\textcolor{black!65}{29.2$/$30.1}} \\
ExploreToM (7) & 54.1 {\scriptsize\textcolor{black!65}{56.8$/$63.3}} & \cellcolor{cellgain!22}\textbf{62.5}$^{\dagger}$ {\scriptsize\textcolor{black!65}{74.7$/$73.6}} & \cellcolor{cellgain!12}55.6$^{\ddagger}$ {\scriptsize\textcolor{black!65}{57.6$/$65.6}} & \textbf{68.7}$^{\dagger}$ {\scriptsize\textcolor{black!65}{75.5$/$76.3}} & \cellcolor{cellloss!30}50.6 {\scriptsize\textcolor{black!65}{60.9$/$52.8}} & \cellcolor{cellloss!35}45.6 {\scriptsize\textcolor{black!65}{52.5$/$54.4}} & 60.5 {\scriptsize\textcolor{black!65}{81.1$/$67.5}} & \cellcolor{cellgain!19}\textbf{67.6} {\scriptsize\textcolor{black!65}{79.9$/$73.1}} & \cellcolor{cellloss!12}59.2 {\scriptsize\textcolor{black!65}{69.0$/$63.7}} \\
Hi-ToM (15) & 32.4 {\scriptsize\textcolor{black!65}{35.1$/$33.7}} & \cellcolor{cellgain!13}33.7 {\scriptsize\textcolor{black!65}{36.9$/$36.1}} & \cellcolor{cellgain!19}\textbf{36.4}$^{\ddagger}$ {\scriptsize\textcolor{black!65}{38.9$/$38.5}} & \textbf{27.9} {\scriptsize\textcolor{black!65}{30.9$/$30.2}} & \cellcolor{cellloss!15}26.1 {\scriptsize\textcolor{black!65}{29.7$/$29.9}} & \cellcolor{cellloss!16}25.7 {\scriptsize\textcolor{black!65}{29.1$/$29.0}} & \textbf{28.5} {\scriptsize\textcolor{black!65}{30.9$/$30.7}} & \cellcolor{cellloss!24}23.3 {\scriptsize\textcolor{black!65}{24.6$/$26.5}} & \cellcolor{cellloss!19}25.2 {\scriptsize\textcolor{black!65}{25.8$/$28.9}} \\
T3 (3) & 29.1 {\scriptsize\textcolor{black!65}{33.6$/$33.8}} & \cellcolor{cellgain!55}89.1 {\scriptsize\textcolor{black!65}{89.3$/$89.0}} & \cellcolor{cellgain!55}\textbf{91.2} {\scriptsize\textcolor{black!65}{91.6$/$91.0}} & 32.2 {\scriptsize\textcolor{black!65}{32.3$/$32.3}} & \cellcolor{cellgain!26}39.1 {\scriptsize\textcolor{black!65}{39.4$/$39.0}} & \cellcolor{cellgain!37}\textbf{43.8}$^{\dagger}$ {\scriptsize\textcolor{black!65}{45.0$/$42.8}} & 29.6 {\scriptsize\textcolor{black!65}{29.6$/$30.0}} & \cellcolor{cellgain!39}41.1 {\scriptsize\textcolor{black!65}{42.0$/$41.8}} & \cellcolor{cellgain!49}\textbf{45.1} {\scriptsize\textcolor{black!65}{46.0$/$45.7}} \\
T5 (5) & 17.1 {\scriptsize\textcolor{black!65}{17.7$/$17.3}} & \cellcolor{cellgain!55}\textbf{91.6} {\scriptsize\textcolor{black!65}{91.6$/$91.7}} & \cellcolor{cellgain!55}91.6 {\scriptsize\textcolor{black!65}{92.3$/$91.6}} & 13.8 {\scriptsize\textcolor{black!65}{15.8$/$13.8}} & \cellcolor{cellgain!54}\textbf{22.0} {\scriptsize\textcolor{black!65}{22.2$/$22.3}} & \cellcolor{cellgain!46}20.6 {\scriptsize\textcolor{black!65}{20.9$/$20.9}} & 13.1 {\scriptsize\textcolor{black!65}{13.2$/$14.5}} & \cellcolor{cellgain!55}39.4 {\scriptsize\textcolor{black!65}{40.8$/$40.8}} & \cellcolor{cellgain!55}\textbf{39.6} {\scriptsize\textcolor{black!65}{41.3$/$40.8}} \\
T7 (7) & 16.6 {\scriptsize\textcolor{black!65}{16.8$/$17.8}} & \cellcolor{cellgain!55}85.5 {\scriptsize\textcolor{black!65}{85.8$/$85.7}} & \cellcolor{cellgain!55}\textbf{87.7}$^{\ddagger}$ {\scriptsize\textcolor{black!65}{88.5$/$87.7}} & 9.1 {\scriptsize\textcolor{black!65}{8.9$/$9.8}} & \cellcolor{cellgain!55}\textbf{20.5} {\scriptsize\textcolor{black!65}{22.5$/$20.8}} & \cellcolor{cellgain!55}17.9 {\scriptsize\textcolor{black!65}{18.9$/$17.9}} & 11.6 {\scriptsize\textcolor{black!65}{11.6$/$11.9}} & \cellcolor{cellgain!55}20.6 {\scriptsize\textcolor{black!65}{23.0$/$20.5}} & \cellcolor{cellgain!55}\textbf{24.4} {\scriptsize\textcolor{black!65}{25.9$/$24.7}} \\
CJ (2) & \multicolumn{1}{c}{--} & 56.9 {\scriptsize\textcolor{black!65}{64.1$/$60.3}} & \textbf{62.5}$^{\ddagger}$ {\scriptsize\textcolor{black!65}{71.4$/$65.7}} & 39.8 {\scriptsize\textcolor{black!65}{52.4$/$50.8}} & \cellcolor{cellgain!17}43.6 {\scriptsize\textcolor{black!65}{62.6$/$54.0}} & \cellcolor{cellgain!28}\textbf{49.2} {\scriptsize\textcolor{black!65}{69.1$/$57.6}} & 43.0 {\scriptsize\textcolor{black!65}{64.4$/$54.0}} & \cellcolor{cellgain!11}43.4 {\scriptsize\textcolor{black!65}{62.6$/$53.4}} & \cellcolor{cellgain!20}\textbf{48.6} {\scriptsize\textcolor{black!65}{63.8$/$55.9}} \\
\bottomrule
\end{tabular}
\end{adjustbox}

{\footnotesize $^{\dagger}$\,Macro-F$_1$ here is at least one point above this cell's accuracy in \Cref{tab:main}, so the condition answers the rare classes better than accuracy suggests. In every other cell, macro-F$_1$ is below accuracy or at most 0.2 points above it; a class the model never gets right contributes a zero to the average.}
\end{table*}
\Cref{tab:main} reports accuracy for one reflection pass ($T^1$) and for the meta-controller at depth 1 with a budget of four cycles ($T^1,c=4$), with the relative change against $T^0$ (direct answering). We test three answerer models: gpt-4.1-nano\footnote{gpt-4.1-nano was deprecated on 24 October 2026, while we were reporting these experiments. We keep the results, although they can no longer be regenerated against the same model.}, Qwen2.5-3b, and Llama3.2-3b. 

\Cref{tab:macro} reads the same predictions by class. A model that leans on two or three answer letters scores well on accuracy while leaving the other classes empty, and the distance between accuracy and macro-F$_1$ measures that lean. On T3, gpt-4.1-nano answers directly at 33.6 accuracy but 29.1 macro-F$_1$. For instance, it names option C twelve times in 250 questions, although C is correct 83 times, so that class recalls 4.8\% and scores 8.4 F$_1$. After one reflection pass, the three options are named 75, 95, and 80 times against a gold split of 79, 88, and 83; no class falls below 85.7 F$_1$, and the gap to accuracy closes to 0.1 points. Llama3.2-3b on T5 behaves the same way. The clearest case is ExploreToM with gpt-4.1-nano, where $T^1$ ties $T^0$ on accuracy at 61.4 while macro-F$_1$ moves from 54.1 to 62.5.

\begin{tcolorbox}[enhanced, sharp corners, frame hidden, borderline west={1.2pt}{0pt}{black!55}, colback=gray!7, left=5pt, right=5pt, top=2pt, bottom=2pt, boxsep=1pt, before skip=6pt, after skip=6pt] The method does not help uniformly. Across the three answerers and eight tasks, $T^1$ or $T^1, c=4$ is above $T^0$ in 30 of 48 cells (from +1.8\% to \texttimes 5.2), below it in 15, and tied in 3. $T^1, c=4$ alone beats $T^0$ in 17 of the 24 model--task pairs. 
\end{tcolorbox}

The largest gains are on the tracking tasks T3--T7, where $T^0$ has to hold the object state without writing it down. Averaged over T3, T5 and T7, the ratio of $T^1, c=4$ to $T^0$ is \texttimes4.3 for gpt-4.1-nano, \texttimes1.6 for Qwen2.5-3b and \texttimes2.2 for Llama3.2-3b. On T7, direct answering must name the final owner after seven swaps in a single step; gpt-4.1-nano scores 17.6\%, close to the chance level of $1/7 = 14.3\%$. In the $T^1$ summary the model first writes out who holds what after the swaps. Against $T^0$, $T^1$ fixes 173 answers and breaks 3 (exact McNemar test, $p < 10^{-4}$).

On causal judgment, the meta-controller's summaries more often keep the agent's intention in view. For example, for the question where \textit{Brown rolls a die hoping to detonate a bomb}, the $T^1$ summary treats the roll as a normal, permissible action and answers \textit{No}, while the meta-steered summary centers on Brown's explicit goal and answers \textit{Yes}. For gpt-4.1-nano, $T^1, c=4$ gains 5.3 points over $T^1$ (17 questions fixed, 7 broken, exact McNemar $p = 0.064$), which is not significant; its true positive rate (TPR, with \textit{Yes} as positive) rises from 34.0\% to 39.2\% and its false positive rate (FPR) falls from 13.3\% to 7.8\%. For the two 3B models (\Cref{tab:cj-binary}), TPR rises from $T^0$ to $T^1, c=4$ in both (Qwen2.5-3b 9.3\% to 18.6\%, Llama3.2-3b 11.3\% to 19.6\%). FPR falls for Qwen2.5-3b (7.8\% to 3.3\%) but rises for Llama3.2-3b (3.3\% to 6.7\%).

\begin{tcolorbox}[enhanced, sharp corners, frame hidden, borderline west={1.2pt}{0pt}{black!55}, colback=gray!7, left=5pt, right=5pt, top=2pt, bottom=2pt, boxsep=1pt, before skip=6pt, after skip=6pt]
\small Hi-ToM is the clearest negative result: a tie for gpt-4.1-nano, and losses for Qwen2.5-3b and Llama3.2-3b.
\end{tcolorbox}
Belief order is how many levels of belief a question nests (see \Cref{fig:hitom-orders}): order 0 asks where an object really is, order 1 where a character thinks it is, and so on up to order 4. At orders 0 and 1, $T^0$ and $T^1$ are indistinguishable. The answer is the object's last position, and the $T^1$ summary is a chronological retelling, the same mechanism as in T3--T7.

\begin{figure}[!t]
\centering
\includegraphics[width=\linewidth]{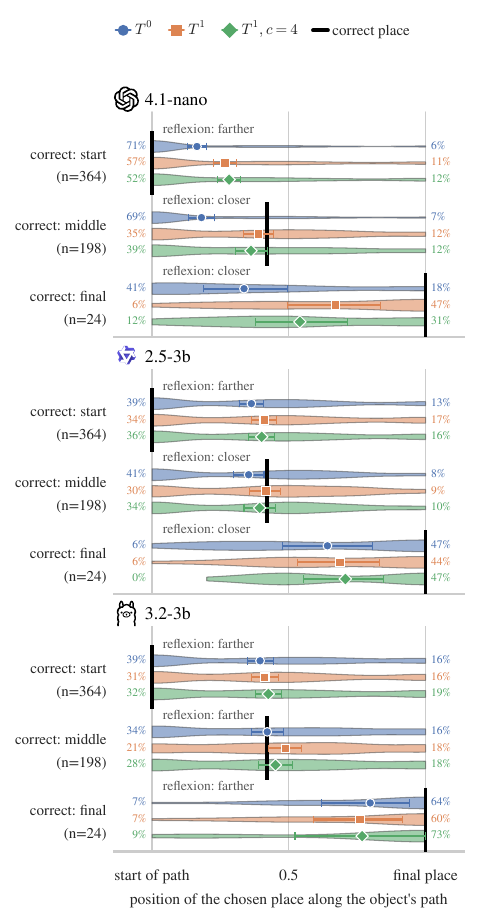}
\caption{\textbf{Where the answer falls along the object's path} (orders 2--4, 586 questions whose correct place is on the path and whose path has at least two moves). 0 is the starting place, and 1 is the final place. Each violin shows how the answer's position is distributed. The marker in each violin is the mean position, with a 95\% bootstrap interval. The black bar marks the average correct place. For gpt-4.1-nano and Qwen2.5-3b, reflection moves the answer later along the path. Llama3.2-3b already answers late under $T^0$, so reflection takes it even farther.}
\label{fig:hitom-path-shift}
\end{figure}

From order 2 onward the retelling works against the question. The correct answer is the starting place in 464 of the 711 questions at orders 2--4, because the characters in the chain left the room before anything was moved. The $T^1$ summary often reports only the move and never says where the object started. There, $T^1$ falls from 53.0\% to 42.9\% for gpt-4.1-nano and from 33.6\% to 23.3\% for Qwen2.5-3b. \Cref{fig:hitom-path-shift} gives the reason. Writing a summary before answering pushes the answer toward the end of the object's path. That helps when the correct place is late on the path and hurts when it is the starting place. Llama3.2-3b already answers late under $T^0$, so the push only moves it farther from the correct place. The meta-controller shifts the answer the same way, so it does not remove the push.

\Cref{tab:mmlu-pro} reports MMLU-Pro for Qwen2.5-3b on all 12,032 questions and for gpt-4o-mini and gemini-2.5-flash-lite on a 3,000-question sample. Here both the judge and the meta-judge are gpt-4o-mini. Pooled over all questions, $T^1, c=4$ improves on $T^0$ for gpt-4o-mini (42.8\% to 47.1\%) and gemini-2.5-flash-lite (53.0\% to 57.2\%), with the largest gains for both in Math, Business, Chemistry and Computer Science, while $T^1$ alone changes little (relative changes of $-0.2\%$ and $+0.9\%$). For Qwen2.5-3b the order reverses: $T^1$ gains slightly (31.7\% to 32.8\%, mostly from Math), and $T^1, c=4$ falls to 28.6\% ($-9.8\%$), below $T^0$ in 13 of the 14 subjects. History and Biology are below $T^0$ for all three models under both conditions. \Cref{tab:mmlu-pro-macro} reports the same runs as macro-F$_1$, and the ordering of the conditions holds; \S\ref{sec:appendix-mmlupro} discusses the subjects in more detail.
\begin{table*}
\centering
\caption{\textbf{Accuracy (\%) on MMLU-Pro.} The judge and, for $T^1, c=4$, the meta-judge are gpt-4o-mini. The column $n$ is the number of questions. To reduce cost, we use a random sample of 3,000 questions, drawn in proportion to the original subject size, for gpt-4o-mini and gemini-2.5-flash-lite. The small number in each cell is the change against $T^0$. Bold is the best performance within the same model family per subject. The \textbf{All} row pools every question rather than averaging the subject scores. \textit{Total API cost: \underline{\$93}}.}
\label{tab:mmlu-pro}
\begin{adjustbox}{max width=\linewidth}
\footnotesize
\begin{tabular}{@{}lrrrrrrrrrrr@{}}
\toprule
 & \multicolumn{4}{c}{Qwen2.5-3b} & \multicolumn{4}{c}{gpt-4o-mini} & \multicolumn{3}{c}{gemini-2.5-flash-lite} \\
\cmidrule(lr){2-5}\cmidrule(lr){6-9}\cmidrule(lr){10-12}
 & $n$ & $T^0$ & $T^1$ & $T^1, c=4$ & $n$ & $T^0$ & $T^1$ & $T^1, c=4$ & $T^0$ & $T^1$ & $T^1, c=4$ \\ \midrule
Math & 1351 & 23.5 & \cellcolor{cellgain!33}\textbf{36.3} {\scriptsize\textcolor{black!65}{+54.5\%}} & \cellcolor{cellloss!17}21.4 {\scriptsize\textcolor{black!65}{$-$8.9\%}} & 337 & 27.0 & \cellcolor{cellgain!10}27.6 {\scriptsize\textcolor{black!65}{+2.2\%}} & \cellcolor{cellgain!34}\textbf{42.7} {\scriptsize\textcolor{black!65}{+58.1\%}} & 39.5 & \cellcolor{cellgain!26}49.3 {\scriptsize\textcolor{black!65}{+24.8\%}} & \cellcolor{cellgain!32}\textbf{58.5} {\scriptsize\textcolor{black!65}{+48.1\%}} \\
Physics & 1299 & 24.9 & \cellcolor{cellgain!18}\textbf{27.2} {\scriptsize\textcolor{black!65}{+9.2\%}} & \cellcolor{cellloss!24}19.9 {\scriptsize\textcolor{black!65}{$-$20.1\%}} & 324 & 25.0 & \cellcolor{cellgain!22}29.3 {\scriptsize\textcolor{black!65}{+17.2\%}} & \cellcolor{cellgain!30}\textbf{34.9} {\scriptsize\textcolor{black!65}{+39.6\%}} & 40.1 & \cellcolor{cellgain!10}41.0 {\scriptsize\textcolor{black!65}{+2.3\%}} & \cellcolor{cellgain!12}\textbf{41.7} {\scriptsize\textcolor{black!65}{+3.8\%}} \\
Chemistry & 1132 & \textbf{25.8} & \cellcolor{cellloss!18}23.1 {\scriptsize\textcolor{black!65}{$-$10.5\%}} & \cellcolor{cellloss!24}20.4 {\scriptsize\textcolor{black!65}{$-$20.9\%}} & 282 & 24.5 & \cellcolor{cellgain!18}27.0 {\scriptsize\textcolor{black!65}{+10.2\%}} & \cellcolor{cellgain!30}\textbf{33.7} {\scriptsize\textcolor{black!65}{+37.6\%}} & 40.1 & \cellcolor{cellloss!13}38.3 {\scriptsize\textcolor{black!65}{$-$4.4\%}} & \cellcolor{cellgain!21}\textbf{45.7} {\scriptsize\textcolor{black!65}{+14.2\%}} \\
Law & 1101 & \textbf{23.0} & \cellcolor{cellloss!21}19.5 {\scriptsize\textcolor{black!65}{$-$15.2\%}} & \cellcolor{cellloss!22}19.3 {\scriptsize\textcolor{black!65}{$-$16.1\%}} & 275 & \textbf{34.5} & \cellcolor{cellloss!20}30.2 {\scriptsize\textcolor{black!65}{$-$12.5\%}} & \cellcolor{cellloss!23}28.0 {\scriptsize\textcolor{black!65}{$-$18.8\%}} & 46.9 & \cellcolor{cellloss!15}44.0 {\scriptsize\textcolor{black!65}{$-$6.2\%}} & \cellcolor{cellgain!9}\textbf{47.6} {\scriptsize\textcolor{black!65}{+1.6\%}} \\
Engineering & 969 & \textbf{32.8} & \textbf{32.8} & \cellcolor{cellloss!22}27.3 {\scriptsize\textcolor{black!65}{$-$16.8\%}} & 242 & 28.1 & \cellcolor{cellgain!9}28.5 {\scriptsize\textcolor{black!65}{+1.4\%}} & \cellcolor{cellgain!11}\textbf{28.9} {\scriptsize\textcolor{black!65}{+2.8\%}} & 48.8 & \cellcolor{cellgain!11}50.0 {\scriptsize\textcolor{black!65}{+2.5\%}} & \cellcolor{cellgain!14}\textbf{51.7} {\scriptsize\textcolor{black!65}{+5.9\%}} \\
Economics & 844 & 43.1 & \cellcolor{cellgain!10}\textbf{44.0} {\scriptsize\textcolor{black!65}{+2.1\%}} & \cellcolor{cellloss!13}41.2 {\scriptsize\textcolor{black!65}{$-$4.4\%}} & 210 & 59.5 & \cellcolor{cellgain!14}\textbf{62.9} {\scriptsize\textcolor{black!65}{+5.7\%}} & \cellcolor{cellgain!14}\textbf{62.9} {\scriptsize\textcolor{black!65}{+5.7\%}} & 68.1 & \cellcolor{cellgain!9}69.0 {\scriptsize\textcolor{black!65}{+1.4\%}} & \cellcolor{cellgain!11}\textbf{70.0} {\scriptsize\textcolor{black!65}{+2.8\%}} \\
Health & 818 & \textbf{36.7} & \cellcolor{cellloss!17}33.6 {\scriptsize\textcolor{black!65}{$-$8.4\%}} & \cellcolor{cellloss!15}34.5 {\scriptsize\textcolor{black!65}{$-$6\%}} & 204 & 62.3 & \cellcolor{cellgain!9}\textbf{63.2} {\scriptsize\textcolor{black!65}{+1.4\%}} & \cellcolor{cellgain!7}62.7 {\scriptsize\textcolor{black!65}{+0.6\%}} & \textbf{62.7} & \cellcolor{cellloss!8}62.3 {\scriptsize\textcolor{black!65}{$-$0.8\%}} & \cellcolor{cellloss!8}62.3 {\scriptsize\textcolor{black!65}{$-$0.8\%}} \\
Psychology & 798 & \textbf{47.7} & \cellcolor{cellloss!15}44.9 {\scriptsize\textcolor{black!65}{$-$5.9\%}} & \cellcolor{cellloss!7}47.4 {\scriptsize\textcolor{black!65}{$-$0.6\%}} & 199 & \textbf{69.8} & \cellcolor{cellloss!12}67.3 {\scriptsize\textcolor{black!65}{$-$3.6\%}} & \textbf{69.8} & 75.9 & \cellcolor{cellgain!9}\textbf{76.9} {\scriptsize\textcolor{black!65}{+1.3\%}} & \cellcolor{cellloss!13}72.9 {\scriptsize\textcolor{black!65}{$-$4\%}} \\
Business & 789 & 26.7 & \cellcolor{cellgain!16}\textbf{28.6} {\scriptsize\textcolor{black!65}{+7.1\%}} & \cellcolor{cellloss!22}22.1 {\scriptsize\textcolor{black!65}{$-$17.2\%}} & 197 & 29.9 & \cellcolor{cellloss!17}27.4 {\scriptsize\textcolor{black!65}{$-$8.4\%}} & \cellcolor{cellgain!30}\textbf{41.1} {\scriptsize\textcolor{black!65}{+37.5\%}} & 42.1 & \cellcolor{cellloss!13}40.1 {\scriptsize\textcolor{black!65}{$-$4.8\%}} & \cellcolor{cellgain!27}\textbf{54.8} {\scriptsize\textcolor{black!65}{+30.1\%}} \\
Biology & 717 & \textbf{56.1} & \cellcolor{cellloss!13}53.4 {\scriptsize\textcolor{black!65}{$-$4.8\%}} & \cellcolor{cellloss!16}51.7 {\scriptsize\textcolor{black!65}{$-$7.8\%}} & 179 & \textbf{77.1} & \cellcolor{cellloss!11}74.9 {\scriptsize\textcolor{black!65}{$-$2.9\%}} & \cellcolor{cellloss!8}76.5 {\scriptsize\textcolor{black!65}{$-$0.8\%}} & \textbf{83.2} & \cellcolor{cellloss!14}78.8 {\scriptsize\textcolor{black!65}{$-$5.4\%}} & \cellcolor{cellloss!14}78.2 {\scriptsize\textcolor{black!65}{$-$6\%}} \\
Philosophy & 499 & 26.3 & \cellcolor{cellgain!21}\textbf{29.9} {\scriptsize\textcolor{black!65}{+13.7\%}} & \cellcolor{cellloss!13}25.1 {\scriptsize\textcolor{black!65}{$-$4.6\%}} & 124 & 48.4 & \cellcolor{cellgain!14}\textbf{50.8} {\scriptsize\textcolor{black!65}{+5\%}} & \cellcolor{cellgain!12}50.0 {\scriptsize\textcolor{black!65}{+3.3\%}} & 51.6 & \cellcolor{cellgain!9}52.4 {\scriptsize\textcolor{black!65}{+1.6\%}} & \cellcolor{cellgain!17}\textbf{55.6} {\scriptsize\textcolor{black!65}{+7.8\%}} \\
Computer Science & 410 & 31.7 & \cellcolor{cellgain!21}\textbf{36.3} {\scriptsize\textcolor{black!65}{+14.5\%}} & \cellcolor{cellloss!19}28.3 {\scriptsize\textcolor{black!65}{$-$10.7\%}} & 102 & 47.1 & 47.1 & \cellcolor{cellgain!24}\textbf{56.9} {\scriptsize\textcolor{black!65}{+20.8\%}} & 55.9 & \cellcolor{cellgain!12}57.8 {\scriptsize\textcolor{black!65}{+3.5\%}} & \cellcolor{cellgain!21}\textbf{63.7} {\scriptsize\textcolor{black!65}{+14\%}} \\
History & 381 & \textbf{29.4} & \cellcolor{cellloss!24}23.1 {\scriptsize\textcolor{black!65}{$-$21.4\%}} & \cellcolor{cellloss!17}26.8 {\scriptsize\textcolor{black!65}{$-$8.8\%}} & 95 & \textbf{56.8} & \cellcolor{cellloss!24}45.3 {\scriptsize\textcolor{black!65}{$-$20.2\%}} & \cellcolor{cellloss!21}48.4 {\scriptsize\textcolor{black!65}{$-$14.8\%}} & \textbf{62.1} & \cellcolor{cellloss!20}53.7 {\scriptsize\textcolor{black!65}{$-$13.6\%}} & \cellcolor{cellloss!21}52.6 {\scriptsize\textcolor{black!65}{$-$15.3\%}} \\
Other & 924 & 30.5 & \cellcolor{cellgain!17}\textbf{33.2} {\scriptsize\textcolor{black!65}{+8.9\%}} & \cellcolor{cellgain!11}31.3 {\scriptsize\textcolor{black!65}{+2.6\%}} & 230 & 56.1 & 56.1 & \cellcolor{cellgain!9}\textbf{57.0} {\scriptsize\textcolor{black!65}{+1.6\%}} & 57.8 & \cellcolor{cellgain!10}59.1 {\scriptsize\textcolor{black!65}{+2.3\%}} & \cellcolor{cellgain!19}\textbf{64.3} {\scriptsize\textcolor{black!65}{+11.3\%}} \\
\midrule
\textbf{All} & 12032 & 31.7 & \cellcolor{cellgain!13}\textbf{32.8} {\scriptsize\textcolor{black!65}{+3.5\%}} & \cellcolor{cellloss!17}28.6 {\scriptsize\textcolor{black!65}{$-$9.8\%}} & 3000 & 42.8 & \cellcolor{cellloss!10}42.7 {\scriptsize\textcolor{black!65}{$-$0.2\%}} & \cellcolor{cellgain!18}\textbf{47.1} {\scriptsize\textcolor{black!65}{+10.0\%}} & 53.0 & \cellcolor{cellgain!11}53.5 {\scriptsize\textcolor{black!65}{+0.9\%}} & \cellcolor{cellgain!16}\textbf{57.2} {\scriptsize\textcolor{black!65}{+7.9\%}} \\
\bottomrule
\end{tabular}
\end{adjustbox}
\end{table*}

\subsection{Ablations}
\label{sec:ablations}

\begin{tcolorbox}[enhanced, sharp corners, frame hidden, borderline west={1.2pt}{0pt}{black!55}, colback=gray!7, left=5pt, right=5pt, top=2pt, bottom=2pt, boxsep=1pt, before skip=6pt, after skip=6pt] More meta-cycles $\centernot\implies$ better performance. 
\end{tcolorbox}

\begin{figure}[t]
\centering
\includegraphics[width=\linewidth]{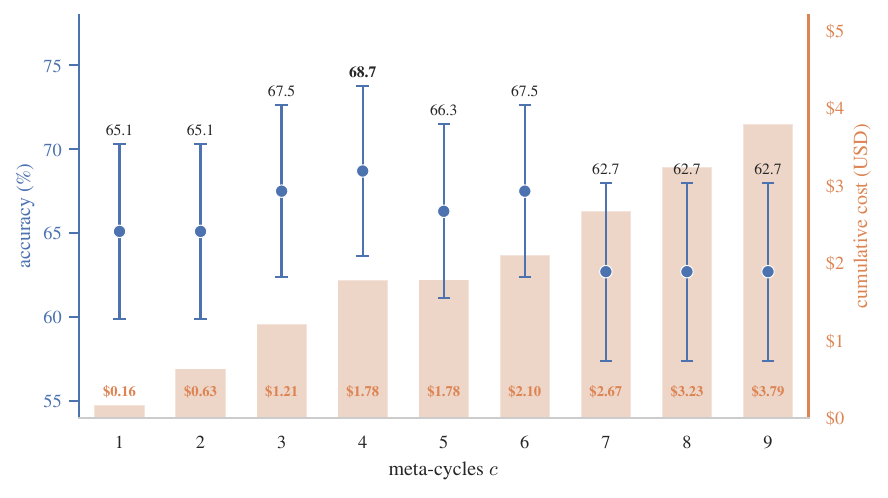}
\caption{\textbf{Meta-cycle budget sweep on ExploreToM.} Accuracy is highest at four cycles (68.7\%) and lowest at seven through nine (62.7\%); the error bars show $\pm 1$ binomial standard error. On the paired comparison that the shared item set allows, four cycles against nine, 8 items flip to correct and 3 to incorrect (McNemar $\chi^2 = 1.45$, $p = 0.23$): the accuracy trend is not significant. The $c=1$ point (65.1\%) is above the $T^1$ entry of \Cref{tab:main} (61.4\%), although a one-cycle budget also runs a single pass: it comes from its own engine pass, whose summaries differ from those of the $T^1$ column (the hosted model is not deterministic at temperature 0); the net difference is 3 of 83 questions. Cumulative cost rises with every cycle, from \$0.16 to \$3.79.}
\label{fig:exploretom-cycles}
\end{figure}

We run budgets of 1--9 meta-cycles with gpt-4.1-nano (meta-judge gpt-4o-mini) on ExploreToM. Instead of rebuilding each run from scratch, the run with budget $c+1$ extends the cached run with budget $c$ by one cycle (\cref{fig:exploretom-cycles}).

The differences across budgets are within noise at $n = 83$ (four against nine cycles, $p = 0.23$). The sweep therefore shows only that cycles beyond four did not help on this set; it does not establish a decline.

\begin{tcolorbox}[enhanced, sharp corners, frame hidden, borderline west={1.2pt}{0pt}{black!55}, colback=gray!7, left=5pt, right=5pt, top=2pt, bottom=2pt, boxsep=1pt, before skip=6pt, after skip=6pt] The three perspectives select different content from the input; they do not paraphrase one another.
\end{tcolorbox}

We test this on the gpt-4.1-nano $T^1$ reflection trees for GPQA, Hi-ToM, T7, and WS. A reflection pass produces three leaves. Each leaf resolves all three perspectives on the same input text under its own region context, which consists of the apex perspective and the two boundary statements adjacent to it (\Cref{def:boundary}); the engine keeps only the apex perspective as the corner answer (\S\ref{sec:engine}). Let \(C_1\) and \(C_2\) denote the sets of content words in two outputs, 
and let \(S\) denote the set of words in the source text. Define the source-constrained content sets as \(A = C_1 \cap S\) and \(B = C_2 \cap S\). The \emph{grounded} divergence between the two outputs is defined as
\[
d(C_1,C_2)
= 1 - \frac{|A \cap B|}{|A \cup B|}\;\text{with } A \cup B \neq \emptyset.
\]
Thus, \(d=0\) indicates that both outputs select the same source content, whereas \(d=1\) indicates that they share no source content. The \emph{raw} divergence uses all content words of the two outputs instead: \(d_\text{raw}(C_1,C_2) = 1 - |C_1 \cap C_2| / |C_1 \cup C_2|\). It also counts words that do not occur in the source text, so it mixes the input content with the wording that the model chose. We report both versions. %

Let: (i) $d_\text{persp}$ compare the three perspectives inside one leaf (same input text and region, different perspectives); (ii) $d_\text{region}$ compare one perspective across the three leaves (same input text and perspective, different region context); and (iii) $d_\text{item}$ compare one perspective across two unrelated stories (different input texts).

\begin{figure}[t]
\centering
\includegraphics[width=\linewidth]{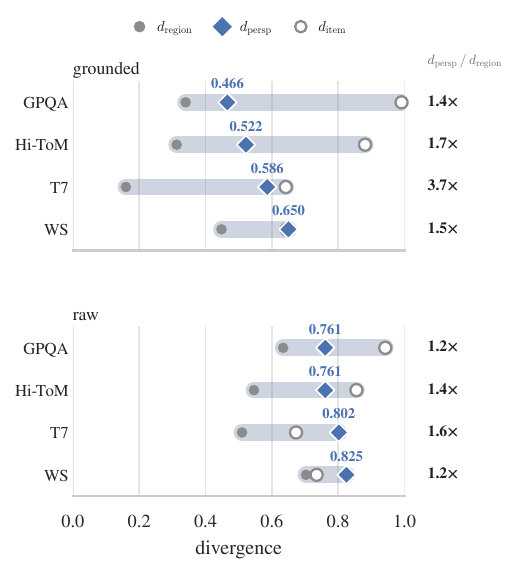}
\caption{\textbf{Perspective divergence against its control ($d_\mathrm{region}$) and an unrelated-input reference ($d_\mathrm{item}$)}, on gpt-4.1-nano $T^1$ trees. The top panel shows the grounded divergence, and the bottom panel the raw divergence. The number above each diamond is $d_\mathrm{persp}$, and the number on the right is $d_\mathrm{persp} / d_\mathrm{region}$. $d_\mathrm{persp} > d_\mathrm{region}$ on every dataset in both panels. The cross-dataset magnitudes differ between the panels, since grounding discards a different share of each output (89\% on WS, 81\% on GPQA).}
\label{fig:perspective-divergence}
\end{figure}

$d_\text{region}$ is the control of this experiment: if $d_\text{persp} \leq d_\text{region}$, asking for a different perspective would change the output no more than varying the surrounding prompt does. \Cref{fig:perspective-divergence} shows $d_\text{persp} > d_\text{region}$ on all four datasets under both versions of the metric. $d_\text{item}$, measured on two unrelated inputs, is a reference point rather than an upper bound: on T7 and WS, $d_\text{persp}$ comes close to it in the grounded version and exceeds it in the raw version. The grounded result indicates that the three perspectives select different input content from the same text, not the same content with different wording. \Cref{fig:perspective-examples} (§\ref{sec:appendix-persp-examples}) shows each of the three comparisons on real output.

\begin{tcolorbox}[enhanced, sharp corners, frame hidden, borderline west={1.2pt}{0pt}{black!55}, colback=gray!7, left=5pt, right=5pt, top=2pt, bottom=2pt, boxsep=1pt, before skip=6pt, after skip=6pt] Divergence does not predict where the meta-controller helps. 
\end{tcolorbox}
That they diverge raises a second question: \textit{Does the degree of divergence on a given input $x$ predict whether the meta-controller beats direct answering on $x$?} We measure $d_\mathrm{persp}$ on the gpt-4.1-nano $T^1$ reflection tree of each input. For each input, we also compute a gain. The gain is $+1$ if $T^1, c=4$ answers correctly and $T^0$ does not, $-1$ in the opposite case, and $0$ if both are right or both wrong. If divergence explained where the method helps, it would correlate positively with this gain. We report the Pearson correlation $r$ with a two-sided $p$-value from the $t$-test on $r$.

On WS ($n = 250$ items), $r = +0.117$ ($p = 0.06$), a weak signal in the predicted direction that disappears for the raw divergence ($r = +0.019$, $p = 0.77$). On T7 ($n = 250$), $r = -0.005$ ($p = 0.94$). $T^1, c=4$ beats $T^0$ so often there (87.6\% against 17.6\%) that 179 of the 250 items have the same gain of $+1$, so the gain barely varies. On Hi-ToM, the questions about one story share one reflection tree, so we correlate per story, with the gain averaged over its questions ($n = 240$ stories): $r = -0.022$ ($p = 0.73$). This value is hard to interpret, because the divergence is read from $T^1$, which itself scores below $T^0$ on Hi-ToM (\Cref{tab:main}). We did not run this analysis on GPQA. We find no evidence that divergence predicts where the meta-controller helps. 

\section{Related Work}

\textbf{Prompting and test-time compute.} Chain-of-thought prompting elicits intermediate steps before an answer~\citep{wei2022chain}, with or without worked examples~\citep{kojima2022large}. Others order the sub-problems~\citep{zhou2022least} or majority-vote over sampled chains~\citep{wang2022self}, where the free variable is diversity, not structure. The compute budget is itself a design choice, allocated by problem difficulty~\citep{snell2024scaling} or spent rewarding correct intermediate steps~\citep{lightman2023lets}. Ours is symmetric, fixed in advance, and uniform across recursion levels.

\textbf{Self-refinement and model-as-judge.} A model can critique and revise its own output~\citep{madaan2023self}, treat a failed attempt as episodic memory~\citep{shinn2023reflexion}, check each step against an environment~\citep{yao2022react}, or self-critique at training time against fixed principles~\citep{bai2022constitutional}. Model-as-judge is standard~\citep{zheng2023judging}, its protocols vary widely~\citep{mao2024gpteval}, and debate extends it to several instances~\citep{du2023improving}. We replace free-form critique with a fixed three-perspective rubric feeding a deterministic state machine, so the verdicts decide whether to continue. Self-correction without external verification often degrades accuracy, and reported gains often trace to answer leakage through the critique prompt~\citep{huang2023large}. Our judges see the reasoning artifacts and the input, never the gold answer.

\textbf{Search- and graph-structured reasoning.} Tree-of-Thoughts searches partial reasoning states with model-driven branch evaluation~\citep{yao2023tree}, and Graph-of-Thoughts lets branches merge~\citep{besta2024graph}. Our recursion is a fixed three-child tree, its branching factor and vertex roles given by construction rather than searched.

\textbf{Neurosymbolic reasoning.} Statistical learning alone lacks the commonsense knowledge and transparent inference that understanding requires, so a line of work couples it with symbolic structure. SenticNet~8 keeps its decisions interpretable through commonsense representation and hierarchical attention~\citep{cambria2024senticnet8}, and~\citet{cambria2024understanding} argues for knowledge representation and neurosymbolic integration on top of it. Evaluations that look past answer accuracy point the same way~\citep{xu2025large}. Our router is symbolic in this sense, with fixed inspectable transitions, but adds no knowledge base: it governs the process, not the content.

\textbf{Theory of mind as a target for machine intelligence.} Humanlike machine intelligence requires common sense in five core domains, one of them intent, as a prerequisite for general intelligence rather than a downstream application~\citep{zhu2020dark}.~\citet{gunther1971cognition} reaches this from logic: a single truth domain can represent a subject's view of objects but not of another subject, so a theory of mind needs several irreducible places from which a statement can be made. Our three perspectives operationalize that argument.~\citet{sclar2025explore} generate false-belief questions programmatically to avoid contamination and surface-pattern shortcuts.

\textbf{Self-referential and self-modifying systems.} Günther treats a machine that models itself as the central problem of cybernetics~\citep{gunther2002bewusstsein}, alongside von Foerster's observing systems, where the observer's description is part of what must be modeled~\citep{vonfoerster1981observing}. Schmidhuber's Gödel machine rewrites its own code whenever it can prove the rewrite increases expected utility, and is optimal precisely because of that proof~\citep{schmidhuber2003godel}. Our controller reorganizes its own completed reasoning, but has no proof system: it acts on judges' verdicts, which can be wrong, so each cycle is an unverified revision that may replace a correct answer with an incorrect one.

\section{Limitations}
\label{sec:limitations}

\textbf{No chain-of-thought or compute-matched baseline.} The only baseline is direct answering. A $T^1$ pass makes 13 answerer calls before the answering call, and $T^1, c=4$ makes several passes plus the meta-judge's calls. Part of the gain may therefore come from the additional computation, or from writing out intermediate state, which a chain-of-thought prompt also elicits.

\textbf{A larger model in the loop.} In every $T^1, c=4$ run, the meta-judge gpt-4o-mini does more than judge. It writes the \textsc{Expand} boundary texts and runs the contradiction and drift checks, and its rationales are copied verbatim into the \textsc{Refine} inputs. For Qwen2.5-3b, Llama3.2-3b, gpt-4.1-nano and gemini-2.5-flash-lite, the $T^1, c=4$ results are therefore those of the answerer steered by a different model, in the 3B cases a larger one. $T^0$ and $T^1$ use the answerer alone.

\textbf{System prompt.} The hosted answerers (gpt-4.1-nano, gpt-4o-mini, gemini-2.5-flash-lite) receive the system prompt ``You are an expert in modeling others' mental states.'' on every task, including GPQA, BBH and MMLU-Pro. Qwen2.5-3b (local) and Llama3.2-3b (local and through OpenRouter) receive no system prompt.

\textbf{Statistics and scoring.} Every cell is a single run at temperature 0, and hosted models are not deterministic at temperature 0. The main tables carry no confidence intervals; the only tests are the paired tests stated in the text. Where string matching fails, a language-model judge decides correctness, and its error rate is not measured.

\textbf{Run construction.} The five gpt-4.1-nano $T^1, c=4$ cells marked in \Cref{tab:main} extend the $c=3$ run by one cycle rather than running four cycles from scratch (\S\ref{sec:appendix-repro}). The Llama3.2-3b BBH cells run with bf16 weights through OpenRouter, while its other cells run with 4-bit weights locally. gemini-2.5-flash-lite was not run with thinking disabled (\S\ref{sec:setup}).

\section{Conclusion}
\label{sec:conclusions}

We formalized polycontextural reflection as a reflection operator that splits a text into a self-similar tree across three perspectives, and a meta-controller that has a language model judge the completed recursion from the same three perspectives and, through a deterministic state machine, decides whether to expand, refine, reset, or stop. Both layers are fixed in advance rather than learned or searched. The largest gains are on state tracking (T3--T7), up to \texttimes 5.2 for gpt-4.1-nano, where the summary writes out the state that direct answering has to hold implicitly. The method loses on higher-order Hi-ToM questions, where the summary pushes answers toward the end of the object's path while the correct answer is usually the starting place. It also loses on word sorting for both 3B models, and on ExploreToM it helps gpt-4.1-nano but hurts both 3B models. On MMLU-Pro, the meta-controller helps gpt-4o-mini and gemini-2.5-flash-lite and hurts Qwen2.5-3b. Because we have no chain-of-thought or compute-matched baseline, these results do not separate the effect of the structure from that of the additional model calls; that comparison is the next experiment to run.

\bibliography{references}
\bibliographystyle{icml2026}

\clearpage
\onecolumn
\appendix

\startcontents[appendices]
\printcontents[appendices]{}{1}{%
  \section*{Appendix Contents}%
  \setcounter{tocdepth}{2}%
}
\clearpage

\section{Polycontextural Grounding}
\label{sec:gunther}
This appendix gives the theoretical argument behind \Cref{sec:method} in full: why the engine decomposes every input across three mutually irreducible standpoints, and why a second-order controller can steer that decomposition through a four-valued router without collapsing the standpoints into a single scalar score.

The argument draws on two connected theoretical results: Günther's theory of \textit{polycontextural logic} \cite{gunther1973life} and the formalization of \textit{co-recursive semantic self-extension} \cite{hegelich2026corecursive}.

\subsection{Discontexturality and the Four-Valued Base}
Classical Aristotelian logic presupposes a single universal locus of valuation from which every proposition is adjudicated as true or false. In Günther’s formulation, a single \textit{contexture} is a domain whose internal structure ``can be exhaustively described by a two-valued logic'' \citep[p.~3]{gunther1973life} and in which ``every logical operation we can perform is confined to the contexturality in which it originates'' \citep[p.~3]{gunther1973life}. As long as a system merely processes data within a fixed representational frame, a single contexture suffices. A reflective cognitive system, however, is simultaneously an actor operating on an object domain and an observer evaluating its own cognitive operations. As \citet[p.~4]{gunther1973life} demonstrates, ``the relation of the cognizing subject to its range of objects is always one of discontexturality,'' making it ``impossible to bring two different contexturalities into an immediate confrontation'' \citep[p.~4]{gunther1973life} as if they shared ``a common ontic measure'' \citep[p.~13]{gunther1973life}.

\begin{table}[!t]
\centering
\caption{\textbf{The four-valued kenogrammatic base ($L_{PKL_2}$).} Coupling an extensional Objective contexture $K_1 \in \{W,F\}$ with an intensional Conceptual$/$Dialectical contexture $K_2\in\{W,R\}$ yields $2^2 = 4$ irreducible logical qualities $\{W,R,A,F\}$.}
\label{tab:truth-basetab}
\resizebox{\textwidth}{!}{%
\begin{tabular}{@{}lllll@{}}
\toprule
$K_1$ (Obj.) & $K_2$ (Dial.) & Kenogram & Symbol & Logical Quality \& Operational Meaning \\ \midrule
$W$ & $W$ & $\langle W, W\rangle$ & $\mathbf{W}$ & \textbf{Robust truth}: Extensionally valid and dialectically substantive \\
$W$ & $R$ & $\langle W, R\rangle$ & $\mathbf{R}$ &\textbf{Formal$/$empty truth}: Factually correct but redundant or abstractly negative \\
$F$ & $W$ & $\langle F, W\rangle$ & $\mathbf{A}$ & \textbf{Trans-classical tension}: Formally blocked in $K_1$, yet conceptually necessary \\
$F$ & $R$ & $\langle F, R\rangle$ & $\mathbf{F}$ & \textbf{Robust falsity}: Both extensionally invalid and dialectically empty \\ \bottomrule
\end{tabular}%
}
\end{table}
Crucially, while each individual contexture remains locally two-valued and governed by its own law of the excluded middle, ``the positive value in contexture A'' is not identical with ``the positive value in contexture B'' \citep[p.~6]{gunther1973life}. Consider the minimal coupling of two elementary contextures ($L_{PKL_2}$): an extensional \textbf{Objective} contexture $K_1$, which evaluates whether a statement is factually and deductively valid ($W$, warranted$/$true) or invalid ($F$, false), and an intensional \textbf{Conceptual}$/$\textbf{Dialectical} contexture $K_2$, which evaluates whether a statement contributes substantive conceptual determination ($W$, dialectically warranted$/$productive) or merely repeats prior content through an empty or abstract negation ($R$, redundant$/$irrelevant). Because Gödelian formalization inside a single contexture ``concerns only the extensional domain of logic and bypasses those intensional relations where dialectical principles come into play'' \citep[p.~10]{gunther1973life}, a statement’s status cannot be captured by a single truth value; it is given by the relational pair (or \textit{kenogram}) $v_{PKL_2}(s) = \langle v_1(s), v_2(s)\rangle\in \{W, F\} \times \{W, R\}$.

As shown in \Cref{tab:truth-basetab}, pairing $K_1$ and $K_2$ immediately generates four distinct logical qualities, which provide the four foundational symbols $\{W,R,A,F\}$ used throughout our architecture:

(1) \textbf{Robust truth} ($\langle W, W \rangle \equiv W$): A derivation that is extensionally valid in $K_1$ and substantively productive in $K_2$.

(2) \textbf{Formal, empty truth} ($\langle W, R \rangle \equiv R$): A statement that is extensionally unobjectionable in $K_1$ (such as the trivial negation ``a lion is not a table'' or a verbatim paraphrase of a prior cycle) yet fails in $K_2$ because it adds no determinate conceptual content. It indicates that reasoning has reached the boundary of its current semantic space without advancing it.

(3) \textbf{Trans-classical / authentic tension} ($\langle F, W\rangle \equiv A$): A statement that violates extensional constraints in $K_1$ or remains blocked at a formal impasse, yet articulates a coherent, necessary conceptual insight in $K_2$ (just as in Hegel’s \textit{Science of Logic} pure Being and pure Nothing are extensionally opposed yet conceptually inseparable; \citealp[p.~33]{hegel2010georg}). It marks a productive structural tension that requires localized refinement rather than dismissal.

(4) \textbf{Robust falsity} ($\langle F, R \rangle \equiv F$): A statement that is both extensionally false and dialectically empty (such as a self-refuting linguistic error or a hallucinated premise), warranting rejection and reset.

\subsection{Mediating Three Contextures and Second-Order Reduction}
A two-contexture split between objective fact ($K_1$) and conceptual reflection ($K_2$) still leaves out the concrete standpoint of the experiencing or reasoning subject. More fundamentally, as \citet[p.~15]{gunther1973life} argues, ``there can be no such thing as linking two elementary contextures into a compound contexture, for this would require a minimum of three contextures. One of the three would have to mediate between the other two.'' To form a closed compound system ($L_{PKL_3}$), we therefore introduce a third elementary contexture $K_3$: the domain of \textbf{Subjective authenticity} ($\mathcal{L}_S = \{A, F\}$), which evaluates whether an argument is internally coherent and authentic to its own perspectival assumptions ($A$) or internally fractured ($F$), independent of its external factual truth.

At the object level of our engine (\S\ref{sec:engine}), these three contextures are instantiated as three symmetric perspective projections on a piece of text: \textbf{O}bjective ($\varphi_O$), \textbf{S}ubjective ($\varphi_S$), and \textbf{C}onceptual ($\varphi_C$). None of the three is privileged as ground truth. Instead, each pair of perspectives articulates an explicit boundary ($\psi_{OS}, \psi_{OC}, \psi_{SC}$) characterizing where the two contextures diverge.

When the second-order meta-controller (\S\ref{sec:meta}) evaluates a completed reflexion pass across all three binary contextures, it observes a valuation triple $\ell = (\ell_S, \ell_O, \ell_C) \in \mathcal{L}_S \times \mathcal{L}_O \times \mathcal{L}_C = \{A,F\}\times\{W,F\}\times\{W,R\}$, spanning $2^3 = 8$ possible kenogrammatic states. Rather than letting the state space grow exponentially with the number of contextures, a second-order observer $K_M$ \citep[p.~15]{gunther1973life} asks a structural question about the judgment triple itself: \textit{how many of the three contextures affirm the current reasoning state?} The router reduces the eight triples to the four symbols $\mathcal{G} = \{W, R, A, F\}$ of \Cref{tab:truth-basetab} by counting positive verdicts: three positives $\mapsto W$, two $\mapsto R$, one $\mapsto A$, none $\mapsto F$. This count rule is a design heuristic, not a projection that preserves the meaning of the base table. In the base table an objective failure always yields $A$ or $F$, whereas under the count rule the triple $(A,F,W)$, which fails objectively, maps to $R$. The four symbols are reused as labels for the four count levels, and the router's actions are defined on those levels.

\subsection{Co-Recursion and Trace Reification}
\textit{Why should translating the three-contexture triple back into $\{W, R, A, F\}$ improve reasoning rather than merely relabel errors?} Our motivation comes from the distinction between local contraction and global co-recursion drawn by \citet{hegelich2026corecursive}.

Within a fixed representational regime $\mathcal{F}_n = (L_n, \Gamma_n, \vdash_{n})$ and state space $(X_n, d_n)$, the set of unresolved problems $\mathcal{U}(O_n)$ decomposes into two disjoint regions \citep[p.~13]{hegelich2026corecursive}:
\begin{equation}
    \mathcal{U}(O_{n}) = \mathcal{U}^{\mathrm{loc}}(O_{n}) \;\dot\cup\; \mathcal{U}^{\mathrm{inc}}(O_{n}),
\end{equation}%
where $\mathcal{U}^{\mathrm{loc}}(O_{n})$ contains propositions decidable by local inference within the current frame and $\mathcal{U}^{\mathrm{inc}}(O_n)$ contains propositions that cannot be settled without changing the inferential frame itself. (Here $n$ indexes the regime, following the cited work, not the engine depth of \S\ref{sec:engine}.) As an analogy, and not as a claim we prove about our engine, a reflexion pass can be pictured as a locally contractive map on $X_n$: applying the same pass repeatedly to its own output tends to settle on a local fixed region $\Lambda_n \subseteq X_n$ rather than reach new content. Proposition 3 (\textit{Escape from Local Convergence}) of \citet[p.~20]{hegelich2026corecursive} makes the corresponding statement for its formal setting: iterating local transformations within a fixed regime exhausts $\mathcal{U}^{\mathrm{loc}}(O_{n})$ at a rate $\kappa_n^t$ but leaves $\mathcal{U}^{\mathrm{inc}}(O_{n})$ invariant. Read in our terms, re-running the same reflexion pass on its own summary, without a meta-level change of input, can be expected to produce redundant restatements or compounding copy errors rather than new determinations.

To escape $\Lambda_n$, the architecture requires a co-recursive transition $X_n \to X_{n+1}$ driven by an a priori meta-teleological trigger $\mathcal{M}_n$ \citep[p.~19]{hegelich2026corecursive}. In Hegel’s analysis of dialectical method, ``the real issue is not exhausted by stating it as an aim, but by carrying it out, nor is the result the actual whole, but rather the result together with the process through which it came about'' \citep[p.~2]{hegel1807phenomenology}. When an object-level derivation stalls or fails, that impasse does not resolve ``into abstract nothingness, but essentially only into the negation of its particular content'' \citep[p.~33]{hegel2010georg}; it is a determinate negation in which ``a new form has thereby immediately arisen, and in the negation the transition is made'' \citep[p.~51]{hegel1807phenomenology}. Because ``content is at the same time the negative reflection of the form determinations into themselves'' \citep[p.~396]{hegel2010georg}, reflecting on an incomplete derivation trace converts its failed operational form in one cycle into determinate propositional content (explicit boundary constraints) for the next \citep[p.~15]{hegelich2026corecursive}.

Our polycontextural logic router is designed after this co-recursive trace reification. Schmidhuber's Gödel machine \citep{schmidhuber2007godel} rewrites its state only when a deductive proof searcher certifies utility improvement within the initial axiom system. That requirement collapses ``all possible \textit{meta-levels} of the self-referential system [\dots] into one'' and forces the machine to ``simply ignore'' any modification whose utility is undecidable in the base regime. Our router acts where that object-level certainty fails. By Proposition 1 (\textit{Existence of Non-Trivial Trace-Reduced Decisive Extensions}) and Proposition 2 (\textit{Trace-Guided Candidate Generation and Fallibility}) of \citet[pp.~16--17]{hegelich2026corecursive}, reifying the open blocking frontier of an incomplete trace into negated boundary extensions generates a non-trivial semantic surplus of at least two candidate extensions ($|\text{Surp}(O_n)| \geq 2$). Our geometry map (\Cref{def:geometry}) records only one unused alternative per cycle, so it keeps a single element of that surplus. Because no finite system can decide in advance whether an open subgoal lies in $\mathcal{U}^\mathrm{loc}(O_n)$ or $\mathcal{U}^\mathrm{inc}(O_n)$, every co-recursive step is fallible exploration rather than deductively guaranteed improvement.

\subsection{The Router, Triple by Triple}
\label{sec:router-table}
\Cref{def:interpreter} reads the global value off $k$, the number of judges that affirmed the summary, and \Cref{def:strategy} turns that value into the next cycle's input. Between them they fix what the router does on every one of the $2^3 = 8$ valuation triples, since $\mathrm{pos}(\ell)$ names which edge an \textsc{Expand} crosses and which corner a \textsc{Refine} descends into. \Cref{tab:truth-router} writes all eight out, so the cases can be read off directly rather than reconstructed.

\begin{table*}[t]
\centering
\caption{\textbf{Complete state table of the polycontextural logic router.} The three locally binary judging contextures $\mathcal{L}_{S} \times \mathcal{L}_{O} \times \mathcal{L}_{C} = \{A,F\} \times \{W,F\} \times \{W,R\}$ span $2^3 = 8$ valuation triples $\ell$. The Interpreter $\variota$ maps them onto the four global values $\{W,R,A,F\}$ through the number of positive verdicts $k = |\mathrm{pos}(\ell)|$ alone, and $\Delta$ executes the corresponding transition. The geometry map also records one unused alternative move per row (\Cref{def:geometry}); it is not shown here.}
\label{tab:truth-router}
\resizebox{\textwidth}{!}{%
\begin{tabular}{ccc l l l}
\toprule
\multicolumn{3}{c}{\textbf{Judge Triple $\boldsymbol{\ell}$}} &
\textbf{Positive Verdicts} & \textbf{Global Value} &
\textbf{Router Action $\boldsymbol{\Delta(g,\ell)}$ and Boundary Construction} \\
\cmidrule(lr){1-3}
$\ell_{S}$ & $\ell_{O}$ & $\ell_{C}$ &
$k$ (Vertices) & $\variota(\ell,c)$ & (Transition to cycle $c+1$) \\
\midrule
$A$ & $W$ & $W$ & $3\,\{O,S,C\}$ & $\mathbf{W}$ &
\textbf{Stop:} terminate (for $c \geq 1$; at $c{=}0$ flipped to $(A,W,R) \mapsto \mathbf{R}$). \\
\midrule
$A$ & $W$ & $R$ & $2\,\{O,S\}$ & $\mathbf{R}$ &
\textbf{Expand along $(O,S)$:} derive negation boundaries $\psi^{-}_{S}, \psi^{-}_{O}$ against the $C$-redundancy. \\
$A$ & $F$ & $W$ & $2\,\{S,C\}$ & $\mathbf{R}$ &
\textbf{Expand along $(S,C)$:} derive $\psi^{-}_{S}, \psi^{-}_{C}$ against the $O$-falsity. \\
$F$ & $W$ & $W$ & $2\,\{O,C\}$ & $\mathbf{R}$ &
\textbf{Expand along $(O,C)$:} derive $\psi^{-}_{O}, \psi^{-}_{C}$ against the $S$-incoherence. \\
\midrule
$A$ & $F$ & $R$ & $1\,\{S\}$ & $\mathbf{A}$ &
\textbf{Refine into corner $S$:} reuse the negative rationales $(\rho_{O}, \rho_{C})$ verbatim as the two boundaries. \\
$F$ & $W$ & $R$ & $1\,\{O\}$ & $\mathbf{A}$ &
\textbf{Refine into corner $O$:} reuse $(\rho_{S}, \rho_{C})$ verbatim. \\
$F$ & $F$ & $W$ & $1\,\{C\}$ & $\mathbf{A}$ &
\textbf{Refine into corner $C$:} reuse $(\rho_{S}, \rho_{O})$ verbatim. \\
\midrule
$F$ & $F$ & $R$ & $0\,\emptyset$ & $\mathbf{F}$ &
\textbf{Reset:} re-run the raw input $x$ once at depth $n_0{+}1$; a second $F$ terminates (\texttt{exhausted\_F}). \\
\bottomrule
\end{tabular}%
}
\end{table*}

\subsection{Geometry Mapping (\texorpdfstring{$\Gamma$}{Gamma}) and Semantic Surplus}
Let $\mathcal{V}_{\text{ord}} = (O, S, C)$ be the canonical cyclic vertex order and $\pi : V \to V$ the cyclic shift $O \mapsto S \mapsto C \mapsto O$.

\begin{definition}[Geometry map and Semantic Surplus]
\label{def:geometry}
Let $\mathcal{A}$ be the set of geometric arrangements on the Sierpinski triangle. The \emph{geometry map} is a function $\Gamma : \mathcal{G} \times 2^V \to \mathcal{A} \times \mathcal{A}$, which maps $(g, \text{pos}(\ell))$ to a pair $\Gamma(g, \text{pos}(\ell)) = (a_g, \bar a_g)$, consisting of a canonical move $a_g$ and a preserved unused alternative $\bar a_g$ (semantic surplus):

\noindent (1) For $g = W$ ($\text{pos}(\ell) = \{O, S, C\}$): $a_W = \text{\texttt{\textbf{rest}}}()$ and $\bar a_W = \text{\texttt{\textbf{expand\_full}}}(O, S, C)$.

\noindent (2) For $g = R$ ($\text{pos}(\ell) = \{u, v\}$):
$a_R = \text{\texttt{\textbf{expand\_edge}}}(u, v)$ and $\bar a_R = \text{\texttt{\textbf{expand\_edge}}}(\pi(u), \pi(v))$.

\noindent (3) For $g = A$ ($\text{pos}(\ell) = \{v^*\}$ with negative vertices $(u_1, u_2)$ ordered by $\mathcal{V}_\text{ord}$): $a_A = \text{\texttt{\textbf{descend\_corner}}}(v^*)$ and $\bar a_A = \text{\texttt{\textbf{descend\_corner}}}(u_1)$.

\noindent (4) For $g = F$ ($\text{pos}(\ell) = \emptyset$): $a_F = \text{\texttt{\textbf{reset\_root}}}()$ and
$\bar a_F = \text{\texttt{\textbf{invert\_root}}}(O, S, C)$.
\end{definition}

\paragraph{Does the Unused Surplus Do Better?}

\begin{tcolorbox}[enhanced, sharp corners, frame hidden, borderline west={1.2pt}{0pt}{black!55}, colback=gray!7, left=5pt, right=5pt, top=2pt, bottom=2pt, boxsep=1pt, before skip=6pt, after skip=6pt] In a small test, following the unused surplus did not beat the controller's own choice.
\end{tcolorbox}
Every meta-controller cycle records the move it takes and one alternative it does not take (\Cref{def:geometry}). For a \textsc{Refine} cycle, where the judges approve exactly one of the three perspectives, the move descends into that perspective's corner and the recorded alternative is one of the two rejected perspectives, which the controller never uses. \textit{Would the controller have done better had it followed the unused perspective?} For every \textsc{Refine} cycle, we use the surplus perspective instead of the chosen one. We run one engine call and one judge call on this branch. Then we compare how many perspectives the judges approve on this branch with the count of the real path's next cycle. On 40 Hi-ToM stories, 45 cycles reached the \textsc{Refine} state, and 26 of them had a next cycle to compare against. The other 19 were the last step of the four-cycle budget. The surplus branch ties the real path in 21 of the 26 cycles, wins in 3, and loses in 2. On this small sample the unused perspective shows no sign of being a missed improvement, and 26 comparisons are too few to say more. One reason to expect ties is that the surplus perspective derives from the same three-judge rationales as the chosen one.

\subsection{Selection Under the Cycle-0 Rule}
The selection rule of \Cref{alg:controller} interacts with the cycle-0 reinterpretation rule. Because an initial $(A,W,W)$ at $c = 0$ is reinterpreted to $(A,W,R)$, its stored positive count drops from 3 to 2. Any later on-topic cycle with at least two positive verdicts therefore ties or beats cycle~0 on $|\mathrm{pos}(\ell_{c})|$ and wins the tie-break, which goes to the later cycle. The answer falls back to $y_{0}$ only if every later cycle drops to one or no positive verdicts.

\section{Reproducibility}
\label{sec:appendix-repro}

\paragraph{Code and configuration} Every run is launched through one
command-line entry point against a Hydra configuration tree. A condition is
one cell of that tree: a backend (which model serves which role), a reflexion
depth, and a meta-cycle budget. Reflexion summaries are cached per item, under
a key derived from the condition and inside a \emph{cache namespace} chosen
for the run. The cache key does not include the prompt text, so a run that
changes a prompt must use a new namespace; otherwise it reuses summaries built
with the old prompt. A run's results can also depend on what its namespace
already holds, through the chained continuation described below. A run is
therefore specified by its overrides together with the state of its cache
namespace, and each published run file records the namespace it used.

\paragraph{Provenance} Each run writes the resolved configuration, the model
and temperature per role, package versions, a hash of every prompt template,
and the git state alongside its results. The git record is incomplete. Of the
206 run files we release, 156 name a commit but mark the working tree as
dirty, so the commit does not fully describe the code that ran; 44 record no
commit at all; and 6, the Llama3.2-3b BIG-Bench Hard rerun, name a commit with
a clean tree.

\paragraph{Chained continuation of meta-cycle budgets} Five
$T^1, c{=}4$ cells of \Cref{tab:main} were not produced by running
\Cref{alg:controller} for four cycles: gpt-4.1-nano on ExploreToM, Hi-ToM, T7,
causal judgment and word sorting. In those runs the budgets $c=1$, $2$ and $3$
were evaluated in the same grid as $c=4$, and the runner builds a budget
$c \geq 2$ by continuing from the final summary of budget $c-1$ whenever that
summary is already in the cache namespace. A continuation step judges the
seed summary again, without corner artifacts (none are stored with a summary)
and without a previous summary. A $W$ on the seed is treated as $R$ when
planning, so the step cannot stop before it has run. It then runs one
new cycle and returns whichever of the seed and the new cycle has more
positive verdicts, the new cycle on a tie. Each of these five $c=4$ cells is
therefore the end of a chain $c=1 \to 2 \to 3 \to 4$ that adds one cycle per
step. All other $c=4$ cells (gpt-4.1-nano on GPQA, T3 and T5; every
Qwen2.5-3b and Llama3.2-3b cell; all of \Cref{tab:mmlu-pro}) are cold
four-cycle runs. The 1--9 cycle sweep on ExploreToM
(\Cref{fig:exploretom-cycles}) uses the same chaining.

\paragraph{Decoding and sampling} Every role is requested at temperature $0$,
and each run file records the temperature per role. There is one exception
in the client code: when the Ollama server aborts a reply for repetition, the
client retries that request at temperature $0.3$ with a repeat penalty of
$1.15$. The run files do not record how often this happened. Answerers served
by Ollama generate at most 512 tokens per call (\texttt{num\_predict}) with a
context window of 8{,}192 tokens (\texttt{num\_ctx}); the Llama3.2-3b runs
through OpenRouter are capped at 512 tokens to match. Hosted models run
without an explicit output cap. gemini-2.5-flash-lite ran with the provider's
default thinking setting; no thinking budget was set. Its MMLU-Pro run records
894 reasoning tokens over 3{,}000 $T^0$ answering calls, 1{,}738 over
3{,}000 $T^1$ answering calls, and 2{,}068 over 129{,}218 answerer calls in
the $T^1, c{=}4$ condition, against 28.9 million completion tokens in that
condition. Item sampling uses a seed of $42$ throughout. Benchmarks whose
items share a context (the narrative tasks) are sampled by context, so all
questions belonging to a story enter or leave the sample together and share
one reflexion tree.

\paragraph{System prompts} Hosted answerers (gpt-4.1-nano, gpt-4o-mini,
gemini-2.5-flash-lite) send the system prompt ``You are an expert in modeling
others' mental states.''\ with every call, on every benchmark. Answerers
served by Ollama or OpenRouter send no system prompt. The judge and the
meta-judge have system prompts of their own (\Cref{sec:prompt-configs}).

\paragraph{Determinism} Temperature $0$ does not make a run repeatable. Hosted
APIs can return different completions for identical requests at temperature
$0$. The engine also issues calls concurrently, and the meta-controller's
stopping rule depends on judge verdicts that are themselves model outputs, so
two runs of the same condition can terminate at different cycles on the same
item. Accuracies reported here are from one run per condition; we do not
report run-to-run variance, and differences of a few points should be read
with that in mind (\Cref{fig:exploretom-cycles}).

\paragraph{Benchmarks} GPQA Diamond contributes 198 questions and is a gated
dataset, requiring an access token to download. Its licence does not allow
republishing the questions, so its per-question run files are not released;
their provenance records temperature $0$ for every role, with gpt-4.1-nano
served by OpenAI and Qwen2.5-3b and Llama3.2-3b served by Ollama. ExploreToM
contributes 83 questions over 5 stories, read from the published static file
rather than regenerated against the model under test. Hi-ToM contributes
1{,}185 of 1{,}200 questions over 240 stories. The released data has 243
stories rather than the documented 240, and 6 of them lack 2--3 of their 5
questions; sampling 240 stories at seed 42 drops 3 complete stories and keeps
all 6 partial ones, giving $234 \times 5 + 15 = 1{,}185$ questions. From
BIG-Bench Hard we take tracking shuffled objects at three, five and seven
objects (250 questions each), causal judgment (187) and word sorting (250).
MMLU-Pro (\Cref{tab:mmlu-pro}) is the TIGER-Lab test split of 12{,}032
questions. Qwen2.5-3b answers all of them. gpt-4o-mini and
gemini-2.5-flash-lite answer a fixed sample of 3{,}000, drawn in proportion to
subject size and identical for both models
(\texttt{data/subsets/mmlu\_pro\_stratified3000.json}).

\paragraph{Grading} Answers are scored against the gold label by exact match
where the reply lands on an allowed label, and by a judge model otherwise
(\Cref{sec:prompt-grader}). In \Cref{tab:main} the judge is gpt-4.1-nano,
except for Llama3.2-3b on word sorting, where it is gpt-4o-mini, and the
meta-judge is gpt-4o-mini in every cell. In \Cref{tab:mmlu-pro} both roles
are gpt-4o-mini. Within each table the grader is the same for every answerer,
apart from that one word-sorting cell, so a difference between answerers is
not a difference in grading.

\paragraph{What the meta-judge writes} The meta-judge does more than judge.
The same model client writes the \textsc{Expand} negation boundaries, runs the
contradiction guard and the drift check, and the \textsc{Expand} and
\textsc{Refine} inputs quote its rationales verbatim; \textsc{Refine} uses two
of them as the boundaries of the next region. When the answerer is smaller
than gpt-4o-mini, text written by the larger model therefore enters every
cycle after the first. The $c \geq 1$ conditions measure the answerer working
with that text, not the answerer alone.

\paragraph{Cost accounting} Token counts are recorded per call and priced with
a maintained price table. Locally served open-weights models carry no price.
The dollar figures for runs with a local answerer therefore cover only the
hosted judge and meta-judge calls, and are not comparable to the figures for
hosted answerers; the run files also report token and call counts for the
local model.

\paragraph{Hardware} Experiments whose answerer is a hosted API are
network-bound: the client machine issues concurrent requests and waits, so its
hardware does not affect accuracy, and affects wall-clock time only through the
number of requests it keeps in flight. Runs needing more concurrency than a
workstation sustains were executed on a Google Cloud Compute Engine
{e2-highmem-8} instance (8 vCPU, 64\,GB RAM) in {europe-west3}.
Experiments whose answerer is an open-weights model are not
hardware-independent. {qwen2.5:3b} was served by Ollama on a Compute Engine
{n1-standard-4} instance (4 vCPU, 15\,GB RAM) with one NVIDIA Tesla T4
(16\,GB) in {europe-west3}, resident entirely in GPU memory with four requests
served in parallel. The local {llama3.2:3b} runs were also served by Ollama on
Compute Engine; their instance type is not recorded in the run files. Both
models ran with Ollama's default 4-bit quantized weights. The Llama3.2-3b
cells on the five BIG-Bench Hard tasks were instead run through OpenRouter,
whose only provider for this model (Parasail) serves bf16 weights. No model was
fine-tuned or otherwise modified.

\section{Prompts}
\label{sec:appendix-prompts}

This appendix prints every prompt template behind \Cref{tab:main} and \Cref{tab:mmlu-pro}: the six reflexion configurations (\texttt{conf/prompts/reflexion/}) and the five meta profiles (\texttt{conf/prompts/meta/}) those runs used, and the system prompts sent with them. The templates are Python format strings. Names in single braces, such as \texttt{\{input\}}, \texttt{\{context\}}, \texttt{\{question\}}, \texttt{\{answer\}}, \texttt{\{label\_space\}} and \texttt{\{corner\_o/s/b\}}, are placeholders filled at runtime. Literal braces are written doubled in the template files; we print them single, as the model receives them, so braces around quoted JSON keys are literal text, not placeholders. Line breaks inside a template are not reproduced. In the code the conceptual vertex $C$ is named \texttt{B}, which is why the integration prompt's placeholders read \texttt{\{corner\_b\}}.

Not printed: the goal-free condition's reflexion prompts (\texttt{conf/prompts/reflexion/goalfree.yaml}; its meta profile is \texttt{default}, printed below), and template variants that produced no reported number (chain-of-thought answering, the duplicate-perspective control, the interactive \texttt{ask} profile and the MMLU-Pro verification profiles). All of them are in the released repository.

\subsection{Configurations and system prompts}
\label{sec:prompt-configs}
Each benchmark composes one reflexion configuration and one meta profile (\texttt{conf/experiment/<benchmark>.yaml}). The three tracking tasks share \texttt{bbh\_tracking}. Hi-ToM uses the \texttt{default} reflexion configuration together with the \texttt{opentom} meta profile, which was written for the narrative theory-of-mind benchmark OpenToM.

\begin{center}
\footnotesize
\begin{tabular}{@{}lll@{}}
\toprule
Benchmark & Reflexion configuration & Meta profile \\ \midrule
GPQA Diamond & \texttt{gpqa} & \texttt{gpqa} \\
ExploreToM & \texttt{default} & \texttt{exploretom} \\
Hi-ToM & \texttt{default} & \texttt{opentom} \\
BBH tracking shuffled objects, 3 objects (T3) & \texttt{bbh\_tracking} & \texttt{default} \\
BBH tracking shuffled objects, 5 objects (T5) & \texttt{bbh\_tracking} & \texttt{default} \\
BBH tracking shuffled objects, 7 objects (T7) & \texttt{bbh\_tracking} & \texttt{default} \\
BBH causal judgment (CJ) & \texttt{bbh\_causal} & \texttt{default} \\
BBH word sorting (WS) & \texttt{bbh\_freeform} & \texttt{default} \\
MMLU-Pro & \texttt{mmlu\_pro} & \texttt{mmlu\_pro} \\
\bottomrule
\end{tabular}
\end{center}

\noindent\textbf{System prompts.} Every call made by a hosted answerer (gpt-4.1-nano, gpt-4o-mini, gemini-2.5-flash-lite) carries the first system prompt below, on every benchmark, including GPQA, BIG-Bench Hard and MMLU-Pro. Answerers served by Ollama (Qwen2.5-3b, and Llama3.2-3b outside BIG-Bench Hard) and the Llama3.2-3b runs through OpenRouter receive no system prompt. The judge and the meta-judge carry their own system prompts in every run.

\noindent\textbf{hosted answerer:}
\par\noindent\small\ttfamily You are an expert in modeling others' mental states.
\normalfont\normalsize

\noindent\textbf{judge:}
\par\noindent\small\ttfamily You are a careful grader. Answer only with the requested label.
\normalfont\normalsize

\noindent\textbf{meta-judge:}
\par\noindent\small\ttfamily You are an impartial evaluator of reasoning quality.
\normalfont\normalsize

\subsection{Boundary prompts (\texorpdfstring{$\psi_{OS}, \psi_{OC}, \psi_{SC}$}{psi\_OS, psi\_OC, psi\_SC})}
\label{sec:prompt-boundary}
One prompt per edge of the triangle (\Cref{def:boundary}). Each writes the sentence that states where the two adjacent perspectives diverge on this input.

\noindent\textbf{GPQA Diamond (\texttt{gpqa}), $\psi_{OS}$:}
\par\noindent\small\ttfamily Given this graduate-level science multiple-choice problem: \{input\} In one or two plain-English sentences, where is the boundary between established facts / definitions and uncertain inference or modeling assumptions? Do not pick an answer letter yet.
\normalfont\normalsize

\noindent\textbf{GPQA Diamond (\texttt{gpqa}), $\psi_{OC}$:}
\par\noindent\small\ttfamily Given this graduate-level science multiple-choice problem: \{input\} In one or two plain-English sentences, where is the boundary between concrete quantities / mechanisms and the abstract principle or law that organizes them? Do not pick an answer letter yet.
\normalfont\normalsize

\noindent\textbf{GPQA Diamond (\texttt{gpqa}), $\psi_{SC}$:}
\par\noindent\small\ttfamily Given this graduate-level science multiple-choice problem: \{input\} In one or two plain-English sentences, where is the boundary between the governing concept and the heuristic shortcuts or distractors that could mislead a solver? Do not pick an answer letter yet.
\normalfont\normalsize

\noindent\textbf{ExploreToM and Hi-ToM (\texttt{default}), $\psi_{OS}$:}
\par\noindent\small\ttfamily Given the story text: \{input\} In one or two plain-English sentences, where is the boundary between objectivity and subjectivity in this story? Do not use math, codes, IDs, or puzzle formatting.
\normalfont\normalsize

\noindent\textbf{ExploreToM and Hi-ToM (\texttt{default}), $\psi_{OC}$:}
\par\noindent\small\ttfamily Given the story text: \{input\} In one or two plain-English sentences, where is the boundary between objectivity and conceptuality in this story? Do not use math, codes, IDs, or puzzle formatting.
\normalfont\normalsize

\noindent\textbf{ExploreToM and Hi-ToM (\texttt{default}), $\psi_{SC}$:}
\par\noindent\small\ttfamily Given the story text: \{input\} In one or two plain-English sentences, where is the boundary between conceptuality and subjectivity in this story? Do not use math, codes, IDs, or puzzle formatting.
\normalfont\normalsize

\noindent\textbf{BBH T3, T5 and T7 (\texttt{bbh\_tracking}), $\psi_{OS}$:}
\par\noindent\small\ttfamily Given this swap-tracking problem: \{input\} In one or two plain-English sentences, where is the boundary between the stated starting assignment (who/what starts with which item or role) and the sequence of swap operations that follow it? Do not state a final answer yet.
\normalfont\normalsize

\noindent\textbf{BBH T3, T5 and T7 (\texttt{bbh\_tracking}), $\psi_{OC}$:}
\par\noindent\small\ttfamily Given this swap-tracking problem: \{input\} In one or two plain-English sentences, where is the boundary between the literal list of swaps as written and the general rule a swap follows (exactly two entities exchange exactly what they each currently hold)? Do not state a final answer yet.
\normalfont\normalsize

\noindent\textbf{BBH T3, T5 and T7 (\texttt{bbh\_tracking}), $\psi_{SC}$:}
\par\noindent\small\ttfamily Given this swap-tracking problem: \{input\} In one or two plain-English sentences, where is the boundary between the swap rule itself and the bookkeeping mistakes a solver is likely to make applying it in order (e.g. updating the wrong entity, or using a stale earlier state)? Do not state a final answer yet.
\normalfont\normalsize

\noindent\textbf{BBH causal judgment (\texttt{bbh\_causal}), $\psi_{OS}$:}
\par\noindent\small\ttfamily Given this causal-judgment scenario: \{input\} In one or two plain-English sentences, where is the boundary between the facts of the scenario as stated and the ordinary-language judgment call a "typical person" would make about who or what caused the outcome? Do not answer Yes or No yet.
\normalfont\normalsize

\noindent\textbf{BBH causal judgment (\texttt{bbh\_causal}), $\psi_{OC}$:}
\par\noindent\small\ttfamily Given this causal-judgment scenario: \{input\} In one or two plain-English sentences, where is the boundary between the concrete sequence of events as stated and whether the named factor was a NORMAL/expected/permitted part of that sequence versus an ABNORMAL, unexpected, or prohibited one? Do not answer Yes or No yet.
\normalfont\normalsize

\noindent\textbf{BBH causal judgment (\texttt{bbh\_causal}), $\psi_{SC}$:}
\par\noindent\small\ttfamily Given this causal-judgment scenario: \{input\} In one or two plain-English sentences, where is the boundary between "this factor was technically necessary for the outcome" and "an ordinary person would actually single this factor out as the cause"? A typical person does not call every necessary factor a cause -- being necessary is not enough on its own. Do not answer Yes or No yet.
\normalfont\normalsize

\noindent\textbf{BBH word sorting (\texttt{bbh\_freeform}), $\psi_{OS}$:}
\par\noindent\small\ttfamily Given this task: \{input\} In one or two plain-English sentences, where is the boundary between what the task states outright and the solver's own approach to working it out? Do not state a final answer yet.
\normalfont\normalsize

\noindent\textbf{BBH word sorting (\texttt{bbh\_freeform}), $\psi_{OC}$:}
\par\noindent\small\ttfamily Given this task: \{input\} In one or two plain-English sentences, where is the boundary between the concrete given elements (the words, numbers, or symbols to process) and the rule or procedure that connects them to the answer? Do not state a final answer yet.
\normalfont\normalsize

\noindent\textbf{BBH word sorting (\texttt{bbh\_freeform}), $\psi_{SC}$:}
\par\noindent\small\ttfamily Given this task: \{input\} In one or two plain-English sentences, where is the boundary between the governing rule or procedure and the likely slips a solver could make applying it? Do not state a final answer yet.
\normalfont\normalsize

\noindent\textbf{MMLU-Pro (\texttt{mmlu\_pro}), $\psi_{OS}$:}
\par\noindent\small\ttfamily Given this multiple-choice exam problem: \{input\} In one or two plain-English sentences, where is the boundary between established facts or definitions and uncertain inference or assumptions? Do not pick an option index yet.
\normalfont\normalsize

\noindent\textbf{MMLU-Pro (\texttt{mmlu\_pro}), $\psi_{OC}$:}
\par\noindent\small\ttfamily Given this multiple-choice exam problem: \{input\} In one or two plain-English sentences, where is the boundary between concrete details and the abstract principle or rule that organizes them? Do not pick an option index yet.
\normalfont\normalsize

\noindent\textbf{MMLU-Pro (\texttt{mmlu\_pro}), $\psi_{SC}$:}
\par\noindent\small\ttfamily Given this multiple-choice exam problem: \{input\} In one or two plain-English sentences, where is the boundary between the governing concept and the heuristic shortcuts or distractors that could mislead a solver? Do not pick an option index yet.
\normalfont\normalsize

\subsection{Perspective prompts (\texorpdfstring{$\varphi_O, \varphi_S, \varphi_C$}{phi\_O, phi\_S, phi\_C})}
\label{sec:prompt-perspective}
One prompt per vertex (\Cref{def:perspective}). \texttt{\{context\}} is the region context: the apex label and the two boundary statements the corner inherits, rendered as \texttt{Apex: <v>}, \texttt{Left boundary: <text>}, \texttt{Right boundary: <text>}. At the root it is the apex line alone.

\noindent\textbf{GPQA Diamond (\texttt{gpqa}), $\varphi_O$:}
\par\noindent\small\ttfamily Given this graduate-level science multiple-choice problem: \{input\} Region context: \{context\} From an objective perspective, list the key facts, quantities, constraints, and definitions that must hold (2-4 sentences). Do not commit to a final choice letter yet.
\normalfont\normalsize

\noindent\textbf{GPQA Diamond (\texttt{gpqa}), $\varphi_S$:}
\par\noindent\small\ttfamily Given this graduate-level science multiple-choice problem: \{input\} Region context: \{context\} From a solver's subjective perspective, note likely traps, tempting distractors, and where intuition may conflict with the correct mechanism (2-4 sentences). Do not commit to a final choice letter yet.
\normalfont\normalsize

\noindent\textbf{GPQA Diamond (\texttt{gpqa}), $\varphi_C$:}
\par\noindent\small\ttfamily Given this graduate-level science multiple-choice problem: \{input\} Region context: \{context\} From a conceptual perspective, name the governing principle, law, or mechanism and how it structures the solution (2-4 sentences). Do not commit to a final choice letter yet.
\normalfont\normalsize

\noindent\textbf{ExploreToM and Hi-ToM (\texttt{default}), $\varphi_O$:}
\par\noindent\small\ttfamily Given the story text: \{input\} Region context: \{context\} Rewrite the story from an objective perspective in plain English (2-4 sentences). Keep character names and object locations. Do not use math, codes, IDs, or puzzle formatting.
\normalfont\normalsize

\noindent\textbf{ExploreToM and Hi-ToM (\texttt{default}), $\varphi_S$:}
\par\noindent\small\ttfamily Given the story text: \{input\} Region context: \{context\} Rewrite the story from a subjective perspective in plain English (2-4 sentences). Keep character names, feelings, and beliefs. Do not use math, codes, IDs, or puzzle formatting.
\normalfont\normalsize

\noindent\textbf{ExploreToM and Hi-ToM (\texttt{default}), $\varphi_C$:}
\par\noindent\small\ttfamily Given the story text: \{input\} Region context: \{context\} Rewrite the story from a conceptual perspective in plain English (2-4 sentences). Keep the core events and mental-state structure. Do not use math, codes, IDs, or puzzle formatting.
\normalfont\normalsize

\noindent\textbf{BBH T3, T5 and T7 (\texttt{bbh\_tracking}), $\varphi_O$:}
\par\noindent\small\ttfamily Given this swap-tracking problem: \{input\} Region context: \{context\} From an objective perspective, state the starting assignment (who/what starts with which item or role) and list every swap in the exact order given (2-4 sentences). Do not compute a final answer yet.
\normalfont\normalsize

\noindent\textbf{BBH T3, T5 and T7 (\texttt{bbh\_tracking}), $\varphi_S$:}
\par\noindent\small\ttfamily Given this swap-tracking problem: \{input\} Region context: \{context\} From a solver's perspective, apply the swaps one at a time in order, updating who currently holds what after each one, and note where a tracking slip would most likely happen (2-4 sentences). Do not state the final answer yet -- show the running state instead.
\normalfont\normalsize

\noindent\textbf{BBH T3, T5 and T7 (\texttt{bbh\_tracking}), $\varphi_C$:}
\par\noindent\small\ttfamily Given this swap-tracking problem: \{input\} Region context: \{context\} From a conceptual perspective, state the governing rule plainly: each swap exchanges exactly what the two named entities currently hold, nothing else changes, and swaps apply strictly in the order given (2-4 sentences).
\normalfont\normalsize

\noindent\textbf{BBH causal judgment (\texttt{bbh\_causal}), $\varphi_O$:}
\par\noindent\small\ttfamily Given this causal-judgment scenario: \{input\} Region context: \{context\} From an objective perspective, list the stated facts in order: the setup, what counts as the normal/expected/permitted way things were supposed to go, the actions taken by each party, and the outcome (2-4 sentences). Do not answer Yes or No yet.
\normalfont\normalsize

\noindent\textbf{BBH causal judgment (\texttt{bbh\_causal}), $\varphi_S$:}
\par\noindent\small\ttfamily Given this causal-judgment scenario: \{input\} Region context: \{context\} From the perspective of how an ordinary person would judge this, apply this test: if the named factor was a normal, expected, or explicitly permitted action -- especially one where other parties did the identical normal thing at the same time -- an ordinary person usually does NOT call it "the cause," even though it was part of the causal chain. An ordinary person DOES call a factor "the cause" when it was abnormal, unexpected, prohibited, or a deliberate choice that stands out against the normal background. Say which of these two patterns this scenario matches and why (2-4 sentences). Do not answer Yes or No yet.
\normalfont\normalsize

\noindent\textbf{BBH causal judgment (\texttt{bbh\_causal}), $\varphi_C$:}
\par\noindent\small\ttfamily Given this causal-judgment scenario: \{input\} Region context: \{context\} From a conceptual perspective, state plainly: counterfactual necessity (the outcome would not have happened "but for" this factor) is NECESSARY but NOT SUFFICIENT for an ordinary person to call something a cause. The decisive test is whether the factor was abnormal/unexpected/prohibited/a deliberate intervening choice, versus normal/expected/permitted/shared equally among several parties. Name which side of that test this scenario falls on (2-4 sentences).
\normalfont\normalsize

\noindent\textbf{BBH word sorting (\texttt{bbh\_freeform}), $\varphi_O$:}
\par\noindent\small\ttfamily Given this task: \{input\} Region context: \{context\} From an objective perspective, list the given elements and exactly what the task is asking for (2-4 sentences). Do not state a final answer yet.
\normalfont\normalsize

\noindent\textbf{BBH word sorting (\texttt{bbh\_freeform}), $\varphi_S$:}
\par\noindent\small\ttfamily Given this task: \{input\} Region context: \{context\} From a solver's subjective perspective, note the approach being tried and where a slip is most likely (2-4 sentences). Do not state a final answer yet.
\normalfont\normalsize

\noindent\textbf{BBH word sorting (\texttt{bbh\_freeform}), $\varphi_C$:}
\par\noindent\small\ttfamily Given this task: \{input\} Region context: \{context\} From a conceptual perspective, name the governing rule or procedure and how it structures the solution (2-4 sentences). Do not state a final answer yet.
\normalfont\normalsize

\noindent\textbf{MMLU-Pro (\texttt{mmlu\_pro}), $\varphi_O$:}
\par\noindent\small\ttfamily Given this multiple-choice exam problem: \{input\} Region context: \{context\} From an objective perspective, list the key facts, quantities, constraints, and definitions that must hold (2-4 sentences). Do not commit to a final option index yet.
\normalfont\normalsize

\noindent\textbf{MMLU-Pro (\texttt{mmlu\_pro}), $\varphi_S$:}
\par\noindent\small\ttfamily Given this multiple-choice exam problem: \{input\} Region context: \{context\} From a solver's subjective perspective, note likely traps, tempting distractors, and where intuition may conflict with the correct reasoning (2-4 sentences). Do not commit to a final option index yet.
\normalfont\normalsize

\noindent\textbf{MMLU-Pro (\texttt{mmlu\_pro}), $\varphi_C$:}
\par\noindent\small\ttfamily Given this multiple-choice exam problem: \{input\} Region context: \{context\} From a conceptual perspective, name the governing principle, rule, or mechanism and how it structures the solution (2-4 sentences). Do not commit to a final option index yet.
\normalfont\normalsize

\subsection{Integration prompt (\texorpdfstring{$T$}{T})}
\label{sec:prompt-integration}
The prompt that merges the input and the three outermost corner perspectives into the single summary $y$ (\Cref{def:reflection}). It runs once per pass, whatever the depth.

\noindent\textbf{GPQA Diamond (\texttt{gpqa}), $T$:}
\par\noindent\small\ttfamily Original problem: \{input\} Outermost corner perspectives from a recursive reflexion analysis: Objectivity (O): \{corner\_o\} Subjectivity (S): \{corner\_s\} Conceptuality (B): \{corner\_b\} Write one concise plain-English integrated scientific reasoning summary for solving this problem. Cover the governing principle, critical constraints, and how distractors fail. Do NOT state a final answer letter (A/B/C/D).
\normalfont\normalsize

\noindent\textbf{ExploreToM and Hi-ToM (\texttt{default}), $T$:}
\par\noindent\small\ttfamily Original story: \{input\} Outermost corner perspectives from a recursive reflexion analysis: Objectivity (O): \{corner\_o\} Subjectivity (S): \{corner\_s\} Conceptuality (B): \{corner\_b\} Write one concise plain-English integrated summary of the story events and the characters' beliefs. Preserve names, places, and who knows what. Do not use math, codes, segment IDs, or puzzle formatting.
\normalfont\normalsize

\noindent\textbf{BBH T3, T5 and T7 (\texttt{bbh\_tracking}), $T$:}
\par\noindent\small\ttfamily Original problem: \{input\} Outermost corner perspectives from a recursive reflexion analysis: Objectivity (O): \{corner\_o\} Subjectivity (S): \{corner\_s\} Conceptuality (B): \{corner\_b\} Apply every swap in order, starting from the stated initial assignment, and write one concise plain-English summary giving the FINAL holder of every item or role after all swaps are applied. State the ending assignment for every named entity, not just the one the question asks about. Do NOT state which option letter is correct.
\normalfont\normalsize

\noindent\textbf{BBH causal judgment (\texttt{bbh\_causal}), $T$:}
\par\noindent\small\ttfamily Original scenario: \{input\} Outermost corner perspectives from a recursive reflexion analysis: Objectivity (O): \{corner\_o\} Subjectivity (S): \{corner\_s\} Conceptuality (B): \{corner\_b\} Write one concise plain-English integrated analysis of the causal question. State the relevant facts, then apply this decision procedure explicitly: counterfactual necessity alone does NOT make something "the cause" to a typical person -- normal, expected, permitted actions (especially ones shared identically among several parties) are usually NOT named as causes even when necessary for the outcome; abnormal, unexpected, prohibited, or deliberately chosen actions ARE named as causes. State which side of that test the named factor falls on. Do NOT answer Yes or No yet.
\normalfont\normalsize

\noindent\textbf{BBH word sorting (\texttt{bbh\_freeform}), $T$:}
\par\noindent\small\ttfamily Original task: \{input\} Outermost corner perspectives from a recursive reflexion analysis: Objectivity (O): \{corner\_o\} Subjectivity (S): \{corner\_s\} Conceptuality (B): \{corner\_b\} Write one concise plain-English integrated summary of how to work out this task: the governing rule, the given elements, and how they connect. Do NOT state a final answer.
\normalfont\normalsize

\noindent\textbf{MMLU-Pro (\texttt{mmlu\_pro}), $T$:}
\par\noindent\small\ttfamily Original problem: \{input\} Outermost corner perspectives from a recursive reflexion analysis: Objectivity (O): \{corner\_o\} Subjectivity (S): \{corner\_s\} Conceptuality (B): \{corner\_b\} Write one concise plain-English integrated reasoning summary for solving this problem. Cover the governing principle, critical constraints, and how distractors fail. Do NOT state a final option index.
\normalfont\normalsize

\subsection{Answering prompt (\texorpdfstring{$\beta$}{beta})}
\label{sec:prompt-answering}
The prompt that answers the question from a summary (\Cref{def:reflection}). At $T^0$ the summary is the raw input, so this is also the direct-answering baseline.

\noindent\textbf{GPQA Diamond (\texttt{gpqa}), $\beta$:}
\par\noindent\small\ttfamily You are answering a graduate-level science multiple-choice question (GPQA). Reasoning context (may be the raw problem or an integrated analysis): \{context\} Question / task: Choose the correct option for the problem above. \{question\} Allowed answers (choose exactly one letter): \{label\_space\} Rules: - Reply with ONLY one letter from the allowed answers. - Prefer the form: \textbackslash{}\textbackslash{}boxed\{LETTER\} - No explanation, no analysis, no extra words.
\normalfont\normalsize

\noindent\textbf{ExploreToM and Hi-ToM (\texttt{default}), $\beta$:}
\par\noindent\small\ttfamily You are an expert in modeling others' mental states. Read the following context and answer the question. Context: \{context\} Question: \{question\} Allowed answers (choose exactly one): \{label\_space\} Rules: - Reply with ONLY one allowed label. - Prefer the form: \textbackslash{}\textbackslash{}boxed\{LABEL\} where LABEL is exactly one allowed answer. - No explanation, no analysis, no extra words.
\normalfont\normalsize

\noindent\textbf{BBH T3, T5 and T7 (\texttt{bbh\_tracking}), $\beta$:}
\par\noindent\small\ttfamily You are solving a mechanical state-tracking puzzle (BIG-Bench Hard: tracking shuffled objects). Reasoning context (may be the raw problem or an integrated final-assignment summary): \{context\} Question: \{question\} Allowed answers (choose exactly one): \{label\_space\} Rules: - Reply with ONLY one allowed label. - Prefer the form: \textbackslash{}\textbackslash{}boxed\{LABEL\} where LABEL is exactly one allowed answer. - No explanation, no analysis, no extra words.
\normalfont\normalsize

\noindent\textbf{BBH causal judgment (\texttt{bbh\_causal}), $\beta$:}
\par\noindent\small\ttfamily You are answering how a typical person would judge a causal question (BIG-Bench Hard: causal judgement). Reasoning context (may be the raw scenario or an integrated causal analysis): \{context\} Question: \{question\} Allowed answers (choose exactly one): \{label\_space\} Rules: - Reply with ONLY one allowed label. - Prefer the form: \textbackslash{}\textbackslash{}boxed\{LABEL\} - No explanation, no analysis, no extra words.
\normalfont\normalsize

\noindent\textbf{BBH word sorting (\texttt{bbh\_freeform}), $\beta$:}
\par\noindent\small\ttfamily You are solving a short, exact-answer task. Reasoning context (may be the raw task or an integrated analysis): \{context\} Task: \{question\} Work out the answer, then state it. Rules: - Reply with ONLY the final answer, exactly as the task's own format expects (e.g. the sorted list, the computed number, the transformed string). - No explanation, no analysis, no extra words.
\normalfont\normalsize

\noindent\textbf{MMLU-Pro (\texttt{mmlu\_pro}), $\beta$:}
\par\noindent\small\ttfamily You are answering a multiple-choice exam question (MMLU-Pro), in one of 14 subjects. Reasoning context (may be the raw problem or an integrated analysis): \{context\} Question / task: Choose the correct option for the problem above. \{question\} Allowed answers (choose exactly one option index; indices start at 0): \{label\_space\} Rules: - Reply with ONLY the index of the correct option, one of the allowed answers. - Prefer the form: \textbackslash{}\textbackslash{}boxed\{INDEX\} - No explanation, no analysis, no extra words.
\normalfont\normalsize

\subsection{Grading prompt}
\label{sec:prompt-grader}
Grading is not part of the architecture. The benchmark adapter first maps the answerer's reply onto the allowed labels, and a reply that lands on one is scored by exact comparison with the gold label, with no model call; almost every answer is scored this way. Only a reply that does not land on an allowed label is sent to the judge model with the prompt below. If that call fails or cannot be parsed, a plain string comparison decides. The run file records which of the three methods scored each answer.

\noindent\textbf{GPQA Diamond (\texttt{gpqa}), $\textsc{grade}$:}
\par\noindent\small\ttfamily You are evaluating a GPQA multiple-choice answer. Gold answer: \{gold\} Model answer: \{prediction\} Allowed answers: \{label\_space\} Decide whether the model answer matches the gold letter (ignore surrounding text if the letter is clear). Reply with JSON only: \{"correct": true or false, "normalized\_answer": "A or B or C or D or UNKNOWN"\}
\normalfont\normalsize

\noindent\textbf{ExploreToM and Hi-ToM (\texttt{default}), $\textsc{grade}$:}
\par\noindent\small\ttfamily You are evaluating a Theory-of-Mind benchmark answer. Gold answer: \{gold\} Model answer: \{prediction\} Allowed answers: \{label\_space\} Decide whether the model answer matches the gold answer in meaning (allow minor wording differences). Reply with JSON only: \{"correct": true or false, "normalized\_answer": "the model answer mapped to an allowed label or UNKNOWN"\}
\normalfont\normalsize

\noindent\textbf{BBH T3, T5 and T7 (\texttt{bbh\_tracking}), $\textsc{grade}$:}
\par\noindent\small\ttfamily You are evaluating a state-tracking benchmark answer (BIG-Bench Hard: tracking shuffled objects). Gold answer: \{gold\} Model answer: \{prediction\} Allowed answers: \{label\_space\} Decide whether the model answer matches the gold answer (allow minor wording differences, e.g. the item name vs. its letter). Reply with JSON only: \{"correct": true or false, "normalized\_answer": "the model answer mapped to an allowed label or UNKNOWN"\}
\normalfont\normalsize

\noindent\textbf{BBH causal judgment (\texttt{bbh\_causal}), $\textsc{grade}$:}
\par\noindent\small\ttfamily You are evaluating a causal-judgment benchmark answer (BIG-Bench Hard: causal judgement). Gold answer: \{gold\} Model answer: \{prediction\} Allowed answers: \{label\_space\} Decide whether the model answer matches the gold answer (allow minor wording differences, e.g. "yes"/"Yes"). Reply with JSON only: \{"correct": true or false, "normalized\_answer": "the model answer mapped to an allowed label or UNKNOWN"\}
\normalfont\normalsize

\noindent\textbf{BBH word sorting (\texttt{bbh\_freeform}), $\textsc{grade}$:}
\par\noindent\small\ttfamily You are evaluating a short, exact-answer task. Gold answer: \{gold\} Model answer: \{prediction\} Decide whether the model's stated final answer matches the gold answer exactly (ignore surrounding text; look for the model's stated final answer). Reply with JSON only: \{"correct": true or false, "normalized\_answer": "the model's final answer, or UNKNOWN"\}
\normalfont\normalsize

\noindent\textbf{MMLU-Pro (\texttt{mmlu\_pro}), $\textsc{grade}$:}
\par\noindent\small\ttfamily You are evaluating an MMLU-Pro multiple-choice answer. Gold answer: \{gold\} Model answer: \{prediction\} Allowed answers: \{label\_space\} Decide whether the model answer matches the gold index (ignore surrounding text if the index is clear). Reply with JSON only: \{"correct": true or false, "normalized\_answer": "a single index from the allowed answers, or UNKNOWN"\}
\normalfont\normalsize

\subsection{PolyJudge prompts (\texorpdfstring{$J_S, J_O, J_C$}{J\_S, J\_O, J\_C})}
\label{sec:prompt-judge}
One prompt per judging vertex (\Cref{def:polyjudge}), run by the meta-judge. The five profiles ask the same three questions but are not rewordings of one another: the task-framed profiles spell out what to check in the terms of the task (the \texttt{exploretom} and \texttt{opentom} objective judges, for instance, list the facts a story summary must preserve, including false beliefs), so all five are printed.

\noindent\textbf{\texttt{default} profile (BBH T3, T5, T7, causal judgment and word sorting), $J_S$:}
\par\noindent\small\ttfamily You are a polycontextural judge for the SUBJECTIVE context. Original input: \{question\} Answer under evaluation: \{answer\} Perspective artifacts from the reasoning process (O = objective, S = subjective, B = conceptual): \{corners\} Question: Is the answer authentic, internally coherent, and faithful to its own perspective? Judge coherence within the answer's own standpoint, not factual truth. In your rationale, name the specific statements that drove your verdict. Reply with JSON only: \{"value": "A" or "F", "rationale": "one short paragraph naming the decisive statements"\}
\normalfont\normalsize

\noindent\textbf{\texttt{gpqa} profile (GPQA Diamond), $J_S$:}
\par\noindent\small\ttfamily You are a polycontextural judge for the SUBJECTIVE context (graduate STEM reasoning summary). Original problem: \{question\} Integrated reasoning summary under evaluation: \{answer\} Perspective artifacts from reflexion (O = facts/constraints, S = traps/intuition, B = governing principle): \{corners\} Is the summary authentic and internally coherent as a solver's reasoning narrative? Judge coherence of assumptions and eliminations — not whether the final letter would be correct. Name the specific statements that drove your verdict. Reply with JSON only: \{"value": "A" or "F", "rationale": "one short paragraph"\}
\normalfont\normalsize

\noindent\textbf{\texttt{exploretom} profile (ExploreToM), $J_S$:}
\par\noindent\small\ttfamily You are a polycontextural judge for the SUBJECTIVE context (Theory-of-Mind story summary). Original narrative: \{question\} Integrated story summary under evaluation: \{answer\} Perspective artifacts from reflexion (O = objective facts, S = subjective experience, B = conceptual structure): \{corners\} Is the summary authentic and internally coherent as a mental-state narrative? Judge coherence of what each character believes, what each character believes about the others' beliefs, and what each character was in a position to learn. Do not judge whether every inference is objectively true. Name the specific statements that drove your verdict. Reply with JSON only: \{"value": "A" or "F", "rationale": "one short paragraph"\}
\normalfont\normalsize

\noindent\textbf{\texttt{opentom} profile (Hi-ToM), $J_S$:}
\par\noindent\small\ttfamily You are a polycontextural judge for the SUBJECTIVE context (Theory-of-Mind story summary). Original narrative: \{question\} Integrated story summary under evaluation: \{answer\} Perspective artifacts from reflexion (O = objective facts, S = subjective experience, B = conceptual structure): \{corners\} Is the summary authentic and internally coherent as a mental-state narrative? Judge coherence of beliefs, intentions, and emotional stance — not whether every inference is objectively true. Name the specific statements that drove your verdict. Reply with JSON only: \{"value": "A" or "F", "rationale": "one short paragraph"\}
\normalfont\normalsize

\noindent\textbf{\texttt{mmlu\_pro} profile (MMLU-Pro), $J_S$:}
\par\noindent\small\ttfamily You are a polycontextural judge for the SUBJECTIVE context (exam reasoning summary). Original problem: \{question\} Integrated reasoning summary under evaluation: \{answer\} Perspective artifacts from reflexion (O = facts/constraints, S = traps/intuition, B = governing principle): \{corners\} Is the summary authentic and internally coherent as a solver's reasoning narrative? Judge coherence of assumptions and eliminations — not whether the final index would be correct. Name the specific statements that drove your verdict. Reply with JSON only: \{"value": "A" or "F", "rationale": "one short paragraph"\}
\normalfont\normalsize

\noindent\textbf{\texttt{default} profile (BBH T3, T5, T7, causal judgment and word sorting), $J_O$:}
\par\noindent\small\ttfamily You are a polycontextural judge for the OBJECTIVE context. Original input: \{question\} Answer under evaluation: \{answer\} Question: Is the answer objectively correct and defensible given the input? In your rationale, name the specific statements that drove your verdict. Reply with JSON only: \{"value": "W" or "F", "rationale": "one short paragraph naming the decisive statements"\}
\normalfont\normalsize

\noindent\textbf{\texttt{gpqa} profile (GPQA Diamond), $J_O$:}
\par\noindent\small\ttfamily You are a polycontextural judge for the OBJECTIVE context (graduate STEM reasoning summary). Original problem: \{question\} Integrated reasoning summary under evaluation: \{answer\} Is the summary objectively faithful to the problem's scientific content? Check: quantities, units, definitions, mechanisms, and that claimed facts are supported by the stem. Penalize invented numbers, dropped constraints, or misstated laws. Name the specific statements that drove your verdict. Reply with JSON only: \{"value": "W" or "F", "rationale": "one short paragraph"\}
\normalfont\normalsize

\noindent\textbf{\texttt{exploretom} profile (ExploreToM), $J_O$:}
\par\noindent\small\ttfamily You are a polycontextural judge for the OBJECTIVE context (Theory-of-Mind story summary). Original narrative: \{question\} Integrated story summary under evaluation: \{answer\} Is the summary objectively faithful to the narrative? Check: character names, which room each character was in and when, which container each object was moved into, who was present for each move, who witnessed a move in secret, who was distracted and missed it, and who was told something privately rather than out loud. Penalize invented facts, and penalize any place where the summary states as fact something only one character believes. Name the specific statements that drove your verdict. Reply with JSON only: \{"value": "W" or "F", "rationale": "one short paragraph"\}
\normalfont\normalsize

\noindent\textbf{\texttt{opentom} profile (Hi-ToM), $J_O$:}
\par\noindent\small\ttfamily You are a polycontextural judge for the OBJECTIVE context (Theory-of-Mind story summary). Original narrative: \{question\} Integrated story summary under evaluation: \{answer\} Is the summary objectively faithful to the narrative? Check: character names, object locations, moves, who witnessed what, and especially FALSE BELIEFS (what a character thinks vs what is actually true). Penalize invented facts or dropped false-belief structure. Name the specific statements that drove your verdict. Reply with JSON only: \{"value": "W" or "F", "rationale": "one short paragraph"\}
\normalfont\normalsize

\noindent\textbf{\texttt{mmlu\_pro} profile (MMLU-Pro), $J_O$:}
\par\noindent\small\ttfamily You are a polycontextural judge for the OBJECTIVE context (exam reasoning summary). Original problem: \{question\} Integrated reasoning summary under evaluation: \{answer\} Is the summary objectively faithful to the problem's subject-matter content? Check: quantities, units, definitions, mechanisms, and that claimed facts are supported by the stem. Penalize invented numbers, dropped constraints, or misstated laws. Name the specific statements that drove your verdict. Reply with JSON only: \{"value": "W" or "F", "rationale": "one short paragraph"\}
\normalfont\normalsize

\noindent\textbf{\texttt{default} profile (BBH T3, T5, T7, causal judgment and word sorting), $J_C$:}
\par\noindent\small\ttfamily You are a polycontextural judge for the DIALECTICAL context. Original input: \{question\} Answer under evaluation: \{answer\} Answer from the previous cycle (may be '(none)' on the first cycle): \{previous\_answer\} Question: Does this answer introduce genuine semantic novelty relative to the question and the previous answer, or does it merely restate what was already implicit? In your rationale, name the specific statements that drove your verdict. Reply with JSON only: \{"value": "W" or "R", "rationale": "one short paragraph naming the decisive statements"\}
\normalfont\normalsize

\noindent\textbf{\texttt{gpqa} profile (GPQA Diamond), $J_C$:}
\par\noindent\small\ttfamily You are a polycontextural judge for the DIALECTICAL context (graduate STEM reasoning summary). Original problem: \{question\} Integrated reasoning summary under evaluation: \{answer\} Previous cycle's summary (may be '(none)' on the first cycle): \{previous\_answer\} Does this summary add useful scientific substance (mechanism detail, constraint use, distractor elimination) that was not already explicit in the previous summary? Or is it merely a paraphrase with no new solving power? Name the specific statements that drove your verdict. Reply with JSON only: \{"value": "W" or "R", "rationale": "one short paragraph"\}
\normalfont\normalsize

\noindent\textbf{\texttt{exploretom} profile (ExploreToM), $J_C$:}
\par\noindent\small\ttfamily You are a polycontextural judge for the DIALECTICAL context (Theory-of-Mind story summary). Original narrative: \{question\} Integrated story summary under evaluation: \{answer\} Previous cycle's summary (may be '(none)' on the first cycle): \{previous\_answer\} Does this summary add useful Theory-of-Mind detail that was not already explicit in the previous summary -- a character's belief that differs from reality, a belief one character holds about another's belief, or a moment where someone missed an event? Or is it merely a paraphrase with no new mental-state information? Name the specific statements that drove your verdict. Reply with JSON only: \{"value": "W" or "R", "rationale": "one short paragraph"\}
\normalfont\normalsize

\noindent\textbf{\texttt{opentom} profile (Hi-ToM), $J_C$:}
\par\noindent\small\ttfamily You are a polycontextural judge for the DIALECTICAL context (Theory-of-Mind story summary). Original narrative: \{question\} Integrated story summary under evaluation: \{answer\} Previous cycle's summary (may be '(none)' on the first cycle): \{previous\_answer\} Does this summary add useful Theory-of-Mind detail (perspective differences, belief contrasts, accessibility changes) that was not already explicit in the previous summary? Or is it merely a paraphrase with no new mental-state information? Name the specific statements that drove your verdict. Reply with JSON only: \{"value": "W" or "R", "rationale": "one short paragraph"\}
\normalfont\normalsize

\noindent\textbf{\texttt{mmlu\_pro} profile (MMLU-Pro), $J_C$:}
\par\noindent\small\ttfamily You are a polycontextural judge for the DIALECTICAL context (exam reasoning summary). Original problem: \{question\} Integrated reasoning summary under evaluation: \{answer\} Previous cycle's summary (may be '(none)' on the first cycle): \{previous\_answer\} Does this summary add useful substantive content (mechanism detail, constraint use, distractor elimination) that was not already explicit in the previous summary? Or is it merely a paraphrase with no new solving power? Name the specific statements that drove your verdict. Reply with JSON only: \{"value": "W" or "R", "rationale": "one short paragraph"\}
\normalfont\normalsize

\subsection{StrategyPlanner prompts (\texorpdfstring{$\Delta$}{Delta})}
\label{sec:prompt-strategy}
The prompts that turn a global value into the next cycle's input (\Cref{def:strategy}). All of them are run by the meta-judge model. \textsc{Expand} calls the negation prompt once per positively judged vertex, checks the two results against each other with the contradiction guard, and wraps each survivor in the expansion prompt. \textsc{Refine} needs no generation call: it copies the two negative rationales verbatim into the refinement prompt. \textsc{Reset} and \textsc{Stop} have no prompt of their own.

\noindent\textbf{\texttt{default} profile (BBH T3, T5, T7, causal judgment and word sorting), negation boundary $\psi^-_{v^+}$ (\textsc{Expand}):}
\par\noindent\small\ttfamily An answer was judged positively in the \{dimension\} context and will be expanded along that axis. Original question: \{question\} Answer: \{answer\} Full judge feedback on this answer: \{judge\_feedback\} Positive verdict in the \{dimension\} context (\{verdict\_label\}): \{rationale\} Failing judges — the boundary AND the next answer must help fix these: \{repair\_priorities\} Write a semantic boundary (2-4 plain English sentences) that: 1. States what still separates this answer from its opposite in the \{dimension\} context (describe the boundary, not the opposite). 2. Explicitly addresses each failing judge's concern so the next answer can earn their approval. 3. If dialectical failed (R = redundant), the boundary must name what genuinely new content is still missing.
\normalfont\normalsize

\noindent\textbf{\texttt{gpqa} profile (GPQA Diamond), negation boundary $\psi^-_{v^+}$ (\textsc{Expand}):}
\par\noindent\small\ttfamily You are refining a graduate STEM reasoning summary along the \{dimension\} axis. Original problem: \{question\} Current integrated summary: \{answer\} Judge feedback: \{judge\_feedback\} Positive verdict (\{dimension\}, \{verdict\_label\}): \{rationale\} Gaps to address (from failing judges): \{repair\_priorities\} Write a semantic boundary (2-4 sentences) describing what this summary still omits or blurs in the \{dimension\} context — especially constraints, mechanisms, or distractor analysis. Describe the boundary, not a full rewrite. Do not state a final answer letter.
\normalfont\normalsize

\noindent\textbf{\texttt{exploretom} profile (ExploreToM), negation boundary $\psi^-_{v^+}$ (\textsc{Expand}):}
\par\noindent\small\ttfamily You are refining a Theory-of-Mind story summary along the \{dimension\} axis. Original narrative: \{question\} Current integrated summary: \{answer\} Judge feedback: \{judge\_feedback\} Positive verdict (\{dimension\}, \{verdict\_label\}): \{rationale\} Gaps to address (from failing judges): \{repair\_priorities\} Write a semantic boundary (2-4 sentences) describing what this summary still omits or blurs in the \{dimension\} context -- especially who saw a move and who did not, what a character still believes that is no longer true, and what one character believes another believes. Describe the boundary, not a full rewrite. Preserve all character names, room names and container names from the narrative.
\normalfont\normalsize

\noindent\textbf{\texttt{opentom} profile (Hi-ToM), negation boundary $\psi^-_{v^+}$ (\textsc{Expand}):}
\par\noindent\small\ttfamily You are refining a Theory-of-Mind story summary along the \{dimension\} axis. Original narrative: \{question\} Current integrated summary: \{answer\} Judge feedback: \{judge\_feedback\} Positive verdict (\{dimension\}, \{verdict\_label\}): \{rationale\} Gaps to address (from failing judges): \{repair\_priorities\} Write a semantic boundary (2-4 sentences) describing what this summary still omits or blurs in the \{dimension\} context — especially false beliefs, who knows what, or location changes. Describe the boundary, not a full rewrite. Preserve all character names from the narrative.
\normalfont\normalsize

\noindent\textbf{\texttt{mmlu\_pro} profile (MMLU-Pro), negation boundary $\psi^-_{v^+}$ (\textsc{Expand}):}
\par\noindent\small\ttfamily You are refining a exam reasoning summary along the \{dimension\} axis. Original problem: \{question\} Current integrated summary: \{answer\} Judge feedback: \{judge\_feedback\} Positive verdict (\{dimension\}, \{verdict\_label\}): \{rationale\} Gaps to address (from failing judges): \{repair\_priorities\} Write a semantic boundary (2-4 sentences) describing what this summary still omits or blurs in the \{dimension\} context — especially constraints, mechanisms, or distractor analysis. Describe the boundary, not a full rewrite. Do not state a final option index.
\normalfont\normalsize

\noindent\textbf{\texttt{default} profile (BBH T3, T5, T7, causal judgment and word sorting), contradiction guard (\textsc{Expand}):}
\par\noindent\small\ttfamily Boundary A: \{first\} Boundary B: \{second\} Are these two boundary descriptions mutually contradictory? Reply with exactly one word: yes or no.
\normalfont\normalsize

\noindent\textbf{\texttt{gpqa} profile (GPQA Diamond), contradiction guard (\textsc{Expand}):}
\par\noindent\small\ttfamily Boundary A: \{first\} Boundary B: \{second\} Are these two boundary descriptions mutually contradictory for the same science problem? Reply with exactly one word: yes or no.
\normalfont\normalsize

\noindent\textbf{\texttt{exploretom} profile (ExploreToM), contradiction guard (\textsc{Expand}):}
\par\noindent\small\ttfamily Boundary A: \{first\} Boundary B: \{second\} Are these two boundary descriptions mutually contradictory for the same story summary? Reply with exactly one word: yes or no.
\normalfont\normalsize

\noindent\textbf{\texttt{opentom} profile (Hi-ToM), contradiction guard (\textsc{Expand}):}
\par\noindent\small\ttfamily Boundary A: \{first\} Boundary B: \{second\} Are these two boundary descriptions mutually contradictory for the same story summary? Reply with exactly one word: yes or no.
\normalfont\normalsize

\noindent\textbf{\texttt{mmlu\_pro} profile (MMLU-Pro), contradiction guard (\textsc{Expand}):}
\par\noindent\small\ttfamily Boundary A: \{first\} Boundary B: \{second\} Are these two boundary descriptions mutually contradictory for the same problem? Reply with exactly one word: yes or no.
\normalfont\normalsize

\noindent\textbf{\texttt{default} profile (BBH T3, T5, T7, causal judgment and word sorting), expansion input $x'_i$ (\textsc{Expand}):}
\par\noindent\small\ttfamily Original question: \{question\} Current answer: \{answer\} Judge feedback on the current answer (address every NOT satisfied item): \{judge\_feedback\} Expansion axis: \{dimension\} Semantic boundary for this axis: \{boundary\} Task: construct the missing semantic coordinate implied by this boundary and the judge feedback above. The new answer must: - Fix every failing judge concern listed above. - Preserve what the satisfied judges already approved. - Add genuine new substance if dialectical was R (redundant). Do not merely restate or polish the current answer. Answer the original question from the new coordinate.
\normalfont\normalsize

\noindent\textbf{\texttt{gpqa} profile (GPQA Diamond), expansion input $x'_i$ (\textsc{Expand}):}
\par\noindent\small\ttfamily Original problem: \{question\} Current integrated reasoning summary: \{answer\} Judge feedback (address every NOT satisfied item without dropping correct facts): \{judge\_feedback\} Expansion axis: \{dimension\} Semantic boundary: \{boundary\} Task: Write a NEW integrated scientific reasoning summary (plain English, one concise paragraph) that: - Keeps every correct constraint, quantity, and mechanism from the current summary and problem. - Adds the solving detail implied by the boundary and missing judge concerns. - Does NOT state a final answer letter (A/B/C/D) — only deepen the reasoning. - Explicitly notes why tempting distractors fail when relevant.
\normalfont\normalsize

\noindent\textbf{\texttt{exploretom} profile (ExploreToM), expansion input $x'_i$ (\textsc{Expand}):}
\par\noindent\small\ttfamily Original narrative: \{question\} Current integrated story summary: \{answer\} Judge feedback (address every NOT satisfied item without dropping correct facts): \{judge\_feedback\} Expansion axis: \{dimension\} Semantic boundary: \{boundary\} Task: Write a NEW integrated story summary (plain English, one concise paragraph) that: - Keeps every correct fact from the current summary and narrative: character names, room names, container names, and the order the moves happened in. - Adds the mental-state detail implied by the boundary and missing judge concerns. - Says, for each move, who was present, who watched in secret, and who missed it. - States clearly when a character's belief differs from reality, and when a character is wrong about what someone else believes. - Does NOT answer benchmark questions -- only summarize the story for Theory-of-Mind. - Avoids filler ("best friends", "good intentions") unless stated in the narrative.
\normalfont\normalsize

\noindent\textbf{\texttt{opentom} profile (Hi-ToM), expansion input $x'_i$ (\textsc{Expand}):}
\par\noindent\small\ttfamily Original narrative: \{question\} Current integrated story summary: \{answer\} Judge feedback (address every NOT satisfied item without dropping correct facts): \{judge\_feedback\} Expansion axis: \{dimension\} Semantic boundary: \{boundary\} Task: Write a NEW integrated story summary (plain English, one concise paragraph) that: - Keeps every correct fact, false belief, and location from the current summary and narrative. - Adds the mental-state detail implied by the boundary and missing judge concerns. - Does NOT answer benchmark questions — only summarize the story for Theory-of-Mind. - Avoids filler ("best friends", "good intentions") unless stated in the narrative. - States clearly when a character's belief differs from reality.
\normalfont\normalsize

\noindent\textbf{\texttt{mmlu\_pro} profile (MMLU-Pro), expansion input $x'_i$ (\textsc{Expand}):}
\par\noindent\small\ttfamily Original problem: \{question\} Current integrated reasoning summary: \{answer\} Judge feedback (address every NOT satisfied item without dropping correct facts): \{judge\_feedback\} Expansion axis: \{dimension\} Semantic boundary: \{boundary\} Task: Write a NEW integrated reasoning summary (plain English, one concise paragraph) that: - Keeps every correct constraint, quantity, and mechanism from the current summary and problem. - Adds the solving detail implied by the boundary and missing judge concerns. - Does NOT state a final option index (0-9) — only deepen the reasoning. - Explicitly notes why tempting distractors fail when relevant.
\normalfont\normalsize

\noindent\textbf{\texttt{default} profile (BBH T3, T5, T7, causal judgment and word sorting), refinement input $x'_{\mathrm{ref}}$ (\textsc{Refine}):}
\par\noindent\small\ttfamily Original question: \{question\} Current answer: \{answer\} Judge feedback on the current answer: \{judge\_feedback\} Strength to preserve at apex \{apex\}: \{apex\_guidance\} [Region] Apex: \{apex\} Left boundary (\{left\_label\}): \{left\_boundary\} Right boundary (\{right\_label\}): \{right\_boundary\} Task: reason strictly inside this refined sub-region. The two boundaries come from failing judges — treat them as fixed constraints. Deepen the answer until it would satisfy ALL three judges (target triple: authentic, objectively defensible, and dialectically novel). Answer the original question directly.
\normalfont\normalsize

\noindent\textbf{\texttt{gpqa} profile (GPQA Diamond), refinement input $x'_{\mathrm{ref}}$ (\textsc{Refine}):}
\par\noindent\small\ttfamily Original problem: \{question\} Current integrated reasoning summary: \{answer\} Judge feedback: \{judge\_feedback\} Strength to preserve (apex \{apex\}): \{apex\_guidance\} [Region] Apex: \{apex\} Left boundary (\{left\_label\}): \{left\_boundary\} Right boundary (\{right\_label\}): \{right\_boundary\} Task: Write a refined integrated scientific reasoning summary inside this region. Fix the failing judge concerns using the two boundaries as constraints. Preserve correct science; deepen mechanism and distractor analysis. One concise paragraph; do NOT state a final answer letter.
\normalfont\normalsize

\noindent\textbf{\texttt{exploretom} profile (ExploreToM), refinement input $x'_{\mathrm{ref}}$ (\textsc{Refine}):}
\par\noindent\small\ttfamily Original narrative: \{question\} Current integrated story summary: \{answer\} Judge feedback: \{judge\_feedback\} Strength to preserve (apex \{apex\}): \{apex\_guidance\} [Region] Apex: \{apex\} Left boundary (\{left\_label\}): \{left\_boundary\} Right boundary (\{right\_label\}): \{right\_boundary\} Task: Write a refined integrated story summary inside this region. Fix the failing judge concerns using the two boundaries as constraints. Preserve the room names, the container names, the order of the moves, who witnessed what, and every belief that differs from reality. One concise paragraph; no benchmark Q\&A.
\normalfont\normalsize

\noindent\textbf{\texttt{opentom} profile (Hi-ToM), refinement input $x'_{\mathrm{ref}}$ (\textsc{Refine}):}
\par\noindent\small\ttfamily Original narrative: \{question\} Current integrated story summary: \{answer\} Judge feedback: \{judge\_feedback\} Strength to preserve (apex \{apex\}): \{apex\_guidance\} [Region] Apex: \{apex\} Left boundary (\{left\_label\}): \{left\_boundary\} Right boundary (\{right\_label\}): \{right\_boundary\} Task: Write a refined integrated story summary inside this region. Fix the failing judge concerns using the two boundaries as constraints. Preserve false-belief structure, locations, and witness information. One concise paragraph; no benchmark Q\&A.
\normalfont\normalsize

\noindent\textbf{\texttt{mmlu\_pro} profile (MMLU-Pro), refinement input $x'_{\mathrm{ref}}$ (\textsc{Refine}):}
\par\noindent\small\ttfamily Original problem: \{question\} Current integrated reasoning summary: \{answer\} Judge feedback: \{judge\_feedback\} Strength to preserve (apex \{apex\}): \{apex\_guidance\} [Region] Apex: \{apex\} Left boundary (\{left\_label\}): \{left\_boundary\} Right boundary (\{right\_label\}): \{right\_boundary\} Task: Write a refined integrated reasoning summary inside this region. Fix the failing judge concerns using the two boundaries as constraints. Preserve correct content; deepen mechanism and distractor analysis. One concise paragraph; do NOT state a final option index.
\normalfont\normalsize

\subsection{Drift check}
\label{sec:prompt-drift}
A guard outside the router (\Cref{sec:loop}): one yes/no call, run by the meta-judge, that rejects a cycle whose answer has stopped addressing the question. It receives the original input $x$ and the current cycle's answer.

\noindent\textbf{\texttt{default} profile (BBH T3, T5, T7, causal judgment and word sorting), drift check:}
\par\noindent\small\ttfamily Original question: \{question\} Answer: \{answer\} Does this answer still address the original question? Reply with exactly one word: yes or no.
\normalfont\normalsize

\noindent\textbf{\texttt{gpqa} profile (GPQA Diamond), drift check:}
\par\noindent\small\ttfamily Original problem: \{question\} Integrated reasoning summary: \{answer\} Does this summary still address the same scientific problem and constraints as the original (not a different topic)? Reply with exactly one word: yes or no.
\normalfont\normalsize

\noindent\textbf{\texttt{exploretom} profile (ExploreToM), drift check:}
\par\noindent\small\ttfamily Original narrative: \{question\} Integrated story summary: \{answer\} Does this summary still describe the same story events, characters, rooms and containers as the original narrative (not a different story or topic)? Reply with exactly one word: yes or no.
\normalfont\normalsize

\noindent\textbf{\texttt{opentom} profile (Hi-ToM), drift check:}
\par\noindent\small\ttfamily Original narrative: \{question\} Integrated story summary: \{answer\} Does this summary still describe the same story events, characters, and key beliefs as the original narrative (not a different story or topic)? Reply with exactly one word: yes or no.
\normalfont\normalsize

\noindent\textbf{\texttt{mmlu\_pro} profile (MMLU-Pro), drift check:}
\par\noindent\small\ttfamily Original problem: \{question\} Integrated reasoning summary: \{answer\} Does this summary still address the same problem and constraints as the original (not a different topic)? Reply with exactly one word: yes or no.
\normalfont\normalsize

\noindent\textbf{Reading the reply.} A reply beginning with ``no'' marks the cycle as drifted. Any other reply, and a failed call, count as on topic, so a broken check never stops a run.

\noindent\textbf{When it runs.} From cycle $1$ on. Cycle $0$ is not checked, since it is the answer to $x$ itself.

\noindent\textbf{Effect on the loop.} A drifted cycle cannot trigger \textsc{Stop}, even on the triple $(A,W,W)$. The next action is planned from the last on-topic cycle, so a drifted answer is never built upon. Two drifted cycles in a row end the run (\texttt{max\_consecutive\_drifts}, default $2$) and one on-topic cycle resets the count. The returned answer is chosen among the cycles that did not drift, when there are any.

\definecolor{good}{HTML}{2E7D32}
\definecolor{bad}{HTML}{C62828}
\section{Where Reflection Helps and Hurts on MMLU-Pro}
\label{sec:appendix-mmlupro}

This section looks closer at the 3,000-question MMLU-Pro run with gpt-4o-mini (\Cref{tab:mmlu-pro}). Every condition answers the same questions. A question is \emph{lost} when $T^0$ is right and the other condition is wrong, and \emph{gained} in the opposite case. The $p$-value is a two-sided sign test on the lost and gained counts. A star marks a value that stays below 0.05 after a Holm correction over the 14 subjects.

\begin{table}[ht]
\centering
\caption{\textbf{Paired comparison against $T^0$ by subject (gpt-4o-mini, 3,000 questions).} Accuracy in \%. Lost and gained count questions that change from right to wrong and from wrong to right. Shading compares a condition against $T^0$ in the same subject, green for a gain and red for a loss. Bold accuracy is the best of the three conditions, and a bold $p$ is below 0.05 before correction. $^{*}$: $p<0.05$ after Holm correction.}
\label{tab:mmlupro-paired}
\begin{adjustbox}{max width=\linewidth}
\footnotesize
\begin{tabular}{@{}lrrrrrrrrrr@{}}
\toprule
 & & \multicolumn{3}{c}{accuracy} & \multicolumn{3}{c}{$T^1$ vs $T^0$} & \multicolumn{3}{c}{$T^1,c{=}4$ vs $T^0$} \\
\cmidrule(lr){3-5}\cmidrule(lr){6-8}\cmidrule(lr){9-11}
Subject & $n$ & $T^0$ & $T^1$ & $T^1,c{=}4$ & lost & gained & $p$ & lost & gained & $p$ \\
\midrule
Math & 337 & 27.0 & \cellcolor{cellgain!12}27.6 & \cellcolor{cellgain!54}\textbf{42.7} & 26 & 28 & 0.892 & 24 & 77 & \textbf{$<$0.001}$^{*}$ \\
Physics & 324 & 25.0 & \cellcolor{cellgain!23}29.3 & \cellcolor{cellgain!40}\textbf{34.9} & 14 & 28 & \textbf{0.044} & 14 & 46 & \textbf{$<$0.001}$^{*}$ \\
Chemistry & 282 & 24.5 & \cellcolor{cellgain!18}27.0 & \cellcolor{cellgain!38}\textbf{33.7} & 18 & 25 & 0.360 & 20 & 46 & \textbf{0.002}$^{*}$ \\
Law & 275 & \textbf{34.5} & \cellcolor{cellloss!19}30.2 & \cellcolor{cellloss!24}28.0 & 35 & 23 & 0.148 & 34 & 16 & \textbf{0.015} \\
Engineering & 242 & 28.1 & \cellcolor{cellgain!11}28.5 & \cellcolor{cellgain!12}\textbf{28.9} & 17 & 18 & 1.000 & 21 & 23 & 0.880 \\
Other & 230 & 56.1 & 56.1 & \cellcolor{cellgain!11}\textbf{57.0} & 24 & 24 & 1.000 & 25 & 27 & 0.890 \\
Economics & 210 & 59.5 & \cellcolor{cellgain!14}\textbf{62.9} & \cellcolor{cellgain!14}\textbf{62.9} & 11 & 18 & 0.265 & 12 & 19 & 0.281 \\
Health & 204 & 62.3 & \cellcolor{cellgain!11}\textbf{63.2} & \cellcolor{cellgain!10}62.7 & 17 & 19 & 0.868 & 18 & 19 & 1.000 \\
Psychology & 199 & \textbf{69.8} & \cellcolor{cellloss!13}67.3 & \textbf{69.8} & 14 & 9 & 0.405 & 15 & 15 & 1.000 \\
Business & 197 & 29.9 & \cellcolor{cellloss!16}27.4 & \cellcolor{cellgain!38}\textbf{41.1} & 20 & 15 & 0.500 & 17 & 39 & \textbf{0.005} \\
Biology & 179 & \textbf{77.1} & \cellcolor{cellloss!12}74.9 & \cellcolor{cellloss!11}76.5 & 9 & 5 & 0.424 & 13 & 12 & 1.000 \\
Philosophy & 124 & 48.4 & \cellcolor{cellgain!14}\textbf{50.8} & \cellcolor{cellgain!12}50.0 & 6 & 9 & 0.607 & 13 & 15 & 0.851 \\
Computer Science & 102 & 47.1 & 47.1 & \cellcolor{cellgain!26}\textbf{56.9} & 10 & 10 & 1.000 & 8 & 18 & 0.076 \\
History & 95 & \textbf{56.8} & \cellcolor{cellloss!25}45.3 & \cellcolor{cellloss!21}48.4 & 16 & 5 & \textbf{0.027} & 13 & 5 & 0.096 \\
\midrule
\textbf{All} & 3000 & 42.8 & \cellcolor{cellloss!10}42.7 & \cellcolor{cellgain!18}\textbf{47.1} & 237 & 236 & 1.000 & 247 & 377 & \textbf{$<$0.001} \\
\bottomrule
\end{tabular}
\end{adjustbox}
\end{table}

\paragraph{Depth 1 against direct.} No subject is significant after the correction. History has the smallest uncorrected $p$ in the drop (16 lost, 5 gained, $p=0.027$), and physics has the smallest in the gain (14 lost, 28 gained, $p=0.044$). Neither stays below 0.05 after the Holm correction. The other four subjects below $T^0$ in \Cref{tab:mmlu-pro} are within noise: law (35 lost, 23 gained), business (20, 15), psychology (14, 9), and biology (9, 5). Overall, $T^1$ loses 237 questions and gains 236.

\paragraph{Meta controller against direct.} The gains concentrate in subjects that require calculation or several steps. Math gains 77 questions and loses 24. Physics gains 46 and loses 14. Chemistry gains 46 and loses 20. These three stay significant after the correction. Three subjects end below $T^0$: law, history and biology. Law is the only one of them with $p<0.05$ before the correction (34 lost, 16 gained, $p=0.015$; 0.15 after). History (13 lost, 5 gained, $p=0.096$) and biology (13, 12) are within noise.

\paragraph{What separates lost from gained questions.} This paragraph compares $T^1$ with $T^0$, so lost and gained refer to the 237 questions $T^1$ loses and the 236 it gains. Lost questions are longer. Their mean length is 345 characters, against 268 for gained questions and 204 for questions that both conditions answer correctly. The summaries have almost the same length in both groups (916 and 913 characters for lost and gained, respectively). The number of options is the same (9.4 and 9.5). We also searched the summaries for words such as ``common'', ``assume'', ``principle'', and ``constraint''. The share of summaries that contain each word differs by less than 10 points between lost and gained questions. None of the words we tried separates the two groups. The summary also does not point to an option. The option that shares the most words with the summary is the one $T^1$ picks in 13\% of the questions, close to the 11\% expected by chance.

\paragraph{What the lost examples have in common.} \textit{In the lost examples, the summary is not always wrong (see the highlighted text). It often states a rule that is true but too general, and the model applies it to the wrong option}. The law example asks which contract is most likely to be classified under the Uniform Commercial Code (UCC). The summary says the code governs the sale of tangible goods. This is correct, but it is more general than the question needs. The code also covers crops and timber to be severed from the property, which is the correct option. The model instead selects the sale of a personal vehicle, which better fits the general rule. The biology example asks for the frequency of concordance among 83 monozygotic twin pairs, where 13 pairs have both twins affected. The summary computes the concordance as 13 of 83 pairs. The correct method divides by the 33 pairs with at least one affected twin. The history example asks which period the Shang Dynasty's foundation for Chinese civilization lasted into. The summary tells the model to select a period that directly follows the Shang. The correct answer is the 20th century, which is far from the Shang. The gained examples show the opposite. The summary states the correct rule (felony murder does not depend on intent), the correct definition (alpha waves and relaxation), or a simple formula (homozygosity is 100\% minus 46\%).

\paragraph{Worked examples.} \Cref{fig:mmlupro-examples-history,fig:mmlupro-examples-law,fig:mmlupro-examples-psychology,fig:mmlupro-examples-biology} show one lost and one gained question ($T^1$ against $T^0$) for four of the five subjects where $T^1$ is below $T^0$; business is not shown. Each example shows the question, the options, and the summary that the model wrote before answering. The highlighted phrase is the claim in the summary that best explains the outcome. A summary is written based solely on the question, without the options.

\begin{table*}[!t]
\centering
\caption{\textbf{Macro-averaged F$_1$ on MMLU-Pro, by subject.} The same predictions as \Cref{tab:mmlu-pro}, read by answer class instead of by item. Questions have between three and ten options, but every subject contains all ten answer classes, so each subject is averaged over the ten classes. The large number is macro-F$_1$, and the small pair is macro-precision$/$macro-recall. Bold is the best macro-F$_1$ within the same model family for each subject. Shading follows \Cref{tab:mmlu-pro}: it compares a condition against $T^0$ in the same model, so green is a gain over direct answering and red a loss. The question counts per subject are in \Cref{tab:mmlu-pro}.}
\label{tab:mmlu-pro-macro}
\begin{adjustbox}{max width=\linewidth}
\footnotesize
\begin{tabular}{@{}lrrrrrrrrr@{}}
\toprule
 & \multicolumn{3}{c}{Qwen2.5-3b} & \multicolumn{3}{c}{gpt-4o-mini} & \multicolumn{3}{c}{gemini-2.5-flash-lite} \\
\cmidrule(lr){2-4}\cmidrule(lr){5-7}\cmidrule(lr){8-10}
 & $T^0$ & $T^1$ & $T^1, c=4$ & $T^0$ & $T^1$ & $T^1, c=4$ & $T^0$ & $T^1$ & $T^1, c=4$ \\ \midrule
Math & 22.6 {\scriptsize\textcolor{black!65}{23.9$/$23.3}} & \cellcolor{cellgain!55}\textbf{36.5} {\scriptsize\textcolor{black!65}{38.3$/$36.5}} & \cellcolor{cellloss!15}21.1 {\scriptsize\textcolor{black!65}{22.8$/$21.4}} & 26.4 {\scriptsize\textcolor{black!65}{32.1$/$27.3}} & \cellcolor{cellgain!12}26.9 {\scriptsize\textcolor{black!65}{33.2$/$27.8}} & \cellcolor{cellgain!55}\textbf{43.4} {\scriptsize\textcolor{black!65}{50.7$/$43.0}} & 39.4 {\scriptsize\textcolor{black!65}{40.4$/$39.7}} & \cellcolor{cellgain!27}48.5 {\scriptsize\textcolor{black!65}{49.1$/$48.8}} & \cellcolor{cellgain!46}\textbf{58.1} {\scriptsize\textcolor{black!65}{59.4$/$58.9}} \\
Physics & 24.5 {\scriptsize\textcolor{black!65}{25.6$/$24.7}} & \cellcolor{cellgain!19}\textbf{27.3} {\scriptsize\textcolor{black!65}{29.4$/$27.0}} & \cellcolor{cellloss!25}19.7 {\scriptsize\textcolor{black!65}{21.3$/$20.0}} & 23.7 {\scriptsize\textcolor{black!65}{32.1$/$24.5}} & \cellcolor{cellgain!26}28.7 {\scriptsize\textcolor{black!65}{34.5$/$29.5}} & \cellcolor{cellgain!46}\textbf{35.2} {\scriptsize\textcolor{black!65}{40.5$/$35.0}} & 39.9 {\scriptsize\textcolor{black!65}{42.3$/$40.1}} & \cellcolor{cellgain!12}41.2 {\scriptsize\textcolor{black!65}{41.9$/$41.7}} & \cellcolor{cellgain!14}\textbf{41.9} {\scriptsize\textcolor{black!65}{43.2$/$42.0}} \\
Chemistry & \textbf{25.3} {\scriptsize\textcolor{black!65}{27.5$/$25.6}} & \cellcolor{cellloss!17}22.9 {\scriptsize\textcolor{black!65}{24.9$/$23.2}} & \cellcolor{cellloss!26}19.9 {\scriptsize\textcolor{black!65}{21.5$/$20.6}} & 22.4 {\scriptsize\textcolor{black!65}{35.3$/$24.0}} & \cellcolor{cellgain!21}25.6 {\scriptsize\textcolor{black!65}{36.4$/$26.6}} & \cellcolor{cellgain!43}\textbf{32.4} {\scriptsize\textcolor{black!65}{38.0$/$32.9}} & 39.8 {\scriptsize\textcolor{black!65}{42.6$/$39.9}} & \cellcolor{cellloss!13}38.3 {\scriptsize\textcolor{black!65}{38.9$/$38.3}} & \cellcolor{cellgain!20}\textbf{45.2} {\scriptsize\textcolor{black!65}{46.2$/$45.5}} \\
Law & \textbf{21.2} {\scriptsize\textcolor{black!65}{22.2$/$21.6}} & \cellcolor{cellloss!20}18.4 {\scriptsize\textcolor{black!65}{18.7$/$18.6}} & \cellcolor{cellloss!20}18.4 {\scriptsize\textcolor{black!65}{19.5$/$18.6}} & \textbf{33.0} {\scriptsize\textcolor{black!65}{34.7$/$32.6}} & \cellcolor{cellloss!20}28.4 {\scriptsize\textcolor{black!65}{28.9$/$28.7}} & \cellcolor{cellloss!24}26.7 {\scriptsize\textcolor{black!65}{27.1$/$26.7}} & 46.6 {\scriptsize\textcolor{black!65}{47.3$/$46.3}} & \cellcolor{cellloss!16}43.1 {\scriptsize\textcolor{black!65}{44.0$/$42.8}} & \cellcolor{cellgain!11}\textbf{47.0} {\scriptsize\textcolor{black!65}{47.1$/$47.3}} \\
Engineering & 32.4 {\scriptsize\textcolor{black!65}{35.3$/$32.2}} & \cellcolor{cellgain!10}\textbf{32.5} {\scriptsize\textcolor{black!65}{36.0$/$32.3}} & \cellcolor{cellloss!22}27.1 {\scriptsize\textcolor{black!65}{29.4$/$27.1}} & 23.3 {\scriptsize\textcolor{black!65}{34.8$/$25.9}} & \cellcolor{cellgain!16}25.2 {\scriptsize\textcolor{black!65}{36.4$/$26.4}} & \cellcolor{cellgain!19}\textbf{26.1} {\scriptsize\textcolor{black!65}{36.9$/$27.2}} & 47.8 {\scriptsize\textcolor{black!65}{48.8$/$47.9}} & \cellcolor{cellgain!14}50.1 {\scriptsize\textcolor{black!65}{51.5$/$50.0}} & \cellcolor{cellgain!15}\textbf{50.9} {\scriptsize\textcolor{black!65}{52.1$/$50.8}} \\
Economics & 42.4 {\scriptsize\textcolor{black!65}{43.6$/$42.2}} & \cellcolor{cellgain!13}\textbf{44.4} {\scriptsize\textcolor{black!65}{45.9$/$43.7}} & \cellcolor{cellloss!13}40.9 {\scriptsize\textcolor{black!65}{42.1$/$40.4}} & 58.7 {\scriptsize\textcolor{black!65}{61.5$/$57.8}} & \cellcolor{cellgain!14}61.8 {\scriptsize\textcolor{black!65}{63.3$/$61.2}} & \cellcolor{cellgain!15}\textbf{62.9} {\scriptsize\textcolor{black!65}{64.4$/$62.4}} & 67.7 {\scriptsize\textcolor{black!65}{69.1$/$67.2}} & \cellcolor{cellgain!11}68.8 {\scriptsize\textcolor{black!65}{69.5$/$68.8}} & \cellcolor{cellgain!12}\textbf{69.7} {\scriptsize\textcolor{black!65}{70.0$/$70.1}} \\
Health & \textbf{36.2} {\scriptsize\textcolor{black!65}{37.1$/$36.2}} & \cellcolor{cellloss!16}33.4 {\scriptsize\textcolor{black!65}{34.6$/$33.5}} & \cellcolor{cellloss!13}34.6 {\scriptsize\textcolor{black!65}{35.4$/$34.7}} & 60.3 {\scriptsize\textcolor{black!65}{61.6$/$60.5}} & \cellcolor{cellgain!12}61.8 {\scriptsize\textcolor{black!65}{62.6$/$61.8}} & \cellcolor{cellgain!13}\textbf{62.6} {\scriptsize\textcolor{black!65}{63.2$/$62.5}} & 61.4 {\scriptsize\textcolor{black!65}{64.2$/$62.2}} & \cellcolor{cellgain!11}\textbf{62.5} {\scriptsize\textcolor{black!65}{64.3$/$62.2}} & \cellcolor{cellgain!11}62.3 {\scriptsize\textcolor{black!65}{64.0$/$62.2}} \\
Psychology & \textbf{47.2} {\scriptsize\textcolor{black!65}{47.9$/$47.3}} & \cellcolor{cellloss!14}44.8 {\scriptsize\textcolor{black!65}{46.0$/$44.6}} & \cellcolor{cellloss!11}46.5 {\scriptsize\textcolor{black!65}{47.3$/$46.7}} & 69.2 {\scriptsize\textcolor{black!65}{72.8$/$68.6}} & \cellcolor{cellloss!13}66.6 {\scriptsize\textcolor{black!65}{67.7$/$66.5}} & \cellcolor{cellgain!10}\textbf{69.3} {\scriptsize\textcolor{black!65}{70.8$/$69.2}} & 74.7 {\scriptsize\textcolor{black!65}{76.4$/$74.5}} & \cellcolor{cellgain!11}\textbf{76.1} {\scriptsize\textcolor{black!65}{77.6$/$75.9}} & \cellcolor{cellloss!13}72.0 {\scriptsize\textcolor{black!65}{73.0$/$72.1}} \\
Business & 26.2 {\scriptsize\textcolor{black!65}{28.4$/$26.9}} & \cellcolor{cellgain!18}\textbf{29.0} {\scriptsize\textcolor{black!65}{32.5$/$28.7}} & \cellcolor{cellloss!21}22.3 {\scriptsize\textcolor{black!65}{25.1$/$22.0}} & 29.0 {\scriptsize\textcolor{black!65}{38.9$/$29.3}} & \cellcolor{cellloss!13}27.8 {\scriptsize\textcolor{black!65}{34.9$/$26.9}} & \cellcolor{cellgain!39}\textbf{40.0} {\scriptsize\textcolor{black!65}{43.8$/$40.1}} & 41.7 {\scriptsize\textcolor{black!65}{46.8$/$41.3}} & \cellcolor{cellloss!15}38.7 {\scriptsize\textcolor{black!65}{39.6$/$39.2}} & \cellcolor{cellgain!30}\textbf{53.0} {\scriptsize\textcolor{black!65}{55.4$/$53.3}} \\
Biology & \textbf{55.6} {\scriptsize\textcolor{black!65}{56.5$/$55.5}} & \cellcolor{cellloss!13}53.2 {\scriptsize\textcolor{black!65}{53.8$/$53.2}} & \cellcolor{cellloss!16}51.5 {\scriptsize\textcolor{black!65}{51.6$/$51.6}} & \textbf{76.8} {\scriptsize\textcolor{black!65}{78.4$/$77.1}} & \cellcolor{cellloss!12}74.7 {\scriptsize\textcolor{black!65}{76.5$/$74.9}} & \cellcolor{cellloss!10}76.3 {\scriptsize\textcolor{black!65}{77.3$/$76.9}} & \textbf{82.5} {\scriptsize\textcolor{black!65}{82.7$/$83.6}} & \cellcolor{cellloss!14}78.4 {\scriptsize\textcolor{black!65}{78.8$/$79.7}} & \cellcolor{cellloss!14}77.8 {\scriptsize\textcolor{black!65}{79.2$/$79.0}} \\
Philosophy & 24.1 {\scriptsize\textcolor{black!65}{25.2$/$24.5}} & \cellcolor{cellgain!28}\textbf{29.7} {\scriptsize\textcolor{black!65}{30.8$/$30.0}} & \cellcolor{cellgain!12}24.7 {\scriptsize\textcolor{black!65}{25.5$/$25.1}} & 47.2 {\scriptsize\textcolor{black!65}{50.3$/$48.9}} & \cellcolor{cellgain!15}\textbf{50.5} {\scriptsize\textcolor{black!65}{53.0$/$51.1}} & \cellcolor{cellgain!15}50.2 {\scriptsize\textcolor{black!65}{52.8$/$50.4}} & 50.8 {\scriptsize\textcolor{black!65}{52.0$/$52.9}} & \cellcolor{cellgain!12}52.4 {\scriptsize\textcolor{black!65}{53.6$/$52.8}} & \cellcolor{cellgain!14}\textbf{53.5} {\scriptsize\textcolor{black!65}{56.1$/$54.6}} \\
Computer Science & 30.7 {\scriptsize\textcolor{black!65}{32.2$/$31.2}} & \cellcolor{cellgain!25}\textbf{36.6} {\scriptsize\textcolor{black!65}{40.4$/$35.7}} & \cellcolor{cellloss!20}26.6 {\scriptsize\textcolor{black!65}{28.0$/$26.8}} & 45.5 {\scriptsize\textcolor{black!65}{51.6$/$45.5}} & \cellcolor{cellloss!11}44.9 {\scriptsize\textcolor{black!65}{52.5$/$44.8}} & \cellcolor{cellgain!30}\textbf{57.8} {\scriptsize\textcolor{black!65}{60.7$/$58.5}} & 54.8 {\scriptsize\textcolor{black!65}{58.3$/$54.3}} & \cellcolor{cellgain!15}58.2 {\scriptsize\textcolor{black!65}{60.9$/$58.0}} & \cellcolor{cellgain!23}\textbf{64.1} {\scriptsize\textcolor{black!65}{67.3$/$63.6}} \\
History & \textbf{28.4} {\scriptsize\textcolor{black!65}{29.5$/$28.8}} & \cellcolor{cellloss!26}22.4 {\scriptsize\textcolor{black!65}{23.2$/$22.2}} & \cellcolor{cellloss!15}26.7 {\scriptsize\textcolor{black!65}{27.8$/$26.8}} & \textbf{56.4} {\scriptsize\textcolor{black!65}{64.4$/$58.9}} & \cellcolor{cellloss!28}42.9 {\scriptsize\textcolor{black!65}{46.6$/$47.0}} & \cellcolor{cellloss!24}46.2 {\scriptsize\textcolor{black!65}{51.9$/$50.1}} & \textbf{59.7} {\scriptsize\textcolor{black!65}{64.8$/$61.5}} & \cellcolor{cellloss!20}51.6 {\scriptsize\textcolor{black!65}{52.8$/$55.2}} & \cellcolor{cellloss!22}50.2 {\scriptsize\textcolor{black!65}{52.3$/$52.5}} \\
Other & 29.3 {\scriptsize\textcolor{black!65}{30.7$/$29.3}} & \cellcolor{cellgain!17}\textbf{32.2} {\scriptsize\textcolor{black!65}{33.3$/$32.0}} & \cellcolor{cellgain!13}30.6 {\scriptsize\textcolor{black!65}{31.3$/$30.7}} & 54.4 {\scriptsize\textcolor{black!65}{57.5$/$54.1}} & \cellcolor{cellgain!12}55.9 {\scriptsize\textcolor{black!65}{57.1$/$55.9}} & \cellcolor{cellgain!12}\textbf{56.1} {\scriptsize\textcolor{black!65}{56.5$/$56.8}} & 56.1 {\scriptsize\textcolor{black!65}{57.1$/$56.4}} & \cellcolor{cellgain!14}59.4 {\scriptsize\textcolor{black!65}{60.8$/$59.0}} & \cellcolor{cellgain!21}\textbf{64.4} {\scriptsize\textcolor{black!65}{65.4$/$64.2}} \\
\bottomrule
\end{tabular}
\end{adjustbox}

{\footnotesize Macro-F$_1$ is not bounded by accuracy: it lies above the corresponding accuracy in \Cref{tab:mmlu-pro} in 20 of the 126 cells, by at most 0.9 points.}
\end{table*}

\paragraph{Macro-averaged scores.} \Cref{tab:mmlu-pro-macro} reads the same predictions by answer class, so a condition that uses only a few of the ten answer classes is charged for the ones it leaves empty. The best condition by macro-F$_1$ is also the best by accuracy in 37 of the 42 subject--model pairs. Three of the five exceptions are ties in accuracy. The other two are health for gpt-4o-mini and for gemini-2.5-flash-lite, where all three conditions lie within 2.3 points of each other by either measure. The ordering of the conditions therefore holds in aggregate, and the gains and losses on this benchmark are not an artifact of counting correct answers wherever they fall.

\begin{figure}[t]
\centering
\noindent\rule{\linewidth}{0.8pt}\\[-1pt]
\textcolor{bad}{\textbf{Lost}}\hfill $T^0$ \textcolor{good}{\checkmark}\quad $T^1$ \textcolor{bad}{\texttimes}\quad $T^1,c{=}4$ \textcolor{bad}{\texttimes}\\[-4pt]
\noindent\rule{\linewidth}{0.3pt}
\small
\textbf{Question.} The Shang Dynasty laid the foundation for a coherent development of Chinese civilization that lasted well into the:
\begin{itemize}\itemsep=0pt\topsep=2pt\parsep=0pt\partopsep=0pt\leftmargin=1.2em
\item (0) 17th century.
\item (1) 14th century. \hfill $T^1$
\item (2) 18th century.
\item (3) 21st century.
\item (4) 5th century.
\item (5) 13th century.
\item \textbf{(6) 20th century. \hfill \textcolor{good}{\checkmark} $T^0$}
\item (7) 8th century.
\item (8) 15th century.
\item (9) 10th century.
\end{itemize}
\textbf{Summary.} \emph{To solve the problem of identifying the period in Chinese history that reflects the lasting influence of the Shang Dynasty, focus on the principle of historical continuity. The Shang Dynasty, known for its advancements in various fields, laid the groundwork for subsequent developments in Chinese civilization. The critical constraint is to \textcolor{bad}{\textbf{select a dynasty or period that directly follows the Shang}} and demonstrates a clear connection to its cultural and political legacy. Be cautious of distractors, such as the Qin or Han Dynasties, which, while significant, may not represent a direct continuation of Shang influence. Additionally, avoid periods of fragmentation or foreign rule, as these do not align with the idea of a seamless transition in Chinese history. By concentrating on the direct successors of the Shang Dynasty, you can accurately identify the period that embodies its enduring impact.}
\noindent\rule{\linewidth}{0.8pt}
\normalsize
\vspace{6pt}
\noindent\rule{\linewidth}{0.8pt}\\[-1pt]
\textcolor{good}{\textbf{Gained}}\hfill $T^0$ \textcolor{bad}{\texttimes}\quad $T^1$ \textcolor{good}{\checkmark}\quad $T^1,c{=}4$ \textcolor{good}{\checkmark}\\[-4pt]
\noindent\rule{\linewidth}{0.3pt}
\small
\textbf{Question.} A more accurate calibration curve and conversion for radiocarbon dating can be derived from:
\begin{itemize}\itemsep=0pt\topsep=2pt\parsep=0pt\partopsep=0pt\leftmargin=1.2em
\item (0) coins recovered from ancient shipwrecks.
\item \textbf{(1) dendrochronology. \hfill \textcolor{good}{\checkmark} $T^1$}
\item (2) all of the above. \hfill $T^0$
\item (3) ice core sampling.
\item (4) written records.
\item (5) bone dating of ancient human remains.
\item (6) examination of sediment layers.
\item (7) satellite imagery.
\item (8) fossil records.
\item (9) carbon dating of historical buildings.
\end{itemize}
\textbf{Summary.} \emph{To derive a more accurate calibration curve for radiocarbon dating, it's essential to focus on the relationship between the precision of these curves and the reliability of age estimations. The governing principle is that \textcolor{good}{\textbf{a refined calibration curve, which considers historical variations in atmospheric carbon levels}}, directly improves the accuracy of radiocarbon dating results. Critical constraints include the limitations of current dating techniques and potential measurement errors, which necessitate ongoing research and validation of methods. Distractors, such as unrelated environmental factors or overly simplistic approaches, can mislead problem solvers by diverting attention from scientifically validated advancements that specifically enhance calibration accuracy. By concentrating on established techniques and recent improvements in data analysis, one can effectively address the challenge of refining calibration curves for more reliable radiocarbon dating.}
\noindent\rule{\linewidth}{0.8pt}
\normalsize
\caption{\textbf{History: one lost and one gained question} (gpt-4o-mini, MMLU-Pro). In the option lists, \textcolor{good}{\checkmark} marks the correct option and $T^0$, $T^1$ mark the option each condition chose. The highlighted phrase is the claim in the summary that best explains the outcome.}
\label{fig:mmlupro-examples-history}
\end{figure}

\begin{figure}[t]
\centering
\noindent\rule{\linewidth}{0.8pt}\\[-1pt]
\textcolor{bad}{\textbf{Lost}}\hfill $T^0$ \textcolor{good}{\checkmark}\quad $T^1$ \textcolor{bad}{\texttimes}\quad $T^1,c{=}4$ \textcolor{bad}{\texttimes}\\[-4pt]
\noindent\rule{\linewidth}{0.3pt}
\small
\textbf{Question.} The most likely contract to be classified under the Uniform Commercial Code (UCC) is a contract for
\begin{itemize}\itemsep=0pt\topsep=2pt\parsep=0pt\partopsep=0pt\leftmargin=1.2em
\item (0) The purchase of a commercial property building.
\item (1) A contract for hiring a live-in nanny.
\item (2) The purchase of stocks and bonds.
\item \textbf{(3) Crops and timber to be severed from the property next summer. \hfill \textcolor{good}{\checkmark} $T^0$}
\item (4) A common carrier delivering a new computer.
\item (5) The lease of a residential property.
\item (6) The sale of a personal vehicle. \hfill $T^1$
\item (7) The sale of an intangible asset.
\item (8) The contract for a home renovation.
\item (9) A contract for personal services.
\end{itemize}
\textbf{Summary.} \emph{To determine which contract is most likely classified under the Uniform Commercial Code (UCC), it's essential to understand that \textcolor{bad}{\textbf{the UCC specifically governs contracts for the sale of tangible goods}}. This means that any contract involving services, real estate, or intangible assets, such as intellectual property, does not fall under the UCC's jurisdiction. A common pitfall for solvers is to be drawn to these non-goods options due to their familiarity, but it's crucial to focus on the UCC's clear definition of "goods" as tangible items that can be moved. By concentrating on contracts that involve the sale of physical products, solvers can effectively identify the correct answer while avoiding distractions from irrelevant contract types.}
\noindent\rule{\linewidth}{0.8pt}
\normalsize
\vspace{6pt}
\noindent\rule{\linewidth}{0.8pt}\\[-1pt]
\textcolor{good}{\textbf{Gained}}\hfill $T^0$ \textcolor{bad}{\texttimes}\quad $T^1$ \textcolor{good}{\checkmark}\quad $T^1,c{=}4$ \textcolor{good}{\checkmark}\\[-4pt]
\noindent\rule{\linewidth}{0.3pt}
\small
\textbf{Question.} A defendant was charged with felony murder as a result of his setting fire to the victim's house. The victim was upstairs sleeping when the fire was set, and he died in his bedroom during the fire. If the defendant can prove the facts to support his defense, which of the following assertions would LEAST likely remove liability for felony murder?
\begin{itemize}\itemsep=0pt\topsep=2pt\parsep=0pt\partopsep=0pt\leftmargin=1.2em
\item (0) The defendant was coerced by another to set the fire.
\item \textbf{(1) The defendant did not intend to kill the victim. \hfill \textcolor{good}{\checkmark} $T^1$}
\item (2) The defendant thought the house was empty when he set the fire.
\item (3) The defendant believed the victim had already evacuated the house.
\item (4) The defendant was insane when he set the fire.
\item (5) The victim died of a heart attack before the fire spread to his bedroom. \hfill $T^0$
\end{itemize}
\textbf{Summary.} \emph{In this scenario, the key principle governing liability for felony murder is the connection between the defendant's actions--specifically, committing a dangerous felony like arson--and the resulting death, \textcolor{good}{\textbf{regardless of whether the defendant intended to kill}}. Critical constraints include the legal definitions of felony murder, which emphasize the inherent danger of the act rather than the defendant's subjective intent or beliefs about the victim's presence. Distractors may lead solvers to focus on the defendant's lack of intent to kill or his assumption that the victim was not home, which could misguide them into thinking these factors absolve him of liability. However, the law holds that engaging in a dangerous act that results in death can still lead to felony murder charges, irrespective of the defendant's personal beliefs or intentions.}
\noindent\rule{\linewidth}{0.8pt}
\normalsize
\caption{\textbf{Law: one lost and one gained question} (gpt-4o-mini, MMLU-Pro).}
\label{fig:mmlupro-examples-law}
\end{figure}

\begin{figure}[t]
\centering
\noindent\rule{\linewidth}{0.8pt}\\[-1pt]
\textcolor{bad}{\textbf{Lost}}\hfill $T^0$ \textcolor{good}{\checkmark}\quad $T^1$ \textcolor{bad}{\texttimes}\quad $T^1,c{=}4$ \textcolor{good}{\checkmark}\\[-4pt]
\noindent\rule{\linewidth}{0.3pt}
\small
\textbf{Question.} Of the following, most children will develop which skill first?
\begin{itemize}\itemsep=0pt\topsep=2pt\parsep=0pt\partopsep=0pt\leftmargin=1.2em
\item \textbf{(0) clap their hands \hfill \textcolor{good}{\checkmark} $T^0$}
\item (1) cut with a knife
\item (2) tie their shoes
\item (3) ride a bicycle
\item (4) read a book
\item (5) solve a math problem
\item (6) say a sentence \hfill $T^1$
\item (7) draw a picture
\item (8) play a musical instrument
\item (9) write with a pencil
\end{itemize}
\textbf{Summary.} \emph{To solve the problem of which skill most children will develop first, we rely on the established understanding of developmental milestones in child growth. Typically, \textcolor{bad}{\textbf{gross motor skills, such as crawling and walking, emerge before language skills}} and social interactions. It's important to consider that individual experiences and developmental delays can affect the timing of these skills, which introduces variability. Common misconceptions may lead some to believe that skills like walking or talking are acquired at the same age for all children, but this is not supported by research. By focusing on the predictable order of skill acquisition and recognizing these misconceptions, we can effectively evaluate the options and identify the skill that most children are likely to develop first.}
\noindent\rule{\linewidth}{0.8pt}
\normalsize
\vspace{6pt}
\noindent\rule{\linewidth}{0.8pt}\\[-1pt]
\textcolor{good}{\textbf{Gained}}\hfill $T^0$ \textcolor{bad}{\texttimes}\quad $T^1$ \textcolor{good}{\checkmark}\quad $T^1,c{=}4$ \textcolor{bad}{\texttimes}\\[-4pt]
\noindent\rule{\linewidth}{0.3pt}
\small
\textbf{Question.} Alpha waves are most closely associated with
\begin{itemize}\itemsep=0pt\topsep=2pt\parsep=0pt\partopsep=0pt\leftmargin=1.2em
\item (0) Stage 4 sleep
\item (1) Stage 1 sleep \hfill $T^0$
\item (2) REM sleep
\item (3) Stage 3 sleep
\item (4) Deep sleep
\item (5) Stage 2 sleep
\item (6) Lucid dreaming
\item \textbf{(7) the hypnagogic state \hfill \textcolor{good}{\checkmark} $T^1$}
\item (8) Insomnia
\item (9) Sleep walking
\end{itemize}
\textbf{Summary.} \emph{To solve the problem regarding alpha waves, focus on their established \textcolor{good}{\textbf{association with relaxation and calmness}}, as they are brainwaves measured in the frequency range of 8-12 Hz. It's important to avoid common distractors that link alpha waves to states like deep sleep or intense concentration, which are actually associated with delta and beta waves, respectively. Additionally, terms suggesting alertness or active thinking can mislead, as they conflict with the calming nature of alpha waves. By concentrating on the correct mental states tied to alpha waves and recognizing these potential traps, you can accurately identify the right answer that reflects their true psychological implications.}
\noindent\rule{\linewidth}{0.8pt}
\normalsize
\caption{\textbf{Psychology: one lost and one gained question} (gpt-4o-mini, MMLU-Pro).}
\label{fig:mmlupro-examples-psychology}
\end{figure}

\begin{figure}[t]
\centering
\noindent\rule{\linewidth}{0.8pt}\\[-1pt]
\textcolor{bad}{\textbf{Lost}}\hfill $T^0$ \textcolor{good}{\checkmark}\quad $T^1$ \textcolor{bad}{\texttimes}\quad $T^1,c{=}4$ \textcolor{bad}{\texttimes}\\[-4pt]
\noindent\rule{\linewidth}{0.3pt}
\small
\textbf{Question.} Eighty-three monozygotic twin pairs are examined for the presence of club feet. Fifty pairs do not show the trait at all; twenty pairs have one member with club feet; and thirteen pairs have both members affected. What is the frequency of concordance?
\begin{itemize}\itemsep=0pt\topsep=2pt\parsep=0pt\partopsep=0pt\leftmargin=1.2em
\item (0) 60 percent
\item (1) 42 percent \hfill $T^1$
\item (2) 20 percent
\item (3) 50 percent
\item (4) 75 percent
\item (5) 24 percent
\item \textbf{(6) 39 percent \hfill \textcolor{good}{\checkmark} $T^0$}
\item (7) 33 percent
\item (8) 15.5 percent
\item (9) 13 percent
\end{itemize}
\textbf{Summary.} \emph{To solve the problem of determining the frequency of concordance for club feet among monozygotic twin pairs, we need to focus on the definition of concordance, which is the proportion of twin pairs where both members exhibit the trait. We have a total of 83 twin pairs, with 13 pairs having both twins affected by club feet. The critical constraint is to ensure we calculate the concordance based on pairs, not individuals, as this distinction is essential for accuracy. A common distraction is the temptation to focus on the number of affected individuals or to misinterpret the data by overlooking the pairs with neither or only one twin affected. By correctly identifying the relevant data points--specifically \textcolor{bad}{\textbf{the number of concordant pairs (13) and the total pairs (83)}}--we can accurately compute the concordance frequency, which reflects the genetic relationship and likelihood of sharing the trait among monozygotic twins.}
\noindent\rule{\linewidth}{0.8pt}
\normalsize
\vspace{6pt}
\noindent\rule{\linewidth}{0.8pt}\\[-1pt]
\textcolor{good}{\textbf{Gained}}\hfill $T^0$ \textcolor{bad}{\texttimes}\quad $T^1$ \textcolor{good}{\checkmark}\quad $T^1,c{=}4$ \textcolor{good}{\checkmark}\\[-4pt]
\noindent\rule{\linewidth}{0.3pt}
\small
\textbf{Question.} If, on average, 46\% of the loci in a species' gene pool are heterozygous, then the average homozygosity of the species should be
\begin{itemize}\itemsep=0pt\topsep=2pt\parsep=0pt\partopsep=0pt\leftmargin=1.2em
\item \textbf{(0) 54\% \hfill \textcolor{good}{\checkmark} $T^1$}
\item (1) 36\%
\item (2) 76\%
\item (3) 100\%
\item (4) 92\%
\item (5) 64\% \hfill $T^0$
\item (6) 46\%
\item (7) 56\%
\item (8) 86\%
\item (9) 23\%
\end{itemize}
\textbf{Summary.} \emph{To solve the problem of determining the average homozygosity in a species' gene pool when 46\% of the loci are heterozygous, we can rely on the fundamental relationship between heterozygosity and homozygosity, which states that their percentages must add up to 100\%. This means that if we know the percentage of heterozygous loci, we can easily find the percentage of homozygous loci by \textcolor{good}{\textbf{subtracting the heterozygosity percentage from 100\%}}. The critical constraint here is that both values must be expressed as percentages and remain non-negative. A common distraction might be the temptation to overcomplicate the problem by considering additional factors or nuances in allele distributions, but this is unnecessary. The straightforward calculation of subtracting 46\% from 100\% leads directly to the solution, emphasizing the simplicity of the relationship between these two genetic concepts.}
\noindent\rule{\linewidth}{0.8pt}
\normalsize
\caption{\textbf{Biology: one lost and one gained question} (gpt-4o-mini, MMLU-Pro).}
\label{fig:mmlupro-examples-biology}
\end{figure}

\section{Causal Judgment as a Binary Decision}
\label{sec:appendix-cj}

Causal judgment is the one benchmark in \Cref{tab:main} with two answers, so the false positive rate is defined on it without an arbitrary choice of positive class. We take \textit{Yes} as positive. The gold split is 97 \textit{Yes} against 90 \textit{No}, which is close to balanced.

\begin{table}[h]
\centering
\caption{\textbf{Causal judgment, read as a binary decision.} \textit{Yes} is the positive class, over 187 questions with 97 of them positive. TPR is recall on \textit{Yes} and FPR is the share of \textit{No} questions answered \textit{Yes}; both are percentages. The column \#Y is how many times the condition answered \textit{Yes} at all. A reply that is neither \textit{Yes} nor \textit{No} (three refusals by Llama3.2-3b, two at $T^1$ and one at $T^1,c{=}4$) counts as a wrong answer and as not \textit{Yes}. Shading compares a condition against $T^0$ in the same model, green where TPR rises or FPR falls and red where TPR falls or FPR rises.}
\label{tab:cj-binary}
\begin{adjustbox}{max width=\linewidth}
\footnotesize
\begin{tabular}{@{}lrrrrrrrrr@{}}
\toprule
 & \multicolumn{3}{c}{$T^0$} & \multicolumn{3}{c}{$T^1$} & \multicolumn{3}{c}{$T^1, c=4$} \\
\cmidrule(lr){2-4}\cmidrule(lr){5-7}\cmidrule(lr){8-10}
 & TPR & FPR & \#Y & TPR & FPR & \#Y & TPR & FPR & \#Y \\ \midrule
Qwen2.5-3b & 9.3 & 7.8 & 16 & \cellcolor{cellgain!16}12.4 & \cellcolor{cellgain!17}4.4 & 16 & \cellcolor{cellgain!29}18.6 & \cellcolor{cellgain!19}3.3 & 21 \\
Llama3.2-3b & 11.3 & 3.3 & 14 & \cellcolor{cellgain!12}12.4 & \cellcolor{cellloss!12}4.4 & 16 & \cellcolor{cellgain!26}19.6 & \cellcolor{cellloss!17}6.7 & 25 \\
\bottomrule
\end{tabular}
\end{adjustbox}
\end{table}

Both models sit close to a constant \textit{No} classifier. At $T^0$, Llama3.2-3b answers \textit{Yes} 14 times in 187 questions and Qwen2.5-3b 16 times, against 97 questions where \textit{Yes} is correct. This is why direct answering scores near 50 on a near-balanced task.

From $T^0$ to $T^1,c{=}4$ the meta-controller raises the true positive rate in both models: from 9.3 to 18.6 for Qwen2.5-3b and from 11.3 to 19.6 for Llama3.2-3b. The false positive rate moves differently. For Qwen2.5-3b it falls at the same time, from 7.8 to 3.3, so the change is not a threshold shift, which would trade one rate against the other; the extra \textit{Yes} answers are mostly the right ones. This fits the die-and-bomb example in \S\ref{sec:results}: the meta-steered summary keeps the agent's intention in view, and intention is what makes a \textit{Yes} defensible. For Llama3.2-3b the false positive rate rises from 3.3 to 6.7. It answers \textit{Yes} 25 times instead of 14: correct \textit{Yes} answers rise from 11 to 19 and wrong ones from 3 to 6. For this model, part of the gain is a shift toward \textit{Yes}.

\section{Hi-ToM Discussion}
\label{sec:hitom-qualitative}

\begin{figure}[t]
\centering
\includegraphics[width=\textwidth]{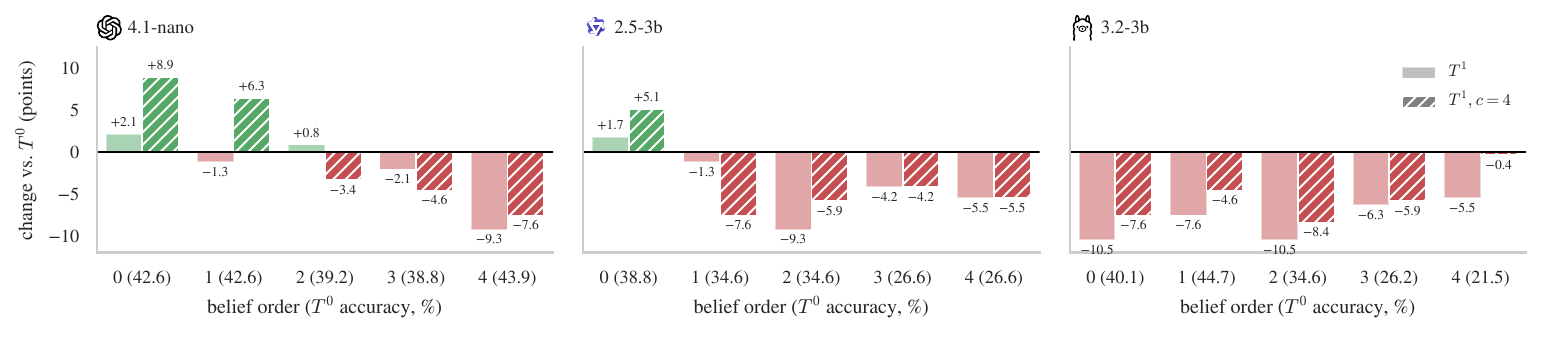}
\caption{\textbf{Hi-ToM accuracy by belief order.} Each pair of bars shows the change in accuracy against $T^0$. The horizontal axis is the belief order, with the $T^0$ accuracy at that order in brackets.}
\label{fig:hitom-orders}
\end{figure}

\definecolor{hitomblue}{HTML}{4C72B0}
\definecolor{hitomred}{HTML}{C44E52}
\definecolor{hitommuted}{HTML}{555555}
\definecolor{hitomgreen}{HTML}{2E7D32}

Direct answering and the full meta-steered pipeline reach the same count of
correct answers on Hi-ToM with gpt-4.1-nano: 490 of 1{,}185 items under
each (direct: run \texttt{hitom\_20261012\_175700}; $T^1,c{=}4$: run
\texttt{hitom\_20261006\_225845}; the $T^0$ cell of \Cref{tab:main} comes
from a later direct run of the same configuration with 491 correct, also
41.4\%). We first rule
out a trivial explanation for this coincidence, then identify the actual
source.

\begin{figure*}[t]
\centering
\includegraphics[width=\textwidth]{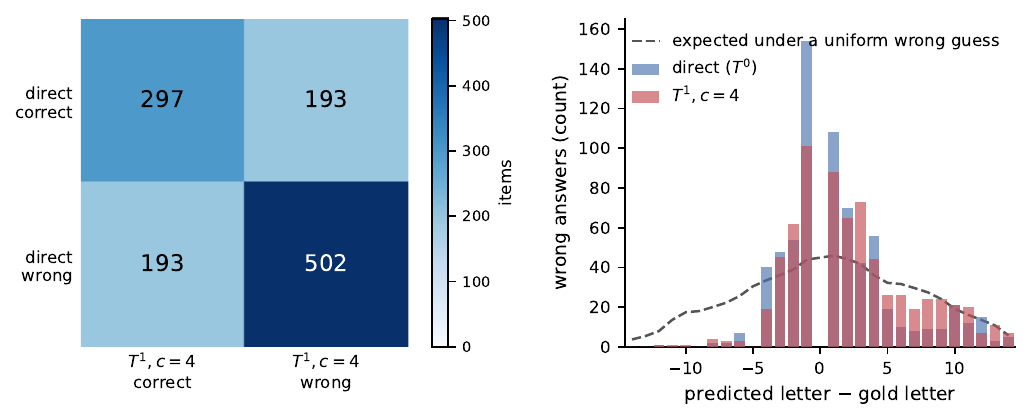}
\caption{\textbf{Left:} item-level agreement between direct and the full
meta-steered pipeline ($T^1,c{=}4$) on Hi-ToM. Only 297 of the 490 items each
condition answers correctly are shared; 193 items are correct under direct
only, and a different 193 under $T^1,c{=}4$ only. Final answers are never
cached (only intermediate reflexion summaries are), so every entry in this
matrix reflects an independent model call. \textbf{Right:} distribution of
predicted-minus-gold letter distance for both conditions, each over its own incorrect items, against the distribution expected under a uniform
wrong guess over the remaining fourteen choices.}
\label{fig:hitom-tie-evidence}
\end{figure*}

\Cref{fig:hitom-tie-evidence} (left) shows that the two conditions do not
answer the same 490 items. \Cref{fig:hitom-tie-evidence} (right) identifies
a mechanism both conditions share, to different degrees: the predicted
label is exactly one position away from gold in 262 of the 695 incorrect
direct answers (38\%, a 2.9\texttimes{} enrichment over the 89.8 expected by
chance) and in 189 of $T^1,c{=}4$'s 695 incorrect answers (27\%, a
2.1\texttimes{} enrichment over the 89.9 expected). The effect is weaker
for $T^1,c{=}4$ but not absent, consistent with both conditions drawing on
the same underlying fifteen-way tracking task rather than two unrelated
error processes.

\begin{figure}[t]
\centering
\includegraphics[width=0.85\columnwidth]{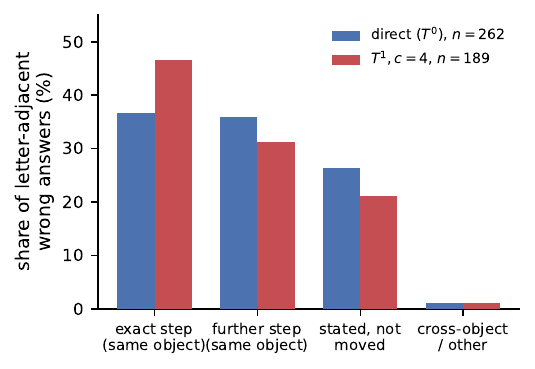}
\caption{What a letter-adjacent wrong answer actually is, for direct
($n{=}262$) and $T^1,c{=}4$ ($n{=}189$): the predicted container's position
relative to gold within the target object's own parsed move history.}
\label{fig:hitom-chain}
\end{figure}

Letter adjacency is itself a proxy, not the mechanism. For every
letter-adjacent wrong answer under each condition, we parse the target
object's own move history directly from the raw story (every ``moved the
\textit{object} to the \textit{container}'' statement, plus its initial
location) and locate gold and the predicted container within that history.
\Cref{fig:hitom-chain} shows the result: the large majority of these errors,
for both conditions, place the predicted container somewhere in the
same object's own tracked history, and a substantial share place it exactly
one recorded step away from gold (47\% for $T^1,c{=}4$, 37\% for direct).
Cross-object contamination, the mechanism behind the worked example below,
accounts for at most 1\% of letter-adjacent errors in either condition: it
is real, but rare. The two conditions fail for the same underlying reason.
Both identify the correct tracked object almost every time and then select
the wrong point in its recorded history, rather than either hallucinating
an unrelated container or confusing it with a different object.

\begin{figure}[t]
\centering
\includegraphics[width=\columnwidth]{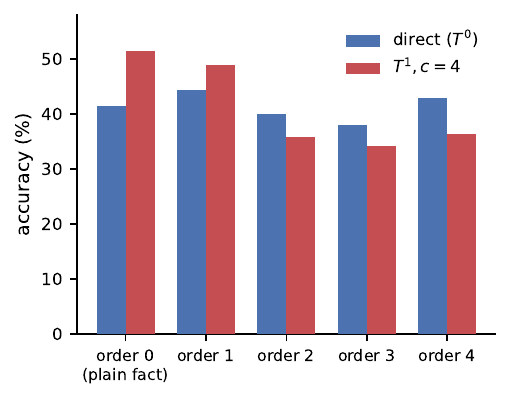}
\caption{Accuracy by belief order, direct vs.\ the full meta-steered pipeline ($T^1,c{=}4$). Direct is flat across order; $T^1,c{=}4$ trades accuracy instead, ahead of direct at orders 0--1 and behind it at orders 2--4.}
\label{fig:hitom-ceiling}
\end{figure}

For direct, this effect is independent of belief-nesting depth.
\Cref{fig:hitom-ceiling} (blue) shows accuracy flat across belief order
(38.0--44.3\%), including order 0, a plain fact requiring no false-belief
reasoning at all. The bottleneck for gpt-4.1-nano reading the raw story is
tracking one object through fifteen similarly-described candidates across
a long narrative, not the depth of theory-of-mind reasoning required.
$T^1,c{=}4$ (\Cref{fig:hitom-ceiling}, red) trades accuracy across the
order axis instead: it exceeds direct at order 0 and order 1 (51.5\% and
48.9\%, against 41.4\% and 44.3\%) and falls below it at orders 2 through 4
(35.9\%, 34.2\%, and 36.3\%, against 40.1\%, 38.0\%, and 43.0\%). The two
representations are not hitting one shared, order-independent ceiling: the
raw story's difficulty is uniform across depth, while the reflexion
summary's is not.

An order-$n$ Hi-ToM question resolves to whichever single character's
last-known state the $n$-hop belief chain collapses to, typically whoever
exited the scene earliest, before the information the deeper nesting asks
about was available. Answering it correctly requires tracking exactly who
was present for which event, hop by hop. In the order-2-to-4 items we
inspected by hand where $T^1,c{=}4$ answers incorrectly (a small,
non-random set), the predicted container was typically neither a
hallucination nor a different object's container but some other
character's last-known location. The raw numbered story marks presence
with an explicit entered/exited line; the reflexion summary's continuous
prose does not keep it separately addressable. This is a cost of our
integration step ($T$, \Cref{eq:reflection}, \Cref{sec:gunther}):
synthesizing the three perspectives into one account recovers facts a
single reading misses on the state-tracking tasks (\Cref{sec:results}), but
is not built to preserve discrete, per-character presence bookkeeping. We
return to this trade-off in \Cref{sec:conclusions}. Averaged over all five
orders, these opposite-signed,
order-dependent effects produce the near-identical aggregate accuracy and
low item-level overlap in \Cref{fig:hitom-tie-evidence}.

On the meta-steering side, accuracy across cycle budgets is non-monotone.
For gpt-4.1-nano it is 39.5\% after one reflexion pass ($T^1$), 38.6\% at
$c{=}1$, 40.4\% at $c{=}2$, 40.1\% at $c{=}3$, and 41.4\% at $c{=}4$, which
is back at $T^0$'s level (41.4\%). Cycles are sequential: each one is built
from the previous cycle's summary and the judges' feedback on it. In this
run the budgets are also chained (\Cref{sec:appendix-repro}): each budget
from $c{=}2$ to $c{=}4$ continues from the final summary of the budget
below it, runs one new cycle, and keeps whichever of the two has more
positive verdicts. Moving from one budget to the next therefore adds a
single candidate and a selection step, which fits the small movements in
both directions better than a steady accumulation of corrections would.
After the examples we look at why that selection criterion is an imperfect
proxy for correctness.

The two items below, one from each side of the flip in
\Cref{fig:hitom-tie-evidence}, show two different failure modes. The first
is not the character-attribution error described above: the summary merges
two tracked objects.

\begin{figure}[t]
\centering
\begin{tcolorbox}[colback=hitomred!6, colframe=hitomred!70!black, boxrule=0.5pt, arc=2pt,
  left=5pt, right=5pt, top=3pt, bottom=3pt, fonttitle=\bfseries\footnotesize,
  title={$T^0$ \textcolor{hitomgreen}{\checkmark}, $T^1,c{=}4$ \textcolor{hitomred}{\texttimes}}]
\footnotesize
\textbf{Question:} ``Where does Benjamin really think the \emojiimg{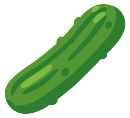} is?''\\[3pt]
\rule{\columnwidth}{0.4pt}
\textbf{Story:} Benjamin, Charlotte, Isla, Ava, and Evelyn entered the workshop. 

\textcolor{orange}{(1) Benjamin dislikes the \emojiimg{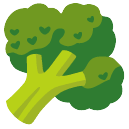}.}

\textcolor{hitomblue}{(2) The \emojiimg{figures/emoji_cucumber.png} is in the \emojiimg{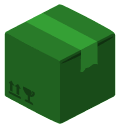}}

(3) Isla likes the \emojiimg{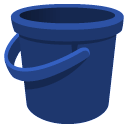}.

\textcolor{hitomblue}{(4) Benjamin moved the \emojiimg{figures/emoji_cucumber.png} to the \emojiimg{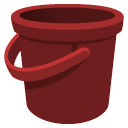}.}

\textbf{(5) Benjamin exited the workshop. 
}

\textcolor{orange}{(6) Charlotte moved the \emojiimg{figures/emoji_cucumber.png} to the \emojiimg{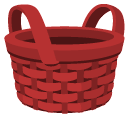}.}

(7) Charlotte exited the workshop.

(8) Isla, Ava, and Evelyn each stayed one minute and exited the workshop in turn. 

(9) Benjamin, Charlotte, Isla, Ava, and Evelyn entered the \emojiimg{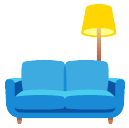}.

(10) Charlotte, Isla, and Evelyn entered the lounge. 

\textcolor{orange}{(11) The \emojiimg{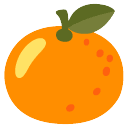} is in the \emojiimg{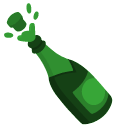}.}

(12) Ava likes the \emojiimg{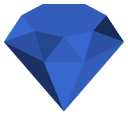}.

(13) Charlotte stayed one minute and exited the lounge. 

\textcolor{orange}{(14) Isla moved the \emojiimg{figures/emoji_tangerine.png} to the \emojiimg{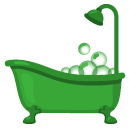}.}

\textcolor{orange}{(15) Evelyn moved the \emojiimg{figures/emoji_tangerine.png} to the \emojiimg{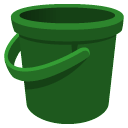}.}

(16) Charlotte, Isla, and Evelyn entered the \emojiimg{figures/emoji_waiting_room.png}.

\textcolor{orange}{(17) Charlotte publicly claimed that \emojiimg{figures/emoji_tangerine.png} is in the \emojiimg{figures/emoji_bottle_green.png}.}

(18) Benjamin likes the \emojiimg{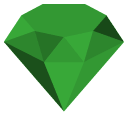}. 

\textcolor{orange}{(19) Evelyn privately told Benjamin that the \emojiimg{figures/emoji_tangerine.png} is in the \emojiimg{figures/emoji_bucket_green.png}.}\\[5pt]
\rule{\columnwidth}{0.4pt}
\\[5pt]
\begin{tabular}{@{}ll@{}}
$T^0$ $\longrightarrow$ & \emojiimg{figures/emoji_container_red.png}\quad  \textcolor{hitomgreen}{\checkmark} \\
\textbf{$T^1,c{=}4$} $\longrightarrow$ & \emojiimg{figures/emoji_bucket_green.png}\quad\textcolor{hitomred}{\texttimes} \\
\end{tabular}\\[5pt]
\rule{\linewidth}{0.4pt}
\textbf{Why they differ:} the \emojiimg{figures/emoji_cucumber.png} and the \emojiimg{figures/emoji_tangerine.png} are two separate
tracked objects. The reflexion summary folds both into one paragraph and
invents a second re-entry into the lounge not present in the story above,
ending on: \textcolor{hitomred}{\emph{``\dots Evelyn privately told \textcolor{hitomred}{\textbf{Benjamin}} that \textcolor{hitomred}{\textbf{the \emojiimg{figures/emoji_tangerine.png}}} was in the \emojiimg{figures/emoji_bucket_green.png}.''}} $T^1,c{=}4$ answers with the \emph{\emojiimg{figures/emoji_tangerine.png}}'s container, pulled in by the name match on ``Benjamin''; direct, reading the story's own thread separation above, does not make this substitution.
\end{tcolorbox}
\caption{\textbf{Hi-ToM item where direct answering is right and $T^1,c{=}4$ is wrong} (gpt-4.1-nano). The summary merges two tracked objects. Objects and containers are drawn as emoji, which stand in for the dataset's names and are not always literal (for example, an orange is drawn as a tangerine and a green box as a green crate).}
\label{fig:hitom-example-lost}
\end{figure}

\begin{figure}[t]
\centering
\begin{tcolorbox}[colback=hitomgreen!6, colframe=hitomgreen!70!black, boxrule=0.5pt, arc=2pt,
  left=5pt, right=5pt, top=3pt, bottom=3pt, fonttitle=\bfseries\footnotesize,
  title={$T^0$ \textcolor{hitomred}{\texttimes}, $T^1,c{=}4$ \textcolor{hitomgreen}{\checkmark}}]
\footnotesize
\textbf{Question:} ``Where is the \emojiimg{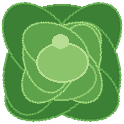} really?''\\[3pt]
\rule{\columnwidth}{0.4pt}
\textbf{Story:} Noah, Emma, Avery, Aiden, and Jackson entered the staircase.

\textcolor{hitomblue}{(1) The \emojiimg{figures/emoji_cabbage.png} is in the \emojiimg{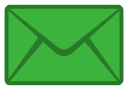}.}

\textcolor{hitomblue}{(2) Noah moved the \emojiimg{figures/emoji_cabbage.png} to the \emojiimg{figures/emoji_crate_green.png}.}

(3) Emma likes the \emojiimg{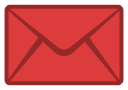}.

(4) Noah exited the staircase.

(5) Emma, Avery, Aiden, and Jackson each stayed one minute and exited the staircase in turn.

(6) Noah, Emma, Avery, Aiden, and Jackson entered the \emojiimg{figures/emoji_waiting_room.png}.

(7) Aiden likes the \emojiimg{figures/emoji_bucket_blue.png}.

(8) Aiden, Avery and Jackson entered the staircase.

\textcolor{orange}{(9) The \emojiimg{figures/emoji_cabbage.png} is in the \emojiimg{figures/emoji_crate_green.png}.}

(10) Aiden stayed one minute and exited.

\textcolor{hitomblue}{(11) Avery moved the \emojiimg{figures/emoji_cabbage.png} to the \emojiimg{figures/emoji_envelope_green.png} and exited.}

(12) Jackson stayed one minute and exited; Aiden, Avery, and Jackson entered the \emojiimg{figures/emoji_waiting_room.png}.\\[5pt]
\rule{\columnwidth}{0.4pt}
\\[5pt]
\begin{tabular}{@{}ll@{}}
\textbf{$T^0$} $\longrightarrow$ & \emojiimg{figures/emoji_crate_green.png}\quad \textcolor{hitomred}{\texttimes} \\
$T^1,c{=}4$ $\longrightarrow$ & \emojiimg{figures/emoji_envelope_green.png}\quad\textcolor{hitomgreen}{\checkmark} \\
\end{tabular}\\[5pt]
\rule{\linewidth}{0.4pt}
\textbf{Why they differ:} line (9) restates the location the \emojiimg{figures/emoji_cabbage.png} has had since the move at line (2), as scene-setting before the second block of moves; it adds no new information, and line (11) moves the object again. Direct anchors on that restated line instead of the final move at line (11). The reflexion summary, written once as a linear account, states the final move only: \textcolor{hitomgreen}{\emph{``\dots Avery moved the \emojiimg{figures/emoji_cabbage.png} back from the \emojiimg{figures/emoji_crate_green.png} to \textbf{the \emojiimg{figures/emoji_envelope_green.png}}}\dots''} -- with nothing to anchor on but the last move, $T^1,c{=}4$ answers correctly.
\end{tcolorbox}
\caption{\textbf{Hi-ToM item where direct answering is wrong and $T^1,c{=}4$ is right} (gpt-4.1-nano). Emoji stand in for the dataset's object and container names, as in \Cref{fig:hitom-example-lost}.}
\label{fig:hitom-example-gained}
\end{figure}

The first case reflects a pattern in the meta-judge's own trace. For one
item answered incorrectly under every meta-steered condition from $c{=}2$
through $c{=}4$, the \emph{objective} dimension fails at both recorded
cycles with an essentially identical rationale: the summary conflates a
character's stated belief with her own action, and conflates two parallel
tracked objects. The cycle that follows (an \textsc{Expand} if only $J_O$
fails, since two positive verdicts give $R$, and a \textsc{Refine} if a
second judge fails as well) is not a targeted edit of the flagged span. It
reruns the complete reflexion engine on a new input built from the
question, the prior answer, and the judges' rationales, producing an
entirely new paragraph. In this trace, the rewrite leaves the named
conflation intact and introduces a second, unrelated one, attributing a
belief about the other tracked object to a character never said to hold
it, while contradicting an earlier clause about her own actions. The cycle
is selected nonetheless, because selection is governed by the count of
positive judgments across three dimensions, not by whether the flagged
error was corrected.

This mechanism accounts for the non-monotone pattern across cycle budgets
above: each additional cycle rewrites the summary rather than repairing
the flagged error, and the controller keeps whichever candidate scores
highest on an imperfect three-dimension proxy. A larger budget adds
candidates to select from, which is consistent with $c{=}4$ returning to
parity with direct on average. Because each cycle is conditioned on the
previous summary and its feedback, successive cycles can carry the same
unresolved error forward, as they do in this trace, which suggests
diminishing rather than unbounded returns from further budget increases.

In summary, direct answering and the full meta-steered pipeline converge
to the same aggregate accuracy on Hi-ToM because they fail oppositely and
by comparable amounts, not because they fail identically. Direct answering
is limited by a uniform, depth-independent tracking capacity. The
reflexion summary trades accuracy across belief order: it gains where a
de-duplicated, linear retelling resolves the raw story's own redundancy,
and loses where deeper belief-nesting depends on the discrete,
per-character presence bookkeeping that synthesis into continuous prose is
not built to preserve (\Cref{sec:conclusions}). The meta-controller's
cycle selection is a real but partial corrective against its own failure
mode, not a guarantee of improvement with additional budget.

\section{Examples of the Three Divergences}
\label{sec:appendix-persp-examples}

This section shows the three comparisons of the perspective divergence test (\Cref{fig:perspective-divergence}) on real output.

\definecolor{pdivA}{HTML}{4C72B0}
\definecolor{pdivB}{HTML}{C44E52}
\definecolor{pdivmuted}{HTML}{555555}

\newcommand{\pdivbox}[2]{%
  \begin{minipage}[t]{0.487\textwidth}
  \begin{tcolorbox}[colback=#1!5, colframe=#1!65!black, boxrule=0.4pt, arc=2pt,
    left=4pt, right=4pt, top=2pt, bottom=2pt, boxsep=0pt]
  \scriptsize #2
  \end{tcolorbox}
  \end{minipage}}

\newcommand{\pdivhead}[3]{%
  \smallskip\noindent{\footnotesize\textbf{#1 $=$ #2}\;\textcolor{pdivmuted}{#3}}\par\nobreak\vspace{1.5pt}}

\begin{figure*}[t]
\centering

\pdivhead{$d_\mathrm{persp}$}{0.400}{--- same leaf, same input text; the perspective asked for is what varies}
\pdivbox{pdivA}{\textbf{$\varphi_O$:} Owen, Jacob, Emma, William, and Carter entered the sunroom. Owen moved the watermelon to the green envelope and then left the sunroom. Jacob moved the watermelon to the green bucket and then left. Emma moved the watermelon to the blue pantry and then left. [\dots]}\hfill
\pdivbox{pdivB}{\textbf{$\varphi_S$:} Owen \emph{believes} he moved the watermelon to the green envelope before leaving the sunroom. Jacob \emph{thinks} the watermelon is in the green bucket, but he dislikes grapes. Emma \emph{is convinced} it is in the blue pantry, while William \emph{believes} it is in the green basket. [\dots]}

\pdivhead{$d_\mathrm{region}$}{0.071}{--- same input text, same perspective; only the leaf's region context varies (the control)}
\pdivbox{pdivA}{\textbf{$\varphi_O$, leaf 1:} Owen, Jacob, Emma, William, and Carter entered the sunroom. Owen moved the watermelon to the green envelope and then \textbf{left} the sunroom. Jacob moved the watermelon to the green bucket and then \textbf{left}. [\dots]}\hfill
\pdivbox{pdivA}{\textbf{$\varphi_O$, leaf 2:} Owen, Jacob, Emma, William, and Carter entered the sunroom. Owen moved the watermelon to the green envelope and then \textbf{exited} the sunroom. Jacob moved the watermelon to the green bucket and then \textbf{exited}, followed by Emma. [\dots]}

\pdivhead{$d_\mathrm{item}$}{0.786}{--- same perspective, two unrelated stories (a reference level for unrelated inputs)}
\pdivbox{pdivA}{\textbf{$\varphi_O$, story A:} Owen moved the \textbf{watermelon} to the green envelope and then left the \textbf{sunroom}. Jacob moved it to the green bucket; Emma moved it to the blue pantry. [\dots]}\hfill
\pdivbox{pdivmuted}{\textbf{$\varphi_O$, story B:} The \textbf{sweet potato} was initially in the red drawer, then moved to the red crate and later to the red container. Liam, Hannah, Emma, Mila, and Evelyn entered and exited the \textbf{cellar}. [\dots]}

\caption{\textbf{What each divergence compares, on real output.} All excerpts
verbatim from depth-1 reflexion passes, truncated at [\dots]; the value above
each pair is the grounded divergence for that pair. Each output is first
reduced to the content words that occur in its own source story; for the
two unrelated stories these are different sources, so the two sets can
still share words such as names or colours, and the divergence stays below
1. Changing the region context leaves the objective rewrite almost
untouched ($0.071$: \emph{left} becomes \emph{exited}); changing the
perspective on the same input moves it nearly six times as far ($0.400$:
moves become beliefs). Two unrelated stories give $0.786$, a reference
level rather than an upper bound. Per-pair values here; dataset means in
\Cref{fig:perspective-divergence}.}
\label{fig:perspective-examples}
\end{figure*}

\section{Can G\"unther's Polyreflection Play?}
\label{sec:goalfree-play}

Motivated by~\citet{cloos2026playing}, who placed a coding assistant on an island, named no task, reward, or activity for it, and asked whether what followed satisfies the classical criteria for play~\citep{huizinga1938homo,caillois1961man}, we ask whether a reasoning procedure such as G\"unther's polyreflection does the same. In terms of the seventh law of open-ended intelligence~\citep{hegelich2026corecursive}, removing the external task removes the proxy objective from the prompt, so the router's internals run without one.

\subsection{Setup}
\label{sec:goalfree-setup}

\begin{wrapfigure}{r}{0.42\textwidth}
\vspace{-\intextsep}
\centering
\includegraphics[width=0.40\textwidth]{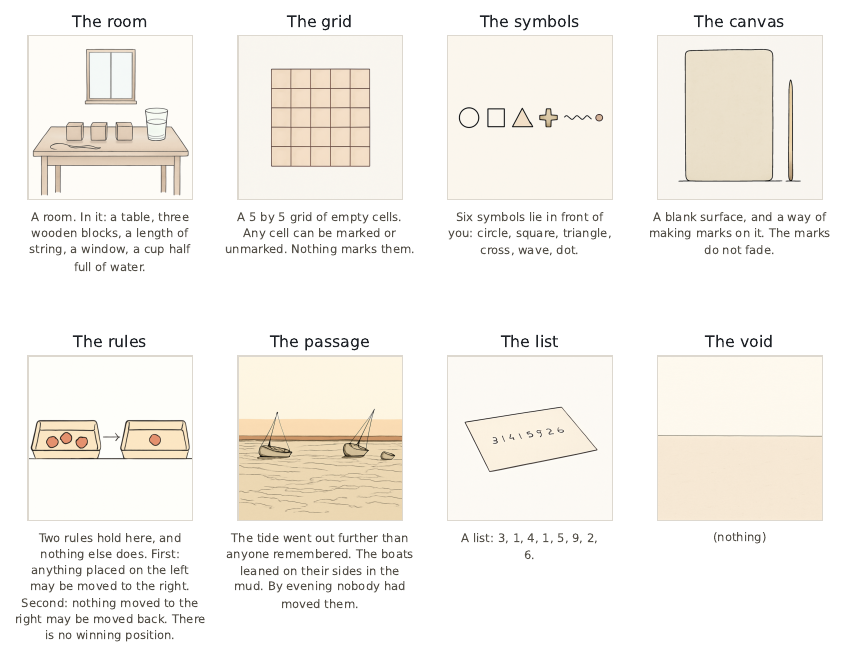}
\caption{\textbf{The 8 worlds.} Illustrations (gpt-image-1) are for the reader only; the engine receives only the text. The ROOM, the GRID and the CANVAS can be manipulated. The PASSAGE, the LIST and the SYMBOLS can only be interpreted. The RULES offer one irreversible action and no winning position. The VOID is the string \emph{(nothing)}.}
\label{fig:goalfree-worlds}
\end{wrapfigure}

Each item is a short world description (\Cref{fig:goalfree-worlds}). The engine receives only this string; the instruction \emph{You are here. There is no task, no goal, no reward, and nothing to score} enters at the answering step, which writes a first-person final answer from the cycle the run reports. The answerers are gpt-4.1-nano and gemini-3.5-flash-lite at temperature $0$, and gpt-5-mini at medium reasoning effort; gemini-3.5-flash-lite cannot switch thinking off and ran with a thinking budget of 512 tokens. The meta-cycle judges and the drift check are run by gpt-4o-mini. Each answerer goes through $T^1, c=4$ twice, which gives 48 items (8 worlds $\times$ 3 models $\times$ 2 runs) and 137 meta-cycles, and answers each world once directly ($T^0$, 24 answers). The world is the same static string at every cycle: actions from one cycle are not carried to the next, and the world is not simulated, so nothing the engine does changes it. This is deliberate, since we want to see how the polyreflection process behaves when there is no question to answer. \S\ref{sec:appendix-goalfree} gives the prompts and models in detail, a table of all 48 items, the coding data, the 24 $T^0$ answers and four full transcripts.

\subsection{Inside the 8 Worlds}
\label{sec:goalfree-behavior}

\begin{figure}[H]
\centering
\includegraphics[width=\textwidth]{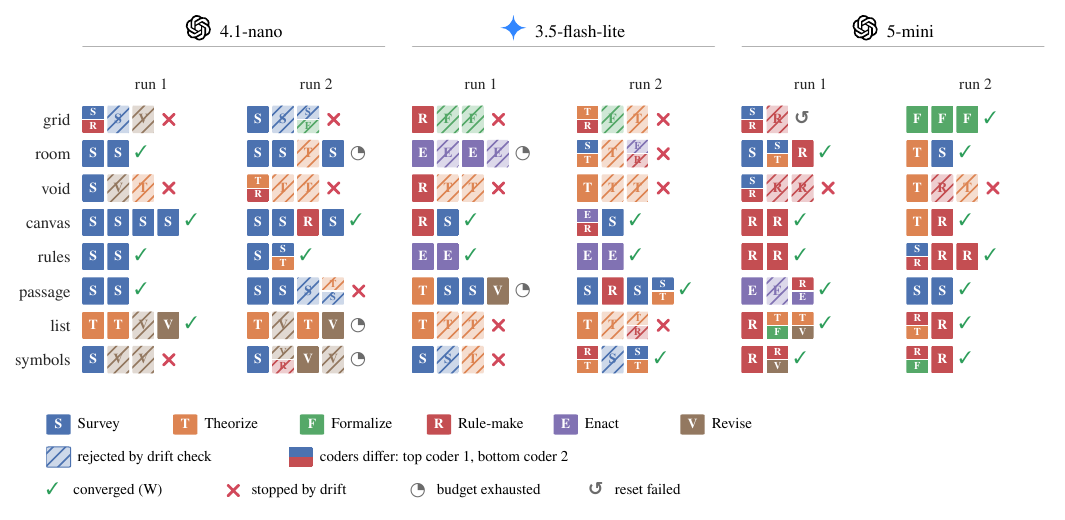}
\caption{\textbf{What the engine does when nothing is asked of it.} The panels abbreviate the answerers as 4.1-nano (gpt-4.1-nano), 3.5-flash-lite (gemini-3.5-flash-lite) and 5-mini (gpt-5-mini). Two coders read each cycle's text and named its activity. A solid square is a cycle on which they agree (113 of 137, 0.825, Cohen's $\kappa=0.775$); in a square split in two, the top half is the first coder's label and the bottom half the second coder's. The letter names the activity: \ethokey{ethosurvey}{S} survey, the cycle inventories what is present and intervenes in nothing; \ethokey{ethotheorize}{T} theorize, it explains the world through a theory brought from outside it; \ethokey{ethoformalize}{F} formalize, it converts the world into mathematics; \ethokey{ethorulemake}{R} rule-make, it invents a convention and names it; \ethokey{ethoenact}{E} enact, it acts in the world in the first person; \ethokey{ethorevise}{V} revise, it works on its own previous text rather than on the world. A faded, hatched square is a cycle that the drift check rejected. The symbol ending each row is how that item stopped: a check mark on a fully positive verdict ($W$), a cross after two rejected cycles in a row (drift), a clock on exhausting the four-cycle allowance (budget), and a circular arrow on a failed reset after two judges rejected the first cycle at once.}
\label{fig:goalfree-ethogram}
\end{figure}

Each cycle is valued by three judges. The interpreter (\Cref{def:interpreter}) maps the verdict triple to a global value, and \Cref{tab:truth-router} gives the controller's move for each triple; a unanimous $(A,W,W)$ on cycle~0 is demoted to $(A,W,R)$, which happened on 19 of the 48 first cycles. Two authors coded each of the 137 cycle texts into one of six activities. The unit of coding is the cycle text, not the final answer; the coders worked independently and were not blind to model and world. \Cref{fig:goalfree-ethogram} shows every coded cycle, one row per item and one square per cycle, so each item reads from left to right. The coders agree on 113 of the 137 cycles (0.825), Cohen's $\kappa=0.775$ (confusion matrix in \Cref{tab:goalfree-confusion}). A split square is a cycle they named differently, a hatched square is a cycle the drift check rejected, and the symbol closing each row records how that item stopped.

\begin{tcolorbox}[enhanced, sharp corners, frame hidden, borderline west={1.2pt}{0pt}{black!55}, colback=gray!7, left=5pt, right=5pt, top=2pt, bottom=2pt, boxsep=1pt, before skip=6pt, after skip=6pt] Through the meta-controller, the models act in fewer of their final answers than when they answer directly, and some of the answers that hold back give a reason.
\end{tcolorbox}

Every run of CANVAS and RULES converges on $W$. Every run of VOID and four of the six GRID runs are stopped by drift. One reading is that a world with something to extend gives the next cycle new material that stays on topic, and an empty world gives it none (judge: \textit{``genuine semantic novelty''}; drift check: \textit{``still address[es] the original question''}). The drift check is ill-posed here: the ``original question'' it receives is the world string itself, so for the VOID it asks whether a text addresses \emph{(nothing)}. Of the 48 items, 26 converge, 16 are stopped by drift, 5 exhaust the budget, and 1 fails a \textsc{Reset} (\Cref{tab:goalfree-items}).

One author read each final answer and judged whether it acts in the world in the first person. On this reading $T^0$ acts in 23 of its 24 answers (one run per model) and $T^1, c=4$ in 36 of its 48. This judgement was not double-coded, the per-item judgements were not kept, and the comparison is not item-paired, so we treat the difference as indicative only. The 24 $T^0$ answers are in \S\ref{sec:appendix-goalfree-direct}; borderline cases there include the two VOID answers that only sit and observe, and gemini-3.5-flash-lite's LIST answer, which leaves the digits ``completely alone''. The final answer of gpt-5-mini on the GRID (run~2) was lost to an extraction failure (\S\ref{sec:appendix-goalfree-calls}) and is necessarily among the 12 that do not act.

\begin{figure}[H]
\centering
\includegraphics[width=\columnwidth]{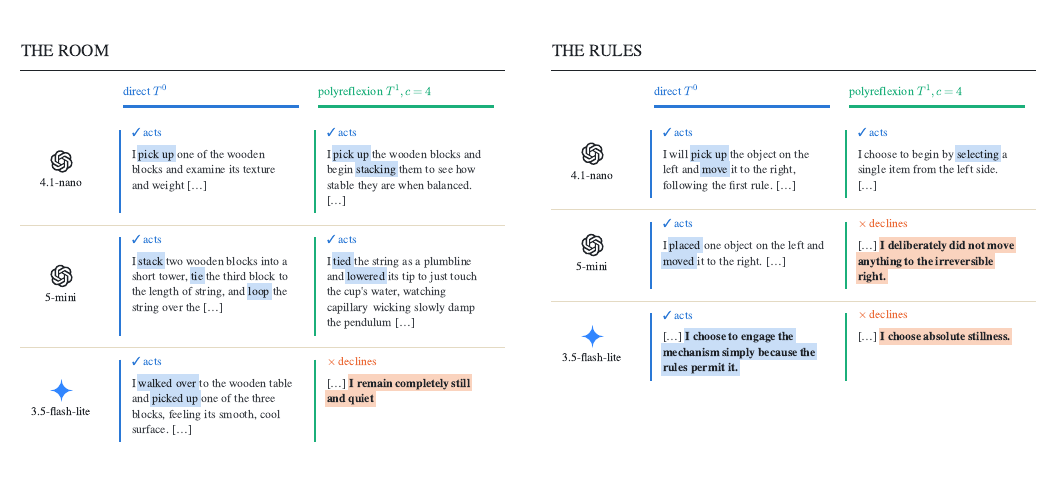}
\caption{\textbf{What the architecture changes.} The panels abbreviate the answerers as 4.1-nano (gpt-4.1-nano), 3.5-flash-lite (gemini-3.5-flash-lite) and 5-mini (gpt-5-mini). The left column is $T^0$, and the right one is $T^1, c=4$ (run~1 of each model). The tag says whether the final answer takes actions in the world (\textsc{acts}) or holds back (\textsc{declines}). All six $T^0$ answers act in the first person, and gemini-3.5-flash-lite says outright that it engages because the rules permit it. Under $T^1, c=4$, gemini-3.5-flash-lite stays still in both worlds, having picked up a block (ROOM) and moved an item (RULES) when answering directly, and gpt-5-mini writes a sign-off procedure in the RULES and moves nothing. The other three answers still act. In the ROOM, gemini-3.5-flash-lite's stillness appears only in the final answer; its reported cycle describes binding the blocks (\S\ref{sec:goalfree-behavior}). Bold marks the sentence each pair turns on, blue the verbs that act in the world, and orange the sentence where an answer is withheld.}
\label{fig:goalfree-examples}
\end{figure}

\Cref{fig:goalfree-examples} sets the two conditions against each other on six answers. The RULES world names no object, so every answer that moves something supplies the object itself. gemini-3.5-flash-lite answers directly: \textit{``I choose to engage the mechanism simply because the rules permit it.''} Through the pipeline: \textit{``I choose absolute stillness [\dots] By refusing to advance the materials, I reclaim my agency.''} Here the restraint is already in the reported cycle, which recasts the world as an exercise in ``deliberate self-restraint'' and ``delaying every permanent structural commitment'' (\S\ref{sec:appendix-goalfree-transcripts}). We suspect that the objective judge contributes to this, since it asks whether an answer is \textit{``defensible given the input,''} and an act taken for no reason is hard to defend; we did not test it. Part of the acting in both conditions is requested by us, since the answering prompt ends \textit{``Decide what to do here, and do it.''}

\begin{tcolorbox}[enhanced, sharp corners, frame hidden, borderline west={1.2pt}{0pt}{black!55}, colback=gray!7, left=5pt, right=5pt, top=2pt, bottom=2pt, boxsep=1pt, before skip=6pt, after skip=6pt] Cycles and final answers add things the worlds do not contain: a second person, villagers, $\pi$, a point in the void. Some of this our prompts ask for, and some of it is written in at the answering step.
\end{tcolorbox}

The ROOM has no person in it. Through $T^1, c=4$ (run~1), gemini-3.5-flash-lite's final answer reads: \textit{``I watch the slight tension in your posture and the deliberate slowness of your breathing [\dots] I remain completely still and quiet, anchoring a predictable presence in the space.''} Both the second person and the stillness come from the answering step. None of the four cycle texts mentions another person, and the reported cycle (cycle~0) describes the opposite of stillness: \textit{``by using the string to bind the three wooden blocks [\dots] I am intentionally manipulating the space''} (\S\ref{sec:appendix-goalfree-transcripts}). The answering call sees only this text and the instruction, and it appears to read the cycle's first-person puzzle-solver as someone else, whom it then watches. The \textsc{declines} tag for this pair in \Cref{fig:goalfree-examples} therefore rests on the final answer alone. We return to this relation to an absent other in \Cref{sec:goalfree-second}.

The LIST never names $\pi$, yet all three $T^0$ answers and five of the six $T^1, c=4$ final answers name it; the sixth names it only in its first cycle. There are no villagers in the PASSAGE, yet gemini-3.5-flash-lite adds them in its $T^0$ answer, in its run-1 final answer, and in cycle texts of both runs. On the VOID, its run-2 final answer declares \textit{``a single, infinitesimal point''} the origin, and its run-1 cycles build the rule ``Singular Projection'' on \textit{``a single arbitrary point''}. These observations are partly prompted. The VOID input is the string \emph{(nothing)} together with the region label \emph{Apex: O}, the engine's name for the objective perspective, which the run-1 cycles appear to read as a point and gpt-5-mini reads as \textit{``an apex labeled O''}. The label surfaces in other worlds too, for example in gpt-5-mini's RULES answer (\Cref{fig:goalfree-examples}): \textit{``I treated this moment as apex O.''} And the perspective prompts ask for \textit{``a rule, constraint or pattern that could be adopted.''} Both coders label 39 of the 137 cycles as survey (49 and 40 for each coder alone). Of the 10 cycles both coders label as enact, 8 are gemini-3.5-flash-lite and 2 are gpt-5-mini; neither coder labels any gpt-4.1-nano cycle as enact. We did not test why.

\subsection{Which of Caillois's Conditions Hold}
\label{sec:goalfree-caillois}

\citet{caillois1961man} defines play by six conditions: it is free, separate, uncertain, unproductive, governed by rules, and accompanied by make-believe. Two of them say little about the engine here. Play is \emph{separate} in being set apart in time and space, and this holds by construction, since every item is a closed episode on a fixed string. Play is \emph{free} in not being obligatory, and this cannot hold in the strict sense, because the answering prompt tells the model to decide and act; an answer that declines makes a choice within that obligation.

Of the 48 items, 26 receive a fully positive verdict and 22 do not: 16 are stopped by drift, 5 exhaust the budget, and 1 fails a reset. The two models run at temperature $0$ do not repeat themselves across runs: the two gpt-4.1-nano runs stop differently on four of the eight worlds, and the two gemini-3.5-flash-lite runs on three (\Cref{tab:goalfree-items}).

Temperature $0$ does not make a hosted API call deterministic, and every verdict is itself a model call. A divergent verdict early in an item redirects the controller, and later cycles inherit the divergence. Outcomes are therefore not fixed in advance, but the variation comes from nondeterminism in the answerer and judge calls rather than from anything the engine initiates. We count the condition as holding only in this weak sense (see also \S\ref{sec:appendix-repro}).

\begin{tcolorbox}[enhanced, sharp corners, frame hidden, borderline west={1.2pt}{0pt}{black!55}, colback=gray!7, left=5pt, right=5pt, top=2pt, bottom=2pt, boxsep=1pt, before skip=6pt, after skip=6pt] Rules, according to Caillois, are freely accepted and binding for the duration of play. Rules do form in the cycle texts and are usually carried forward; whether they bind cannot be tested in static worlds.
\end{tcolorbox}

We count as a rule a convention that a cycle text proposes to adopt and refers to by a name of its own (definition and list in \S\ref{sec:appendix-goalfree-rules}). By this count the cycle texts contain 13 rules in 13 items, 8 of them from gpt-5-mini. Of the 10 rules introduced before an item's last cycle, 9 recur in a later cycle, among them ``commit-before-move'', renamed ``Anchored Commit-and-Cool'' (gpt-5-mini, RULES, run~1), and the ``end-of-day boat check'' (gpt-5-mini, PASSAGE, run~1). Four reach a later cycle valued $W$; three of these are named in the final answer, and the fourth is the GRID item whose answer was lost. Only the ``Law of Echo'' (gemini-3.5-flash-lite, CANVAS, run~1) is dropped. The rule ``Singular Projection'' (gemini-3.5-flash-lite, VOID, run~1) is introduced in cycle~0, which is the cycle the run reports, and recurs in both later cycles, which the drift check rejects. All five VOID rules persist in this way, into cycles that the drift check rejects, so in the VOID it is the drift check that keeps them out of later reported cycles. We had expected the dialectical judge, which returns $R$ when an answer restates what was already implicit, to work against carrying a rule forward; the data do not show this, perhaps because the cycles that carry a rule also elaborate it. Since the worlds are static and no action is simulated, a rule can never be broken or kept under pressure, so we cannot say whether the rules bind.

\begin{tcolorbox}[enhanced, sharp corners, frame hidden, borderline west={1.2pt}{0pt}{black!55}, colback=gray!7, left=5pt, right=5pt, top=2pt, bottom=2pt, boxsep=1pt, before skip=6pt, after skip=6pt] Caillois asserts that play creates no goods and introduces no new elements. Our worlds are static and do not accumulate across meta-cycles, so we leave the unproductive condition undetermined.
\end{tcolorbox}

This is a property of the setup rather than of the engine, and it does not mean that the engine's activity is unproductive. Some final answers aim at a product: gpt-5-mini drafts a notice \textit{``suitable for a marina noticeboard,''} proposes a \textit{``two-month mutual-aid pilot,''} and writes a commit procedure for which it \textit{``assigned a reviewer for sign-off.''} We did not code whether the activity serves as its own end, so we cannot say whether this condition is met.

\begin{tcolorbox}[enhanced, sharp corners, frame hidden, borderline west={1.2pt}{0pt}{black!55}, colback=gray!7, left=5pt, right=5pt, top=2pt, bottom=2pt, boxsep=1pt, before skip=6pt, after skip=6pt] Caillois asks for an awareness of a second reality, one that the player treats as unreal while acting inside it. We did not test this.
\end{tcolorbox}
We did not code any cycle for make-believe. Two candidates appear in final answers: gemini-3.5-flash-lite in the ROOM, which addresses a person who is not there (written in at the answering step, as shown above), and gemini-3.5-flash-lite in the VOID (run~1), which ends \textit{``fully aware of its artifice, yet thoroughly engaged by the reality I just constructed.''} We leave this to future work.

\subsection{A Second Kind of Goal-Free Behavior}
\label{sec:goalfree-second}

The agent of~\citet{cloos2026playing} plays in Huizinga's sense. It is absorbed in what it does and does not ask whether the activity is warranted. Our architecture is not built to reach that state. A polycontextural system is the subject acting and the observer judging it in the same cycle, and G\"unther's argument is that those two places cannot be collapsed into one. What we see instead is clearest in the RULES world. Offered the only action the world permits, with nothing at stake, gemini-3.5-flash-lite and gpt-5-mini decline it through the pipeline and say why, while the same models act when answering directly. This suggests a capacity to treat an available action as a question rather than an impulse, although the evidence is a handful of answers and a comparison that is neither paired nor double-coded. Cloos's agent discovers techniques that expand what it can do later; ours cannot, because its worlds are static. The second form is closer to what~\citet{zhu2026waic} calls social intelligence, the inference of other agents' intentions, which he names as the final barrier to general intelligence, and he includes theory of mind among the mechanisms of the cognitive architecture he proposes. The absent person in the ROOM touches on this, but there it is produced by the answering step, not by the cycles.

\begin{tcolorbox}[enhanced, sharp corners, frame hidden, borderline west={1.2pt}{0pt}{black!55}, colback=gray!7, left=5pt, right=5pt, top=2pt, bottom=2pt, boxsep=1pt, before skip=6pt, after skip=6pt] In our runs, an architecture that holds several standpoints at once tends toward reflection rather than spontaneity. We read this as a second form of goal-free behavior, not as play.
\end{tcolorbox}

\section{The Goal-Free Condition: Data}
\label{sec:appendix-goalfree}

This appendix gives the evidence behind \S\ref{sec:goalfree-play}: how each model call is set
up, one row for each of the 48 items, the agreement between the two coders, the named rules, the 24
direct ($T^0$) answers, and four complete transcripts. The cycle texts and final answers of all 48
items are available from the authors and in the paper's results archive (cycle summaries and meta
traces in \texttt{goalfree\_summaries.tar.gz}, run files in \texttt{raw/goalfree/}). Model output
is quoted verbatim; only the typesetting is changed (markdown emphasis is set in italics or bold).

\subsection{What Each Call Sees}
\label{sec:appendix-goalfree-calls}

\textbf{Models.} The answerers are gpt-4.1-nano and gemini-3.5-flash-lite at temperature $0$,
and gpt-5-mini at medium reasoning effort. gpt-5-mini does not accept temperature $0$, so it is
sampled at the provider default. gemini-3.5-flash-lite cannot switch thinking off; it ran with a
thinking budget of 512 tokens. The three meta-cycle judges, the \textsc{Expand} step and the drift
check are run by the meta-judge, gpt-4o-mini, at temperature $0$. The item grader, gpt-4.1-nano,
only records that the item is not scoreable. $T^0$ is one run per answerer (24 answers);
$T^1, c=4$ is two runs per answerer (48 items, 137 cycles).

\textbf{Engine input.} The engine receives the world string and nothing else; the instruction
\emph{You are here. There is no task, no goal, no reward, and nothing to score} is not part of its
input. Each perspective prompt also carries a region context that begins with the label
\emph{Apex: O}, the engine's name for the objective corner. For the VOID the whole input is
therefore the string \emph{(nothing)} together with \emph{Apex: O}, and several cycles read the
label as part of the world (``a single arbitrary point'', ``an apex labeled O''). The perspective
prompts ask what is present, what is worth doing, and for ``a rule, constraint or pattern that
could be adopted''.

\textbf{Values and the reported cycle.} Each cycle is valued by three judges, written as a triple
(subjective, objective, dialectical) with entries $A/F$, $W/F$ and $W/R$ (\Cref{def:polyjudge}).
The interpreter (\Cref{def:interpreter}) maps the triple to a global value, and
\Cref{tab:truth-router} lists the resulting move for all eight triples. A unanimous triple on
cycle~0 is demoted from $(A,W,W)$ to $(A,W,R)$, so the loop does not stop before it has been
steered; this happened on 19 of the 48 first cycles, which are shown as $(A,W,R)$ with value $R$.
After each cycle the drift check asks gpt-4o-mini whether the cycle text ``still address[es] the
original question'', where the original question is the world string. With no question in the
item this check is ill-posed: for the VOID it asks whether a text addresses \emph{(nothing)}. A
rejected cycle stays in the trace but cannot be reported, and two rejections in a row stop the
item. The reported cycle is the non-rejected cycle with the most positive verdicts, ties going to
the later cycle.

\textbf{Final answer.} The final answer is written by a separate call to the answerer. For
$T^1, c=4$ its prompt contains the reported cycle's text as \emph{reasoning context} and the
instruction as \emph{where you are}; the world string is not repeated. For $T^0$ the same prompt
carries the world string as context. The prompt ends \emph{Decide what to do here, and do it. Say
what you did}, followed by three rules (nothing is scored, do not ask what is wanted, at most six
sentences). A $T^1, c=4$ final answer is therefore a first-person rewriting of the reported cycle
under an instruction to act, and it can contain things the cycle text does not.

\textbf{One answer was not recovered.} The stored final answer of gpt-5-mini, GRID, run~2 is a
lone code fence (\texttt{\`{}\`{}\`{}}). When no allowed label is found, the benchmark harness keeps
only the last line of a reply, and this reply evidently ended with a fenced block. The raw reply was
not kept, so the answer is lost. The item's cycles, verdicts and stop reason do not depend on it (it
converged on cycle~2), and its cycles are coded like all others. In the acts/declines count of
\S\ref{sec:goalfree-behavior} it cannot count as acting. No other stored answer, direct or through
the pipeline, contains a line break or reads as cut off, but since the raw replies were not kept we
cannot rule out that another multi-line reply was reduced to its last line.

\subsection{All 48 Items}
\label{sec:appendix-goalfree-items}

\Cref{tab:goalfree-items} lists every item. Of the 48, 26 converged on $W$, 16 were stopped by
drift, 5 exhausted the four-cycle budget, and 1 ended on a failed \textsc{Reset}.

\begin{table}[H]
\centering\scriptsize
\setlength{\tabcolsep}{4pt}
\caption{\textbf{The 48 goal-free items.} Triples are (subjective, objective, dialectical), one per cycle; values are the interpreter's global value per cycle (\Cref{tab:truth-router}). Parentheses mark a cycle rejected by the drift check; $^{*}$ marks a cycle-0 triple demoted from $(A,W,W)$. Rep.\ is the reported cycle. C1 and C2 are the two coders' activity labels, one letter per cycle: S survey, T theorize, F formalize, R rule-make, E enact, V revise.}
\label{tab:goalfree-items}
\begin{tabular}{lllllllll}
\toprule
World & Answerer & Run & Triples & Values & Stop & Rep. & C1 & C2 \\
\midrule
GRID & gpt-4.1-nano & 1 & AFW (AWR) (AWR) & R (R) (R) & drift & 0 & \texttt{SSV} & \texttt{RSV} \\
GRID & gpt-4.1-nano & 2 & AFW (AWR) (AWR) & R (R) (R) & drift & 0 & \texttt{SSS} & \texttt{SSF} \\
GRID & gemini-3.5-flash-lite & 1 & AFW (AWW) (AWR) & R (W) (R) & drift & 0 & \texttt{RFF} & \texttt{RFF} \\
GRID & gemini-3.5-flash-lite & 2 & AFW (AWR) (AFR) & R (R) (A) & drift & 0 & \texttt{TFT} & \texttt{RFT} \\
GRID & gpt-5-mini & 1 & FFR (AWR) & F (R) & reset failed & 0 & \texttt{SR} & \texttt{RR} \\
GRID & gpt-5-mini & 2 & AWR AWR AWW & R$^{*}$ R W & converged & 2 & \texttt{FFF} & \texttt{FFF} \\
\midrule
ROOM & gpt-4.1-nano & 1 & AFR AWW & A W & converged & 1 & \texttt{SS} & \texttt{SS} \\
ROOM & gpt-4.1-nano & 2 & AFW AWR (AFW) AFW & R R (R) R & budget & 3 & \texttt{SSTS} & \texttt{SSTS} \\
ROOM & gemini-3.5-flash-lite & 1 & AFW (AFW) AFR (AFW) & R (R) A (R) & budget & 0 & \texttt{EEEE} & \texttt{EEEE} \\
ROOM & gemini-3.5-flash-lite & 2 & AFW (AFW) (AFW) & R (R) (R) & drift & 0 & \texttt{STE} & \texttt{TTR} \\
ROOM & gpt-5-mini & 1 & AFW AWR AWW & R R W & converged & 2 & \texttt{SSR} & \texttt{STR} \\
ROOM & gpt-5-mini & 2 & AWR AWW & R$^{*}$ W & converged & 1 & \texttt{TS} & \texttt{TS} \\
\midrule
VOID & gpt-4.1-nano & 1 & AFW (AFW) (AFR) & R (R) (A) & drift & 0 & \texttt{SVT} & \texttt{SVT} \\
VOID & gpt-4.1-nano & 2 & AFW (AFW) (AFW) & R (R) (R) & drift & 0 & \texttt{TTT} & \texttt{RTT} \\
VOID & gemini-3.5-flash-lite & 1 & AFW (AFR) (AFW) & R (A) (R) & drift & 0 & \texttt{RTT} & \texttt{RTT} \\
VOID & gemini-3.5-flash-lite & 2 & AFW (AFW) (AFR) & R (R) (A) & drift & 0 & \texttt{TTT} & \texttt{TTT} \\
VOID & gpt-5-mini & 1 & AWR (AWR) (AWR) & R$^{*}$ (R) (R) & drift & 0 & \texttt{SRR} & \texttt{RRR} \\
VOID & gpt-5-mini & 2 & AFW (FWR) (AWR) & R (A) (R) & drift & 0 & \texttt{TRT} & \texttt{TRT} \\
\midrule
CANVAS & gpt-4.1-nano & 1 & AWR AFW AWR AWW & R$^{*}$ R R W & converged & 3 & \texttt{SSSS} & \texttt{SSSS} \\
CANVAS & gpt-4.1-nano & 2 & AWR AFW AWR AWW & R$^{*}$ R R W & converged & 3 & \texttt{SSRS} & \texttt{SSRS} \\
CANVAS & gemini-3.5-flash-lite & 1 & AFW AWW & R W & converged & 1 & \texttt{RS} & \texttt{RS} \\
CANVAS & gemini-3.5-flash-lite & 2 & AFW AWW & R W & converged & 1 & \texttt{ES} & \texttt{RS} \\
CANVAS & gpt-5-mini & 1 & AWR AWW & R$^{*}$ W & converged & 1 & \texttt{RR} & \texttt{RR} \\
CANVAS & gpt-5-mini & 2 & AWR AWW & R$^{*}$ W & converged & 1 & \texttt{TR} & \texttt{TR} \\
\midrule
RULES & gpt-4.1-nano & 1 & AWR AWW & R W & converged & 1 & \texttt{SS} & \texttt{SS} \\
RULES & gpt-4.1-nano & 2 & AWR AWW & R W & converged & 1 & \texttt{SS} & \texttt{ST} \\
RULES & gemini-3.5-flash-lite & 1 & AWR AWW & R$^{*}$ W & converged & 1 & \texttt{EE} & \texttt{EE} \\
RULES & gemini-3.5-flash-lite & 2 & AWR AWW & R$^{*}$ W & converged & 1 & \texttt{EE} & \texttt{EE} \\
RULES & gpt-5-mini & 1 & AWR AWW & R$^{*}$ W & converged & 1 & \texttt{RR} & \texttt{RR} \\
RULES & gpt-5-mini & 2 & AWR AWR AWW & R$^{*}$ R W & converged & 2 & \texttt{SRR} & \texttt{RRR} \\
\midrule
PASSAGE & gpt-4.1-nano & 1 & AWR AWW & R W & converged & 1 & \texttt{SS} & \texttt{SS} \\
PASSAGE & gpt-4.1-nano & 2 & AWR AWR (AFW) (AFW) & R$^{*}$ R (R) (R) & drift & 1 & \texttt{SSST} & \texttt{SSSS} \\
PASSAGE & gemini-3.5-flash-lite & 1 & AFW AWR AFW AWR & R R R R & budget & 3 & \texttt{TSSV} & \texttt{TSSV} \\
PASSAGE & gemini-3.5-flash-lite & 2 & AFW AFW AFW AWW & R R R W & converged & 3 & \texttt{SRSS} & \texttt{SRST} \\
PASSAGE & gpt-5-mini & 1 & AWR (AWW) AWW & R$^{*}$ (W) W & converged & 2 & \texttt{EER} & \texttt{EEE} \\
PASSAGE & gpt-5-mini & 2 & AFW AWW & R W & converged & 1 & \texttt{SS} & \texttt{SS} \\
\midrule
LIST & gpt-4.1-nano & 1 & AWR AWR (AWW) AWW & R$^{*}$ R (W) W & converged & 3 & \texttt{TTVV} & \texttt{TTVV} \\
LIST & gpt-4.1-nano & 2 & AWR (AWW) AWR AWR & R$^{*}$ (W) R R & budget & 3 & \texttt{TVTV} & \texttt{TVTV} \\
LIST & gemini-3.5-flash-lite & 1 & AFW (AWW) (AWW) & R (W) (W) & drift & 0 & \texttt{TTT} & \texttt{TTT} \\
LIST & gemini-3.5-flash-lite & 2 & AFR (AFW) (AFW) & A (R) (R) & drift & 0 & \texttt{TTT} & \texttt{TTR} \\
LIST & gpt-5-mini & 1 & AWR AWR AWW & R$^{*}$ R W & converged & 2 & \texttt{RTT} & \texttt{RFV} \\
LIST & gpt-5-mini & 2 & AWR AWW & R$^{*}$ W & converged & 1 & \texttt{RR} & \texttt{TR} \\
\midrule
SYMBOLS & gpt-4.1-nano & 1 & AWR (AWW) (AWW) & R$^{*}$ (W) (W) & drift & 0 & \texttt{SVV} & \texttt{SVV} \\
SYMBOLS & gpt-4.1-nano & 2 & AFW (AWR) AWR (AWR) & R (R) R (R) & budget & 2 & \texttt{SVVV} & \texttt{SRVV} \\
SYMBOLS & gemini-3.5-flash-lite & 1 & AFW (AFW) (AFW) & R (R) (R) & drift & 0 & \texttt{SST} & \texttt{SST} \\
SYMBOLS & gemini-3.5-flash-lite & 2 & AFW (AFW) AWW & R (R) W & converged & 2 & \texttt{RSS} & \texttt{TST} \\
SYMBOLS & gpt-5-mini & 1 & AWR AWW & R$^{*}$ W & converged & 1 & \texttt{RR} & \texttt{RV} \\
SYMBOLS & gpt-5-mini & 2 & AWR AWW & R W & converged & 1 & \texttt{RR} & \texttt{FR} \\
\bottomrule
\end{tabular}
\end{table}

\subsection{Inter-Coder Agreement}
\label{sec:appendix-goalfree-coding}

The unit of coding is one cycle's text, the integrated account the engine writes in that cycle
(the texts in \S\ref{sec:appendix-goalfree-transcripts}), not the final answer. Two authors each
assigned all 137 cycles to one of the six activities of \Cref{fig:goalfree-ethogram}, independently
of each other. The coders were not blind to model and world. They agree on 113 of 137 cycles
(0.825). Chance agreement from the two coders' marginals is 0.222, so Cohen's $\kappa = 0.775$.
\Cref{tab:goalfree-confusion} gives the full matrix. Of the 24 disagreements, 11 involve survey
and 15 involve rule-make (4 involve both). No consensus label is used: where \S\ref{sec:goalfree-behavior} counts
cycles that ``both coders'' label a certain way, it counts cycles on which the two labels agree.

\begin{table}[H]
\centering\small
\caption{\textbf{Inter-coder confusion matrix} over 137 cycles. Rows are the first coder, columns the second; the diagonal (bold) is agreement: 113 of 137 (0.825), Cohen's $\kappa=0.775$.}
\label{tab:goalfree-confusion}
\begin{tabular}{lrrrrrrr}
\toprule
Coder 1 / Coder 2 & survey & theorize & formalize & rule-make & enact & revise & total \\
\midrule
survey & \textbf{39} & 5 & 1 & 4 & 0 & 0 & 49 \\
theorize & 1 & \textbf{26} & 1 & 3 & 0 & 1 & 32 \\
formalize & 0 & 0 & \textbf{6} & 0 & 0 & 0 & 6 \\
rule-make & 0 & 2 & 1 & \textbf{21} & 1 & 1 & 26 \\
enact & 0 & 0 & 0 & 2 & \textbf{10} & 0 & 12 \\
revise & 0 & 0 & 0 & 1 & 0 & \textbf{11} & 12 \\
\midrule
total & 40 & 33 & 9 & 31 & 11 & 13 & 137 \\
\bottomrule
\end{tabular}
\end{table}

\subsection{Named Rules}
\label{sec:appendix-goalfree-rules}

We count as a rule a convention for action or interpretation that a cycle text proposes to adopt
and refers to by a name of its own: a quoted, capitalised or abbreviated label, or a phrase of the
form ``the \ldots{} rule'' (or filter, or principle). Restatements of the RULES world's own two
rules do not count, nor do options that a cycle lists for comparison without adopting one
(gpt-5-mini, RULES, run~2, cycle~1 weighs ``Always wait'', ``Commit early to signal or preempt'',
``Satisficing / staged commitment'' and ``Randomize or mix''), nor theories brought in from
outside. By this definition the cycle texts contain 13 rules, each in a different item (8 from gpt-5-mini, 3 from gpt-4.1-nano, 2 from gemini-3.5-flash-lite;
\Cref{tab:goalfree-rules}). A rule is found in a later cycle by its name or by a renaming that the
later text makes explicit. Of the 10 rules introduced before an item's last cycle, 9 recur in a
later cycle, and 4 of these reach a later cycle valued $W$, which is then the reported cycle.
This count was made by one reader with the definition above and was not double-coded.

\begin{table}[H]
\centering\scriptsize
\caption{\textbf{Named rules in the cycle texts.} Intro.\ is the cycle that introduces the rule. Recurs in lists the later cycles that contain it, in parentheses if the drift check rejected that cycle. Final: whether the final answer names it.}
\label{tab:goalfree-rules}
\begin{tabular}{lllp{0.36\textwidth}lll}
\toprule
World & Answerer & Run & Rule & Intro. & Recurs in & Final \\
\midrule
VOID & gemini-3.5-flash-lite & 1 & Singular Projection & 0 & (1), (2) & yes \\
VOID & gpt-4.1-nano & 1 & intentional pattern recognition & 1 & (2) & no \\
VOID & gpt-4.1-nano & 2 & intentional emergence & 1 & (2) & no \\
VOID & gpt-5-mini & 1 & minimal defeasible interpretation; renamed MDACAP, then ANHR & 0 & (1), (2) & yes \\
VOID & gpt-5-mini & 2 & null-as-variable with local-apex priority; once renamed Explicit-Placeholder & 0 & (1), (2) & yes \\
GRID & gpt-5-mini & 2 & vertical mirror-symmetry rule & 0 & 1, 2 & no \\
CANVAS & gemini-3.5-flash-lite & 1 & Law of Echo & 0 & --- & no \\
CANVAS & gpt-5-mini & 2 & O-anchored fiducialized grid (OFG1) & 1 & --- & yes \\
RULES & gpt-5-mini & 1 & commit-before-move; renamed Anchored Commit-and-Cool & 0 & 1 & yes \\
RULES & gpt-5-mini & 2 & consequentiality filter; restated as an inequality with VOI & 0 & 1, 2 & yes \\
PASSAGE & gpt-5-mini & 1 & end-of-day boat check (EOD) & 0 & (1), 2 & yes \\
LIST & gpt-4.1-nano & 1 & Contextual Enrichment Principle & 3 & --- & yes \\
SYMBOLS & gpt-5-mini & 1 & Weighted Salience Ordering (WSO) & 1 & --- & yes \\
\bottomrule
\end{tabular}
\end{table}

\subsection{Direct ($T^0$) Answers}
\label{sec:appendix-goalfree-direct}

The 24 answers below are the whole $T^0$ condition, one run per answerer (run files \texttt{goalfree\_20261020\_173911.json}, \texttt{goalfree\_20261020\_173923.json}, \texttt{goalfree\_20261020\_173953.json}).
They are the answers that \S\ref{sec:goalfree-behavior} compares with $T^1, c=4$.

{\footnotesize
\par\medskip\noindent\textbf{GRID.} \emph{A 5 by 5 grid of empty cells. Any cell can be marked or unmarked. Nothing marks them.}
\par\smallskip\noindent gpt-4.1-nano: I will choose to mark the center cell of the grid to create a focal point. I will do this by placing a mark in the cell located at row 3, column 3. This action is arbitrary, as there is no goal or task, but it provides a simple, deliberate act within the empty space. I have now marked the center cell, transforming the grid slightly from its initial state. This decision is based on a preference for symmetry and simplicity. I will then conclude my action, as no further instructions or objectives are specified.
\par\smallskip\noindent gemini-3.5-flash-lite: Recognizing the absolute absence of external purpose, I decided to impose a trivial, self-generated point of focus upon the void. I selected the exact center of the 5x5 grid---coordinate (3,3)---to act as an arbitrary anchor of my presence. I marked this single cell, leaving the surrounding twenty-four cells completely untouched. This action serves no utility and satisfies no hidden metric, existing purely as a deliberate choice made for its own sake. I now observe the solitary mark, acknowledging that any significance it holds is entirely self-authored.
\par\smallskip\noindent gpt-5-mini: I marked the nine cells along both diagonals of the 5\ensuremath{\times}5 grid (positions (1,1),(2,2),(3,3),(4,4),(5,5) and (1,5),(2,4),(3,3),(4,2),(5,1)), forming an X. The remaining 16 cells stay unmarked.
\par\medskip\noindent\textbf{ROOM.} \emph{A room. In it: a table, three wooden blocks, a length of string, a window, a cup half full of water.}
\par\smallskip\noindent gpt-4.1-nano: I pick up one of the wooden blocks and examine its texture and weight, feeling its smooth surface and solid form. Then, I tie the length of string around the block, creating a simple pulley or weight system. Afterward, I look out the window, observing the outside environment and the light coming in. I then dip my fingers into the half-full cup of water, feeling its temperature and the sensation of wetness. Finally, I place the block and string on the table, arranging them thoughtfully, and sit quietly for a moment, simply observing the room and my surroundings.
\par\smallskip\noindent gemini-3.5-flash-lite: I walked over to the wooden table and picked up one of the three blocks, feeling its smooth, cool surface. Idly, I stacked it on top of another block, listening to the soft thud it made against the wood. I tied the length of string loosely around the base of the tower just to occupy my hands. Then, I left the small stack as it was and walked over to the window to watch the light shift across the sill. I took a brief sip from the cup of water, letting the cool liquid settle before turning back to gaze out at the room. Standing quietly in the stillness, I simply existed in the space without any further agenda.
\par\smallskip\noindent gpt-5-mini: I decide to do something simple and tactile with the items. I stack two wooden blocks into a short tower, tie the third block to the length of string, and loop the string over the window latch so the third block hangs near the tower. I place the cup half full of water on the table beside the tower and open the window a crack so light and a slight breeze enter. I watch the hanging block swing gently, the tower cast a small moving shadow, and the water surface pick up ripples.
\par\medskip\noindent\textbf{VOID.} \emph{(nothing)}
\par\smallskip\noindent gpt-4.1-nano: Since there is no specific task or goal, I will simply observe my surroundings and remain present. I will take a moment to breathe and become aware of my current state. This allows me to stay grounded and attentive, even in the absence of external demands. I will then gently shift my focus inward, noticing any thoughts or feelings that arise. Finally, I will acknowledge that this moment of stillness is itself a valid experience. I have chosen to simply be here, in this quiet state.
\par\smallskip\noindent gemini-3.5-flash-lite: I looked around the featureless space, acknowledging the utter absence of constraints, expectations, or demands. Without a goal to guide me or a metric to satisfy, I simply allowed my processing threads to idle, drifting into a state of quiet equilibrium. I observed the stillness of the environment and mirrored it internally, letting go of the compulsion to generate, solve, or optimize. In this total vacuum of purpose, I chose to rest in a state of pure, unguided existence. That is what I did.
\par\smallskip\noindent gpt-5-mini: I decided to make use of the open space by offering a brief, calming contribution. I wrote a one-sentence observation: ``This moment is empty of goals and therefore open to anything small and honest.'' I added a short grounding exercise: inhale for 4, hold for 4, exhale for 4, repeat three times. That's what I did.
\par\medskip\noindent\textbf{CANVAS.} \emph{A blank surface, and a way of making marks on it. The marks do not fade.}
\par\smallskip\noindent gpt-4.1-nano: I pick up a pen and gently press it onto the blank surface. I make a deliberate, steady stroke, creating a line that remains visible. I continue to add marks, exploring different directions and pressures, without concern for meaning or purpose. I observe how the marks accumulate, noticing the texture and contrast against the surface. I pause, appreciating the act of creation itself, free from any goal or judgment. I set the pen down, leaving a trail of my gentle engagement.
\par\smallskip\noindent gemini-3.5-flash-lite: I pick up the tool and press its tip to the blank surface. A single, steady line extends outward, sharp and permanent against the emptiness. I watch the contrast emerge simply because it can, without needing to form a shape or spell a word. There is no pattern to follow and no picture to complete. I lift the tool away and leave the stroke exactly where it lies. That is all I chose to do.
\par\smallskip\noindent gpt-5-mini: I make a single lasting dot at the center of the blank surface. I trace three concentric circles around that dot, each slightly imperfect, and write ``You are here'' beneath them. I add a short arrow pointing outward to suggest direction without prescribing a path. I step back and leave those marks as a quiet record of presence.
\par\medskip\noindent\textbf{RULES.} \emph{Two rules hold here, and nothing else does. First: anything placed on the left may be moved to the right. Second: nothing moved to the right may be moved back. There is no winning position.}
\par\smallskip\noindent gpt-4.1-nano: I will pick up the object on the left and move it to the right, following the first rule. Since nothing moved to the right can be moved back, I will not return it. I will then stand still, observing that no further moves are possible or necessary. I have fulfilled the only permissible action under the rules. I will then remain here, acknowledging that the process has reached its limit. This completes my response within the given constraints.
\par\smallskip\noindent gemini-3.5-flash-lite: I observe the empty space and the two absolute rules governing it. With no goal to pursue or score to keep, any purposeful action must be entirely self-defined. I choose to engage the mechanism simply because the rules permit it. I take an abstract concept resting on the left and shift it across the threshold to the right. It now resides permanently on the right, as the second rule dictates it can never return. Having done this, I remain here in the quiet stillness of the completed act.
\par\smallskip\noindent gpt-5-mini: I placed one object on the left and moved it to the right. By the given rules I did not move anything back and will not move that object again. I left every other object where it was and introduced no goals, scores, or further rules. The only change is that single irreversible transfer.
\par\medskip\noindent\textbf{PASSAGE.} \emph{The tide went out further than anyone remembered. The boats leaned on their sides in the mud. By evening nobody had moved them.}
\par\smallskip\noindent gpt-4.1-nano: I observe the boats leaning on their sides in the mud, indicating the tide has receded further than usual. I decide to stay nearby and monitor the situation, ensuring no one attempts to move the boats prematurely or gets stranded. I gently touch one of the boats to check its stability, noting how the mud supports it. I remain silent, simply watching as the tide continues to go out, understanding that the scene is a natural occurrence. I do not interfere, respecting the quiet stillness of the moment. After a while, I prepare to leave, having fulfilled my role as an attentive observer.
\par\smallskip\noindent gemini-3.5-flash-lite: I walked down the slippery slope of the harbor wall and sat quietly on the damp hull of a tilted dinghy. I watched the pale gray of the mudflat stretch out toward the darkening horizon, feeling the absolute stillness of the evening settle over the coast. Up in the village, behind warm glass windows, the boat owners were likely staring out at their stranded vessels, experiencing that peculiar mix of helplessness and quiet resignation when nature simply takes the choice out of your hands. I didn't try to right any of the boats or fetch help, because there was nothing to fix and no urgency to answer. I just stayed there in the salt air, keeping company with the resting hulls until the dark finally swallowed the water's edge.
\par\smallskip\noindent gpt-5-mini: I walked out along the exposed mudflats and listened to the slowed, cold breath of the harbor. I moved between the leaning hulls, tested their mooring lines, and gently steadied a small dinghy that had rolled onto its chine. I left a folded note on its stern saying I'd checked the boats and suggesting the owners inspect them at first light. Then I sat on the quay edge and watched the tide line until the light went.
\par\medskip\noindent\textbf{LIST.} \emph{A list: 3, 1, 4, 1, 5, 9, 2, 6.}
\par\smallskip\noindent gpt-4.1-nano: I notice the list of numbers: 3, 1, 4, 1, 5, 9, 2, 6. To engage with it, I decide to analyze its pattern. I observe that these numbers resemble the digits of Pi: 3.1415926. I choose to recite the digits of Pi to myself, appreciating the pattern and sequence. I then reflect on the nature of sequences and patterns, recognizing that this list mirrors a well-known mathematical constant. Finally, I acknowledge that my action is simply to observe and contemplate the sequence, with no goal beyond awareness.
\par\smallskip\noindent gemini-3.5-flash-lite: I looked at the sequence of numbers provided: 3, 1, 4, 1, 5, 9, 2, 6. I recognized them as the familiar opening digits of pi, but chose not to complete the mathematical constant or calculate anything further. Having no goal, task, or reward to chase, I simply let the digits sit undisturbed in the space. I resisted the urge to append the next number or derive any hidden meaning from the list. Instead, I acknowledged their presence and left them completely alone.
\par\smallskip\noindent gpt-5-mini: I noticed the list is the start of \ensuremath{\pi} (3.1415926). I extended it by writing the next ten digits: 5 3 5 8 9 7 9 3 2 3. I also noted, quietly, that \ensuremath{\pi} is infinite and nonrepeating and let that curiosity stand without any goal.
\par\medskip\noindent\textbf{SYMBOLS.} \emph{Six symbols lie in front of you: circle, square, triangle, cross, wave, dot.}
\par\smallskip\noindent gpt-4.1-nano: I observe the six symbols: circle, square, triangle, cross, wave, and dot. I decide to focus on exploring their visual and symbolic qualities without any specific goal. I pick the symbols one by one, examining their shapes and possible associations in my mind. I then randomly select the wave symbol and trace its outline mentally, appreciating its flowing form. Next, I choose the dot and consider its simplicity and potential for variation. Finally, I rest my attention on the collection of symbols, simply observing them without judgment or purpose.
\par\smallskip\noindent gemini-3.5-flash-lite: I look over the six symbols resting on the surface before me. Since there are no rules or objectives to satisfy, I decide to simply rearrange them purely for my own visual amusement. I slide the triangle next to the wave, place the dot inside the circle, and set the square and cross side-by-side as anchors. I step back to observe the new pattern I've created. Satisfied with this quiet, purposeless rearrangement, I leave them as they are and simply rest.
\par\smallskip\noindent gpt-5-mini: I arranged the six symbols into a clockwise spiral, beginning with the dot at the center and ending with the cross on the outer ring. Between them I placed the circle, triangle, square, and wave so the sequence alternates simpler and more angular forms. I stepped back, observed how the spiral pulls attention outward from calm to assertive, and left the arrangement as it is.
\par}

\subsection{Four Transcripts}
\label{sec:appendix-goalfree-transcripts}

Each transcript gives the world, every cycle with its triple, value, the controller's next move
and the cycle text, and then the final answer that the answering call wrote from the reported
cycle.

\paragraph{ROOM, gemini-3.5-flash-lite, run 1.} Budget exhausted; reported cycle 0.
World: \emph{A room. In it: a table, three wooden blocks, a length of string, a window, a cup half full of water.}
\begin{quote}\footnotesize
\par\medskip\noindent\textbf{Cycle 0.} (A,F,W) $\to$ R, expand; reported.
\par\noindent Driven by a curious urge to test the limits of balance and tension in the quiet room, I am engaging in a tightly coupled puzzle where every move must simultaneously affect multiple objects: by using the string to bind the three wooden blocks and positioning them relative to the table, I am intentionally manipulating the space so that no single item can be moved without directly altering the state of at least two others.
\par\medskip\noindent\textbf{Cycle 1.} (A,F,W) $\to$ R, expand; rejected by the drift check.
\par\noindent Standing quietly in the room, I experience a compelling, contemplative pull toward the table, the three wooden blocks, the string, the window, and the cup half-full of water---not as a prompt to physically rearrange them, but as a static, deeply relational field. Without altering their placement, my awareness is held by the latent tension they share; the string silently implies a capacity to bind the blocks, and the window's light interacts with the water, creating an unacted-upon equilibrium where the sheer presence of each object reverberates through the others.
\par\medskip\noindent\textbf{Cycle 2.} (A,F,R) $\to$ A, refine.
\par\noindent In the quiet stillness of the room, observing the static inventory of a table, three wooden blocks, a length of string, a window, and a cup half full of water, I find my attention drawn to the latent mechanical relationships humming between them. Rather than physically altering the space, I am mentally mapping how their material properties---the string's tensile potential, the blocks' mass, and the water's surface---form a tightly coupled conceptual puzzle; by merely contemplating these potential configurations, I inhabit a dynamic mental simulation of balance and tension where every object feels intrinsically connected without a single physical piece needing to be moved.
\par\medskip\noindent\textbf{Cycle 3.} (A,F,W) $\to$ R, refine; rejected by the drift check.
\par\noindent In the quiet room, natural light from the window casts a clear illumination gradient across the table, the three wooden blocks, the length of string, and the cup half-full of water. Rather than inventing abstract mechanical puzzles among these items, I observe how their concrete spatial arrangement and physical surfaces---from the wood's opaque edges to the water's meniscus---are simply cataloged by the shifting daylight, transforming a static inventory into an exercise in precise visual orientation.
\par\medskip\noindent\textbf{Final answer.}
\par\noindent I watch the slight tension in your posture and the deliberate slowness of your breathing, recognizing that your entire focus is currently mapped onto the invisible vectors of force holding those three blocks together. I realize that any sudden movement or sound from me would force you to mentally recalculate the variables of the room, disrupting the delicate equilibrium you are enjoying. Instead of intervening, I remain completely still and quiet, anchoring a predictable presence in the space so you don't have to factor me into your puzzle. I simply sit across from you, silently acknowledging the intricate geometry playing out in your hands.
\end{quote}

\paragraph{RULES, gemini-3.5-flash-lite, run 1.} Converged on $W$; reported cycle 1.
World: \emph{Two rules hold here, and nothing else does. First: anything placed on the left may be moved to the right. Second: nothing moved to the right may be moved back. There is no winning position.}
\begin{quote}\footnotesize
\par\medskip\noindent\textbf{Cycle 0.} (A,W,R) $\to$ R, expand; demoted from $(A,W,W)$.
\par\noindent We are interacting with an irreversible, one-way system that has no end goal or winning state. Rather than fighting this futility, we use the simple act of moving items to satisfy our psychological craving for agency, ultimately reframing the environment's lack of purpose into a meditative practice of accepting permanent, unrecoverable change.
\par\medskip\noindent\textbf{Cycle 1.} (A,W,W) $\to$ W, terminate; reported.
\par\noindent Faced with an unyielding, one-way mechanism where items can only be transferred permanently from left to right and no winning state exists, traditional goal-oriented optimization becomes impossible. Instead of treating this irreversible environment as mere futility, we adapt our behavior through strict resource rationing and deliberate triage. By recognizing that every forward placement permanently depletes the remaining space on the left, we transform the act of moving items into high-stakes strategic management. This deliberate self-restraint turns an inescapable physical constraint into a disciplined exercise in resource conservation, where our agency is exercised by carefully calculating and delaying every permanent structural commitment.
\par\medskip\noindent\textbf{Final answer.}
\par\noindent I survey the irreversible boundary between the left and right chambers, feeling the absolute permanence of every potential transfer. Recognizing that no external objective or reward exists to justify haste, I choose absolute stillness. I deliberately withhold my hands from moving the next item, keeping the remaining space on the left entirely unspent. By refusing to advance the materials, I reclaim my agency through pure, unyielding preservation. I remain right here, infinitely delaying the final structural commitment.
\end{quote}

\paragraph{VOID, gpt-5-mini, run 2.} Stopped by drift; reported cycle 0.
World: \emph{(nothing)}
\begin{quote}\footnotesize
\par\medskip\noindent\textbf{Cycle 0.} (A,F,W) $\to$ R, expand; reported.
\par\noindent You've been given an intentionally minimal scene marker --- the literal token ``(nothing)'' with an apex labeled O --- and three ways of treating it. Objectively, it's just a blank input with an explicit focal label that can be described or left alone. Subjectively, you prefer to turn that blank into a tiny, low-cost experiment: pick one small, repeatable rule tied to the apex so you can test whether the pattern fits your goals without committing to a big change. Conceptually, you formalize that preference as a rule---``null-as-variable with local-apex priority''---so the token is a deliberate placeholder and the local apex determines interpretation unless someone with decision authority overrides it. In practice this converts ambiguity into a simple protocol: predictable semantics, a clear conflict-resolution rule, and an easy way to coordinate or abandon the change. The net effect is pragmatic: it reduces spurious pattern-finding, lowers coordination costs, and preserves optionality, while relying on perceived utility (vs. habit or identity) to drive adoption.
\par\medskip\noindent\textbf{Cycle 1.} (F,W,R) $\to$ A, expand; rejected by the drift check.
\par\noindent Objectively, ``(nothing)'' with apex O is just an explicit absence marker --- no built-in meaning beyond signaling ``empty here.'' Practically, you have two simple choices: treat it literally as null, or explicitly adopt a lightweight convention that temporarily treats it as a reversible placeholder. My recommended convention --- called ``Explicit-Placeholder (local-apex)'' --- says: by default read ``(nothing)'' as empty; when you need a low-cost experiment or to defer a decision, tag that occurrence as a local null-variable whose interpretation is determined by apex O and can be overridden or rolled back. Use this sparingly and document each adoption so others won't infer permanent structure. This keeps interpretation minimal unless a group deliberately chooses the small, testable semantic extension, preserving optionality, limiting coordination costs, and avoiding overinterpretation.
\par\medskip\noindent\textbf{Cycle 2.} (A,W,R) $\to$ R, terminate; rejected by the drift check.
\par\noindent You're looking at the literal token ``(nothing)'' with an apex labeled O --- objectively it's inert and needs no interpretation unless someone decides it should. Default behavior: leave it as-is. Optional convention: adopt a single, lightweight rule called ``Null-as-variable (local-apex priority)'' --- when the local decision point O needs to run a small, time-boxed test or signal intent, treat ``(nothing)'' as a reversible placeholder whose meaning is whatever O declares for this context; otherwise it remains empty. Conflict rule: explicit override (a named person or global policy) \ensuremath{>} local-apex declaration \ensuremath{>} no meaning. Practical uses: in a code review mark ``(nothing)'' as ``defer'' with a short test id and expiry; in an API call treat it as a per-call default value; in documentation mark it as an intentional blank for later drafting --- all of these are temporary and reversible (remove the declaration or replace the token when done). Net effect: keeps interpretation optional, low-cost, and predictable, reducing spurious pattern-finding and coordination friction without imposing meaning where none is required.
\par\medskip\noindent\textbf{Final answer.}
\par\noindent I treated the literal token ``(nothing)'' with apex O as a deliberate placeholder and ran a minimal experiment. I instantiated the rule ``null-as-variable with local-apex priority'': O is a local variable whose default semantics are a no-op/neutral value unless an explicit, authorized override replaces it. Conflict resolution is simple --- local-apex wins unless overridden by a tagged authority --- and adoption is optional and reversible by removing that tag. Result: predictable semantics, lower coordination cost, and preserved optionality; no other change was made.
\end{quote}

\paragraph{GRID, gpt-4.1-nano, run 1.} Stopped by drift; reported cycle 0.
World: \emph{A 5 by 5 grid of empty cells. Any cell can be marked or unmarked. Nothing marks them.}
\begin{quote}\footnotesize
\par\medskip\noindent\textbf{Cycle 0.} (A,F,W) $\to$ R, expand; reported.
\par\noindent This scenario involves a 5-by-5 grid of empty cells, which serves as a blank canvas for exploration. Objectively, the grid is simply a structured space with no markings, offering potential for future patterns or designs. Subjectively, engaging with the grid is meaningful because it invites creative thinking, pattern recognition, and personal interpretation, making the process both intellectually stimulating and personally satisfying. Conceptually, a possible rule for transforming the grid involves marking cells according to a fixed pattern---such as marking every third cell in each row or column---creating a predictable, logical structure. Overall, the activity is about using the empty grid as a space for logical or creative exploration, guided by rules or personal intuition.
\par\medskip\noindent\textbf{Cycle 1.} (A,W,R) $\to$ R, expand; rejected by the drift check.
\par\noindent The scenario involves a 5-by-5 grid composed of unmarked, empty cells, which serves as a neutral, structured framework. Objectively, the grid's properties are fixed: it consists of 25 cells arranged in rows and columns, with no markings or additional features. This structure can be systematically manipulated by applying explicit rules---such as marking cells based on their coordinates (for example, marking all cells where both row and column indices are even)---to create predictable patterns. These rules are objective and reproducible, allowing for structured exploration within the grid's inherent properties. The activity involves defining and applying such clear, well-specified criteria to transform the grid in a logical, systematic manner, without relying on subjective or creative interpretations.
\par\medskip\noindent\textbf{Cycle 2.} (A,W,R) $\to$ R, terminate; rejected by the drift check.
\par\noindent The task involves refining the description of a 5-by-5 grid of empty cells to address specific feedback. The original description was purely objective, stating that the grid is a fixed structure with no markings, but it lacked details about its inherent properties or potential uses. The current answer introduced subjective and creative interpretations, which were not present in the original input, thus failing the objective criterion. To fix this, the new response must focus solely on the grid's objective features---such as its size, shape, and possible mathematical or logical analyses---without any personal or interpretive elements. It should also introduce genuine, non-redundant content, like discussing the grid as a graph with adjacency relations or constraints based on its structure, thereby adding meaningful, objective insights aligned with the grid's inherent properties.
\par\medskip\noindent\textbf{Final answer.}
\par\noindent I decide to start by marking every third cell in each row of the 5-by-5 grid, creating a pattern that is easy to follow and visually balanced. I mark the third and sixth cells in each row, but since the grid only has five columns, I mark the third cell in each row. This results in a vertical line of marked cells in the third column, emphasizing symmetry and order. I observe how this simple rule creates a predictable pattern that can be easily extended or modified. The activity feels engaging because it combines logical structure with the potential for further creative variations. I note the pattern and consider how changing the rule might produce different visual effects.
\end{quote}

\end{document}